\documentclass[letterpaper]{article}
\usepackage{aaai24}
\makeatletter
\@ifpackageloaded{theapa}{%
}{
 \usepackage{natbib}
}
\makeatother

\usepackage{xparse}

\usepackage{amsmath}
\usepackage{amsthm}
\usepackage{amssymb}
\usepackage{amsfonts}

\usepackage{tabularx}
\usepackage{booktabs}   
\usepackage{multirow}
\usepackage{subcaption}

\usepackage{manfnt}             

\usepackage{xspace}             
\usepackage{relsize}            
\usepackage{adjustbox}          
\usepackage{diagbox}            
\usepackage{pdflscape}          

\usepackage{microtype}

\usepackage{graphicx}
\usepackage{url}
\usepackage{comment}            
\usepackage{colortbl}
\usepackage[dvipsnames,svgnames]{xcolor}

\usepackage{algorithm}
\usepackage{algpseudocode}

\input{math.sty}
\input{general.sty}
\input{abbrev.sty}
\input{strips.sty}

\allowdisplaybreaks

\definecolor{mygray}{gray}{.9}

\begin{document}
%
\title{Transferable Adversarial Face Attack with Text Controlled Attribute}
\author {
    Wenyun Li\textsuperscript{\rm 1,2},
    Zheng Zhang\textsuperscript{\rm 1,2 },
    Xiangyuan Lan\textsuperscript{\rm 2,3 },
    Dongmei Jiang\textsuperscript{\rm 2}
}
\affiliations {
    \{
    \textsuperscript{\rm 1}Harbin Institute of Technology,   
    \textsuperscript{\rm 2}Peng Cheng Laboratory \}, Shenzhen, China\\
    \textsuperscript{\rm 3}Pazhou Laboratory (Huangpu), Guangzhou, China\\
    \textbf{Correspondence to}: darrenzz219@gmail.com, lanxy
    @pcl.ac.cn
}
\maketitle
\begin{abstract}
\begin{quote}
Traditional adversarial attacks typically produce adversarial examples under norm-constrained conditions, whereas unrestricted adversarial examples are free-form with semantically meaningful perturbations. Current unrestricted adversarial impersonation attacks exhibit limited control over adversarial face attributes and often suffer from low transferability. In this paper, we propose a novel Text Controlled Attribute Attack (TCA$^2$) to generate photorealistic adversarial impersonation faces guided by natural language. Specifically, the category-level personal softmax vector is employed to precisely guide the impersonation attacks. Additionally, we propose both data and model augmentation strategies to achieve transferable attacks on unknown target models. Finally, a generative model, \textit{i.e}, Style-GAN, is utilized to synthesize impersonated faces with desired attributes. Extensive experiments on two high-resolution face recognition datasets validate that our TCA$^2$ method can generate natural text-guided adversarial impersonation faces with high transferability. We also evaluate our method on real-world face recognition systems, \textit{i.e}, Face++ and Aliyun, further demonstrating the practical potential of our approach. The source code is available at \url{https://github.com/phd-research-ai/TCA2.git}.
\end{quote}
\end{abstract}

\section{Introduction}
\label{sec:intro}
Recent studies have shown that deep learning-based face recognition (FR) model systems are vulnerable to adversarial examples \cite{vakhshiteh2021adversarial,dong2019efficient,ZhangWLSZ22}. Adding deliberately designed but imperceptible noise to a clean image can fool even state-of-the-art commercial FR models \cite{ali2021classical}. This vulnerability poses a direct threat to socially critical applications such as customs inspection and mobile device face identification. Consequently, the security community has increasingly focused on studying adversarial examples to improve the robustness and generalization ability of existing FR systems.

Early well-studied works focus on norm-constrained attacks, where the adversarial image lies within an $\epsilon$-neighborhood of a real sample using the $L_p$ distance metric to evaluate the strength of the adversarial example \cite{szegedy2013intriguing,Wang0WSL21}. Common values for $p$ include 0, 2, and $\infty$. With a sufficiently small $\epsilon$, the adversarial image is quasi-indistinguishable from the natural sample. Such norm-based attacks have demonstrated outstanding adversarial performance against face recognition (FR) systems. However, there are limitations: 1) Despite being designed to be indistinguishable, norm-based attacks may still contain visible perturbations, making them detectable by human eyes or specially designed detectors \cite{massoli2021detection};  2) Numerous adversarial defense methods have been introduced to counter norm-based attacks, leading to a relatively low attack success rate against real-world FR models \cite{madry2017towards}.

\begin{figure*}
     \centering
     \begin{subfigure}[b]{0.155\textwidth}
         \centering
         \includegraphics[width=\textwidth]{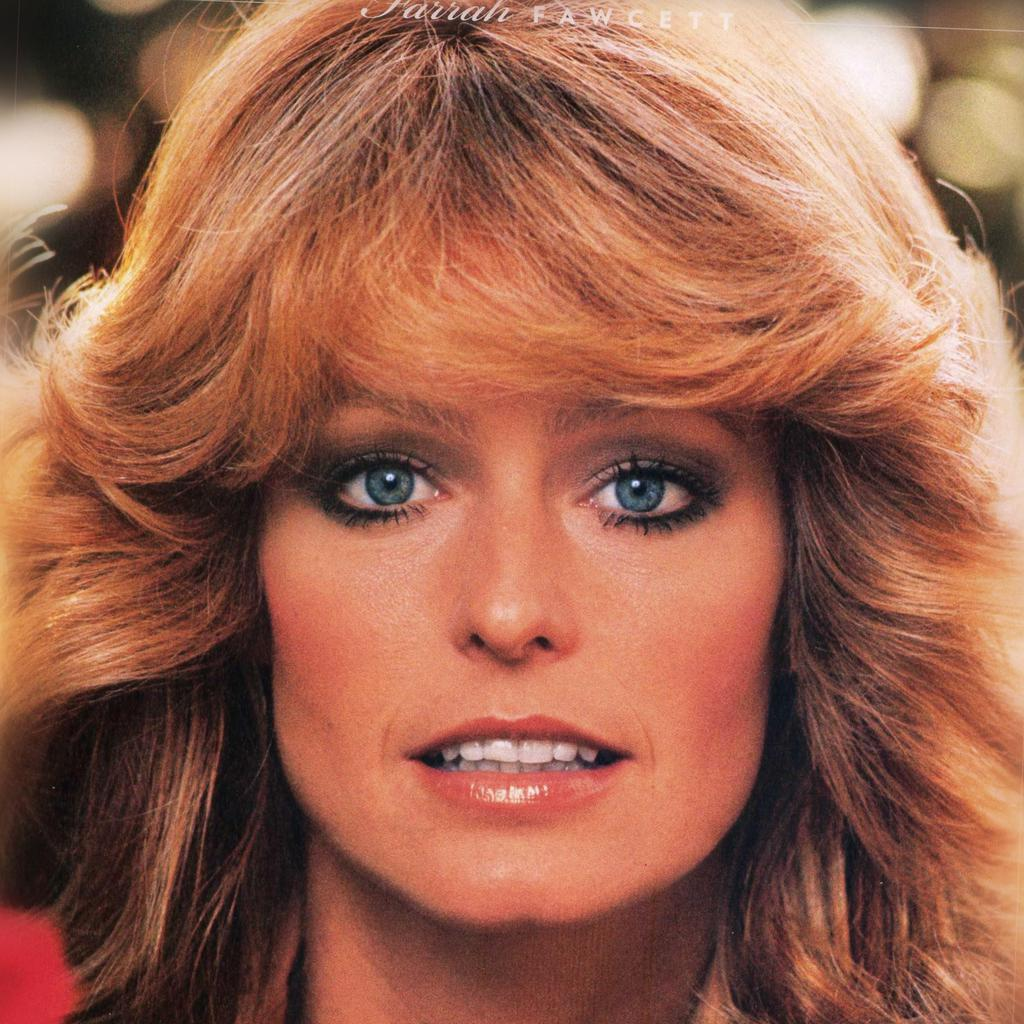}
         \caption{\textcolor{DarkGreen}{Source}}
     \end{subfigure}
     \hfill
     \begin{subfigure}[b]{0.155\textwidth}
         \centering
         \includegraphics[width=\textwidth]{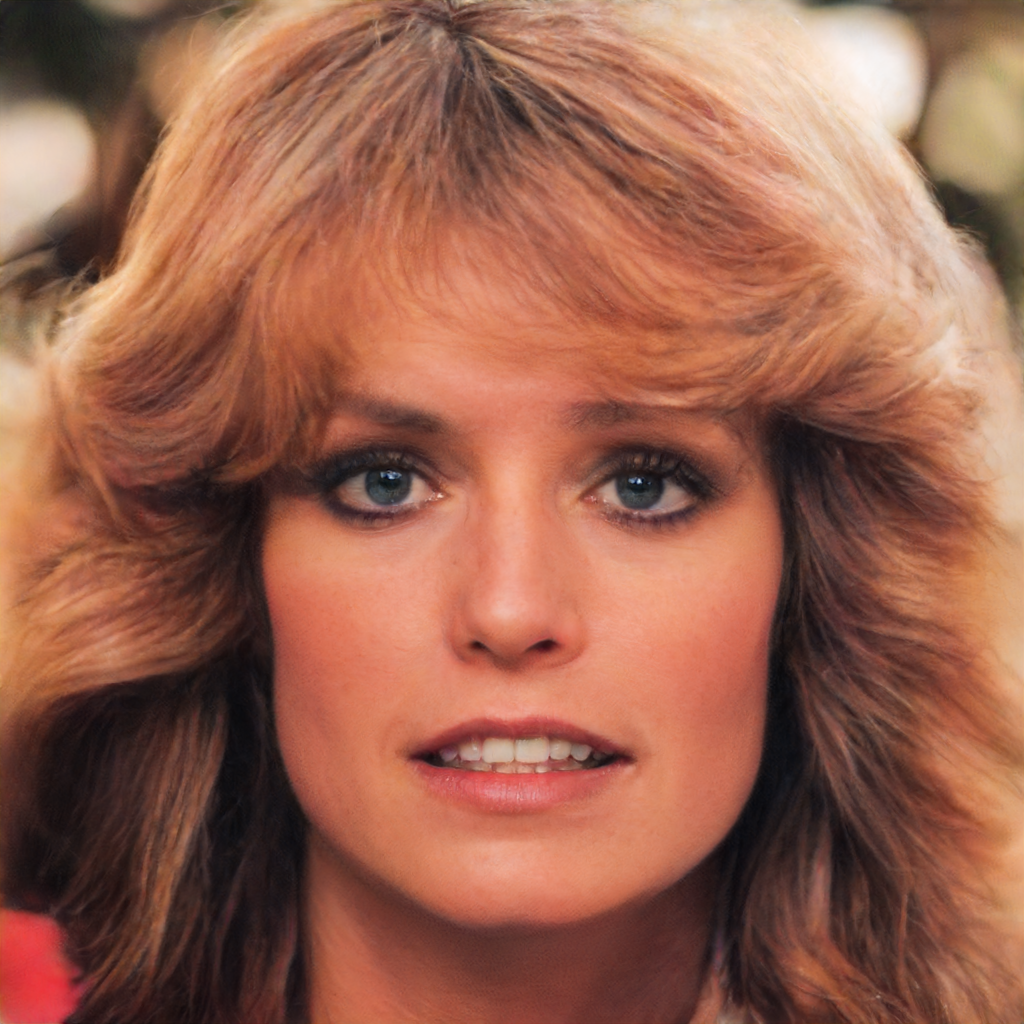}
         \caption{Inversion}
     \end{subfigure}
     \hfill
     \begin{subfigure}[b]{0.155\textwidth}
         \centering
         \includegraphics[width=\textwidth]{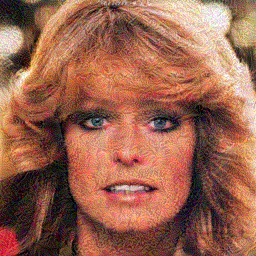}
         \caption{PGD}
     \end{subfigure}
     \hfill
     \begin{subfigure}[b]{0.155\textwidth}
         \centering
         \includegraphics[width=\textwidth]{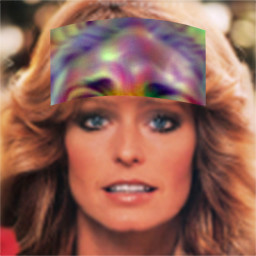}
         \caption{AdvHat}
     \end{subfigure}
     \hfill
     \begin{subfigure}[b]{0.155\textwidth}
         \centering
         \includegraphics[width=\textwidth]{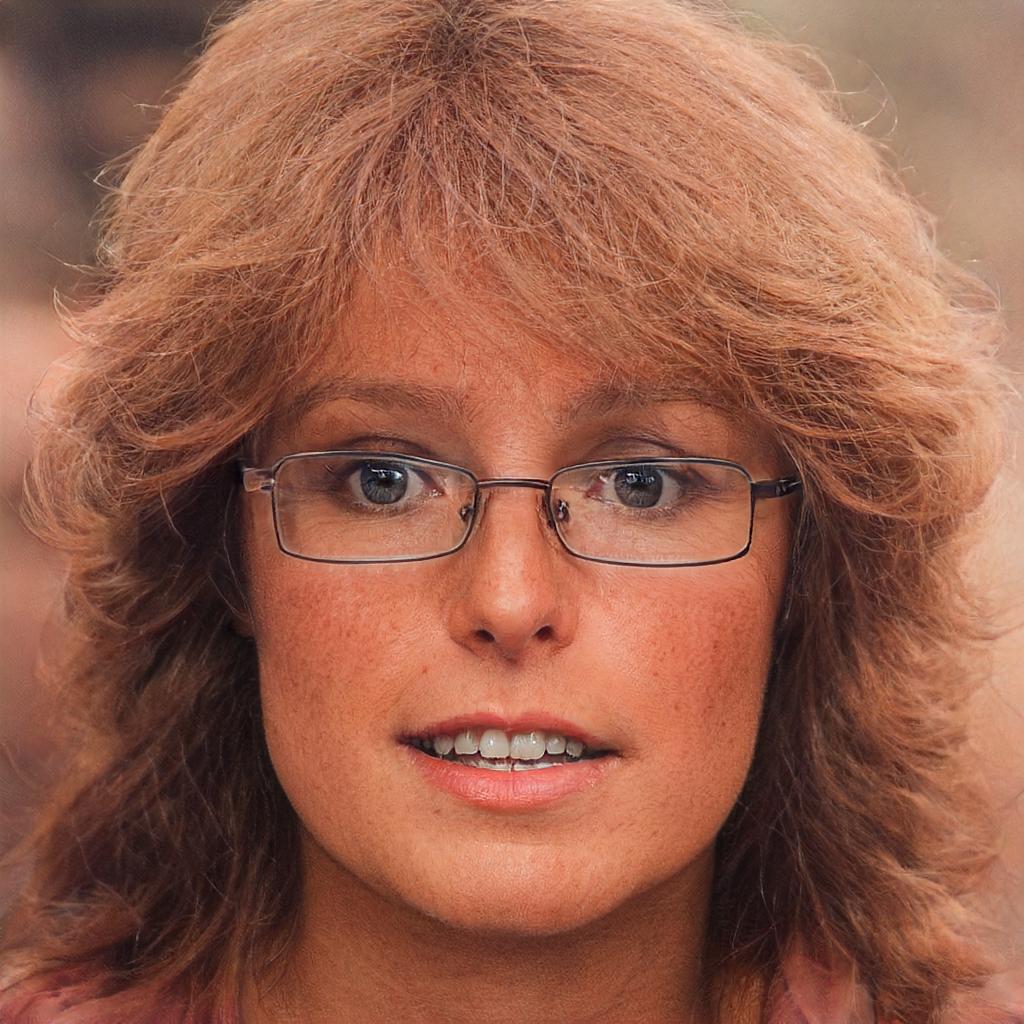}
         \caption{TCA$^2$}
     \end{subfigure}
     \hfill
     \begin{subfigure}[b]{0.155\textwidth}
         \centering
         \includegraphics[width=\textwidth]{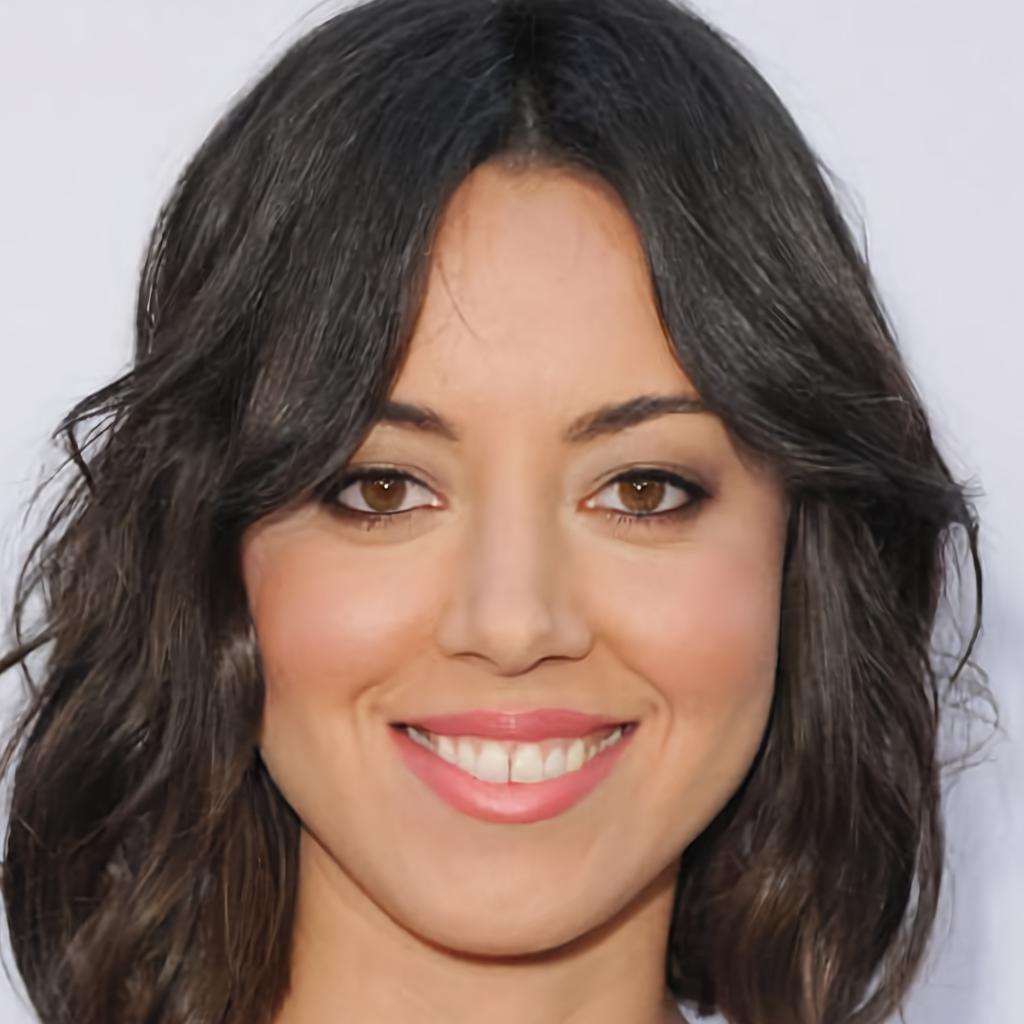}
         \caption{\textcolor{red}{Target}}
     \end{subfigure}
        \caption{The visualization of source face, other different adversarial face and target face. The second image is from GAN inversion. Some representations of norm-based adversarial examples\cite{madry2017towards} and unrestricted adversarial examples\cite{komkov2021advhat} are shown. TCA$^2$ generates the 5$^{th}$ image.}
        \label{fig:abstrct_examples}
\end{figure*}

Recently proposed unrestricted adversarial attack (UAA) methods generate adversarial images with more stealthy and semantically meaningful perturbations compared to noise-based adversarial attacks. Some unrestricted adversarial attacks hide the adversarial perturbations as decorative accessories like glasses \cite{sharif2016accessorize} or hat \cite{komkov2021advhat} to improve stealthiness. Notably, makeup-based adversarial attacks \cite{yin2021adv} generate perturbations as natural makeup. Moreover, \cite{li2021exploring} generate more natural face images than previous methods with the help of pre-trained GANs. Adv-Attribute \cite{jia2022adv} automatically injects a pre-defined pattern from a target image to complete an adversarial edit. Adv-Diffusion \cite{DBLP:conf/aaai/LiuWP0H024} extracts latent codes from both source and target images, then exploits a diffusion model to generate adversarial faces. Although these UAA methods demonstrate improved stealthiness, they have limited ability to change attributes and primarily edit a pre-defined set of semantic information from the target image. We argue that it is essential to rapidly generate adversarial images with specific attributes, such as skin color, expression, or hairstyle, guided by attribute text. Such adversarial attacks can help security researchers expose vulnerabilities in existing FR models due to changes in facial attributes that are likely to occur.

Notwithstanding their effectiveness in attacking face recognition (FR) systems, these adversarial attacks have significant limitations. Although some previous works achieve relative transferability in black-box scenarios, they still struggle to attack FR models in real-world scenarios. Specifically, works such as Adv-Attribute \cite{jia2022adv} and Adv-Diffusion \cite{DBLP:conf/aaai/LiuWP0H024} tend to generate impersonated faces optimized for a specific model, leading to overfitting to the source model. This limitation drives us to develop a more transferable attack that can generalize well to real-world FR systems.

To address the aforementioned shortcomings, this paper proposes a Text Controlled Attribute Attack (TCA$^2$) to generate adversarial perturbations guided by text prompts as shown in \refig{fig:abstrct_examples}. By feeding the targeted image into the FR model, a discriminative category-level softmax vector is produced to guide the impersonation attack. To enhance black-box transferability, we employ both data and model augmentation strategies. For data augmentation, we adopt simple random resizing and padding. For model augmentation, we apply a meta-learning paradigm to simulate white-box and black-box FR environments, further improving transferability. The framework of our TCA$^2$ is shown in \refig{framework}.

 Our contributions can be summarized as follows:
\begin{enumerate}
    \item We propose a novel Text Controlled Attribute Attack (TCA$^2$) to generate semantically meaningful perturbations guided by text prompts. Significantly, unlike existing unrestricted adversarial attacks (UAA), our TCA$^2$ offers rich face attribute editing capabilities under text guidance;
    \item Both data and model augmentation techniques are employed to generate adversarial images that are more transferable to unknown black-box face recognition (FR) models;
    \item Extensive experiments validate the superior effectiveness and transferability of our method compared to other state-of-the-art attack techniques on two high-resolution datasets.
\end{enumerate}

\section{Related Work}

\subsection{Norm-based Adversarial Examples}
Many adversarial attack algorithms \cite{DBLP:journals/istr/RyuPC21} have demonstrated that deep learning face recognition (FR) models are vulnerable to adversarial samples. Traditional adversarial examples against FR focus on norm-constrained conditions. For a given FR model $\mathcal{F}(x): \mathcal{X} \rightarrow \mathbb{R}^{d}$ and a face image $x \in \mathbb{R}^{n}$, the adversarial image $\hat{x} \in \mathbb{R}^{n}$ satisfies the condition $\left| x - \hat{x} \right|p < \epsilon$ and $\mathcal{F}(x) \ne \mathcal{F}(\hat{x})$. Common values for $p$ are 0, 2, and $\infty$, and $\epsilon$ is a sufficiently small value to ensure the perturbation is imperceptible. Adversarial attacks against FR models can be categorized as impersonation (targeted) attacks and dodging (untargeted) attacks based on whether their goal is to make the FR classify the adversarial face image as a specified $\hat{y}$ or any $\hat{y} \ne y$. Representative norm-based methods include the Fast Gradient Sign Method (FGSM) \cite{goodfellow2014explaining}, which uses a first-order approximation of the function for faster adversarial example generation, and Projected Gradient Descent (PGD) \cite{madry2017towards}, an iterative variant of FGSM that provides a strong first-order attack through multiple steps of gradient ascent. Carlini and Wagner (C\&W) \cite{carlini2017towards} proposed stronger optimization-based attacks for $L_0$, $L_2$, and $L\infty$ via improved objective functions. AdvGAN \cite{xiao2019generating} proposes a GAN network to efficiently generate adversarial examples. These methods can easily fool victim neural networks. However, norm-based adversarial examples can still be detected by humans or adversarial detectors \cite{massoli2021detection}. Consequently, several defense mechanisms against such attacks have been proposed, such as adversarial training \cite{madry2017towards}.

\begin{figure*}[!t]
\centerline{\includegraphics[width=32pc]{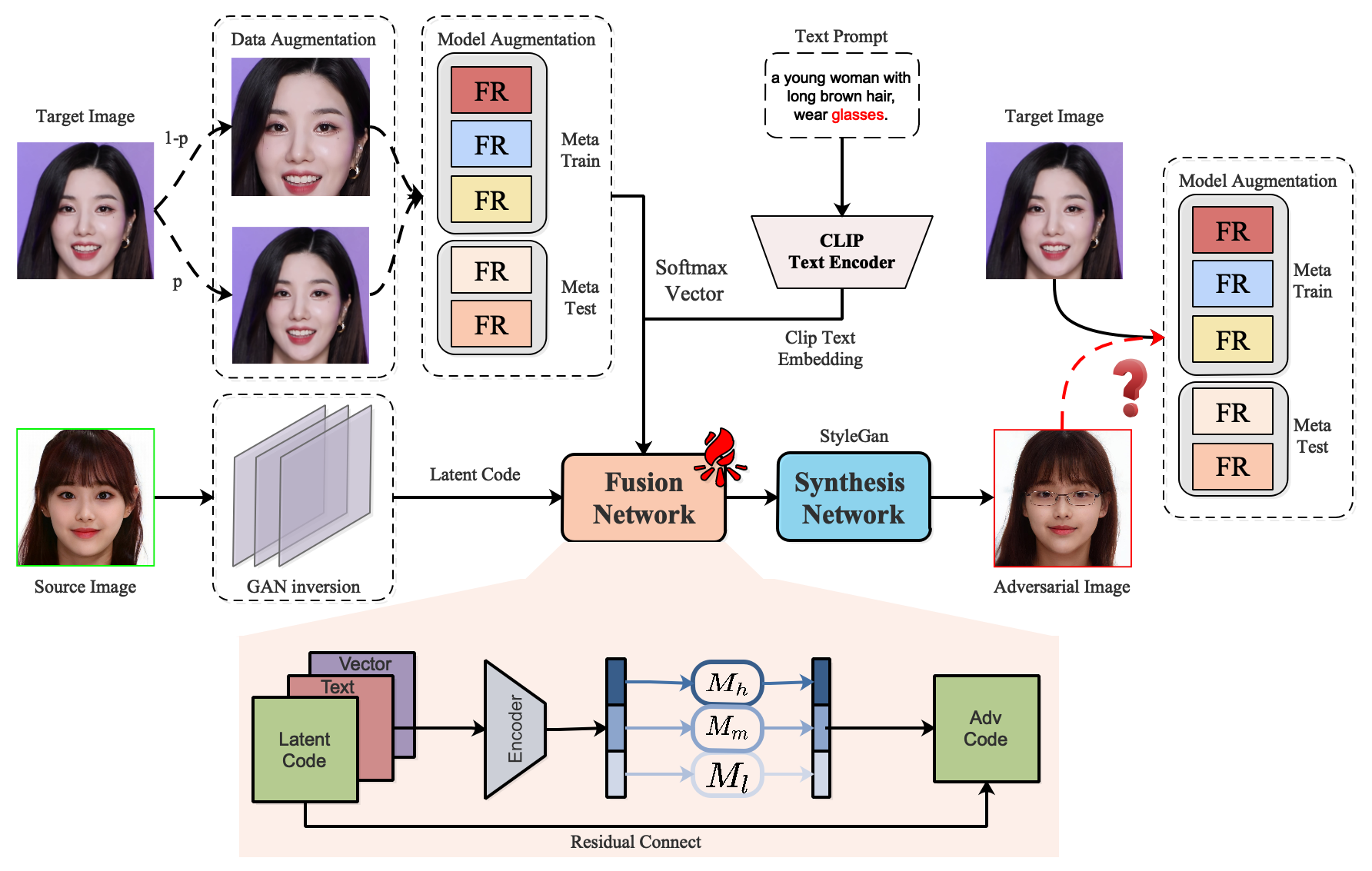}}
\caption{The overall framework of our proposed Text Controlled Attribute Attack (TCA$^2$). Our proposed method involves searching for adversarial faces on the generative StyleGAN manifold by optimizing the parameters of a Fusion Network. The generated photo-realistic adversarial faces can deceive state-of-the-art face recognition (FR) systems under the guidance of text prompts. To effectively represent the semantics of an impersonated person, a softmax vector is employed to perform a targeted attack against the FR system. The FR models in the framework are randomly selected from either the meta-train set or the meta-test set.
}
\label{framework}
\end{figure*}
\subsection{Unrestricted Adversarial Attack}
Traditional adversarial perturbations are constrained by norm bounds, whereas unrestricted adversarial attacks (UAA) are not subject to such limitations. These attacks have been extensively studied in image classification tasks \cite {sharif2016accessorize,brown2017adversarial,karmon2018lavan}. UAA generates adversarial images with semantically meaningful perturbations compared to noise-based adversarial attacks. Some UAAs have been proposed by generating adversarial wearable accessories like glasses \cite{sharif2016accessorize} or hat\cite{komkov2021advhat} to fool the FR model. However, such colorful patches are easily noticeable, leading to poor stealthiness. Makeup-based UAA \cite{yin2021adv,guetta2021dodging} are developed against FR models by generating perturbations in the form of makeup, but these generated makeups still appear unnatural to humans. More recently, some attacks \cite{li2021exploring} have been introduced using pre-trained generative models. These works demonstrate an excellent ability to generate adversarial examples containing fewer artifacts compared to previous UAA.

\subsection{Transferable Attack}
Adversarial transferability refers to the ability of adversarial examples, generated on a white-box model (the source model), to successfully deceive a black-box model (the target model). Black-box adversarial attacks, which follow the most common practice, are more feasible in real-world scenarios and hold greater research significance compared to white-box attacks. While most adversarial attacks are designed for white-box models \cite{madry2017towards}, they exhibit poor transferability when applied directly to black-box models. To address this limitation, various black-box attack strategies have been proposed, including query-based and transfer attacks. Query-based attacks \cite{ilyas2018black} estimate gradients through a large number of queries to the target model. However, these methods require an extensive number of queries, making them susceptible to detection by the target system. In contrast, transfer attacks are more efficient. Although some previous works have achieved relatively high transferability in black-box settings, they often overfit to the source model. In our work, we employ both data and model augmentation techniques to enhance the transferability of our TCA$^2$  method in black-box scenarios.

\section{Method}
\subsection{Problem Formulation}
\label{Problem_Formulation}
The objective of adversarial face attacks is to deceive the target face recognition (FR) model using adversarial perturbations. Specifically, an impersonation attack seeks to cause the FR model to misclassify a face as another specific identity by introducing subtle perturbations. Most prior studies have learned these perturbations under norm constraints to ensure stealthiness. In our work, we relax these strict constraints to explore unconstrained adversarial attacks (UAA).

Let $x_s \in \mathcal{X} \subset  \mathbb{R} ^{n}$ denote the given source face image, and let $x_t \in \mathcal{X} \subset  \mathbb{R} ^{n}$ denote the target face image to be impersonated. Let $\mathcal{F} (x):\mathcal{X} \rightarrow \mathbb{R} ^{d}$ be a face recognition model that extracts the normalized facial feature representation for identification. The optimization process of the impersonation face attack can be expressed as follows:
\begin{align}
\label{targeted}
\underset{\hat{x}_s }{arg min} & \texttt{S}(\mathcal{F}(\hat{x}_s),\mathcal{F}(x_t)) \\
\underset{\hat{x}_s }{arg min} & \mathcal{D}(\hat{x}_s,x_s)
\end{align}
where $\texttt{S}(\cdot)$ denotes to the identity similarity, $\mathcal{D}$ represents the perceptual distance used in unrestricted adversarial attacks. We adopt the most common perceptual network LPIPS\cite{zhang2018unreasonable} to measure the difference between the clean face image $x_s$ and its corresponding adversarial image $\hat{x}_s$. A natural language text prompt $t$ is adopted to control the generation of $\hat{x}_s$ according to the intent of adversary.

The objective of our approach is to generate a high-quality adversarial face, denoted as $\hat{x}_s$, which closely resembles its original image except for the attribute controlled by the text prompt $t$. Additionally, $\hat{x}_s$ is designed to effectively mislead the black-box face recognition system, causing it to misidentify $\hat{x}_s$ as $x_t$.

\subsection{Preliminaries}
A latent code in the style space of StyleGAN \cite{karras2020analyzing} can be projected into a specific image. Following the approach of \cite{li2021exploring}, we manipulate in the latent space of StyleGAN directly. Let $G_L$ denote the generator network with $L$ layers in StyleGAN. The random noise $z$ is sampled from a uniform distribution $Z$ and then transformed into a style vector $\omega$ via a nonlinear mapping network $f$. The intermediate latent code $\omega$ consists of $L$ copies, i.e., $\omega = \left[\omega_1, \omega_2, \cdots, \omega_L\right] \in \mathbb{R}^{L \times 512}$. Each $\omega_m$ within $\omega$ represents the latent code input to the $L_m$ layer of $G_{L_m}$. This $\omega_m$ is projected into the $L_m$ layers and controls the $m^{th}$ level of style in the synthesized image, where $m \in {1, 2, \cdots, L}$. The corresponding attribute at the $m^{th}$ level varies with changes in the value of $\omega_m$. It is important to note that $\omega_m$ at different depths influences generated attributes to varying degrees: shallow layers control coarse attributes, middle layers control intermediate attributes, and deep layers control fine attributes. This impact is further illustrated in the Supplementary Materials.

In addition to the latent code $\omega$, a noise term $\eta$, also sampled from the uniform distribution $Z$, is introduced to control the stochastic variations of the generated image at each layer. The noise term $\eta$ typically affects uncorrelated attributes, such as the fine details of hair strands in a generated face. Since $\omega$ is entangled with semantically meaningful attributes, this work aims to control $\omega$ with a text prompt $t$ to generate the desired adversarial image capable of fooling the target FR. 

\subsection{Text Controlled Attribute Attack}
Previous works \cite{jia2022adv} have introduced semantically meaningful perturbations to create transferable adversarial examples against face recognition (FR) systems by injecting specific styles or patterns from a target image. However, these methods face two significant limitations: 1) \textbf{Inability to control the adversarial attributes.} While a real attribute vector corresponding to specific styles, such as smiling or glasses, is provided to control the generation details, the process automatically injects a pre-defined pattern from the target image. This means that the attacker cannot control the type of injected attribute, nor can they introduce a pattern outside the predefined attribute candidates. This limitation severely restricts the adversary's ability to generate the desired adversarial face. 2) \textbf{Low adversarial transferablity.} The semantically meaningful adversarial perturbations \cite{jia2022adv, qiu2020semanticadv, DBLP:conf/aaai/LiuWP0H024} are optimized based on a single target FR model. As a result, the generated adversarial examples are highly coupled with the white-box FR model, which significantly reduces their transferability when used to attack black-box FR models with different architectures and parameters. Our work addresses these two challenges by focusing on enhancing both the control over adversarial attributes and the transferability of the generated adversarial faces.

\subsubsection{Text-controlled Adversarial Face Generation}
Our text-controlled adversarial face generator (illustrated in \refig{framework}) leverages the robust joint multimodal representation capabilities of the vision-language pretrained model, specifically CLIP \cite{radford2021learning}. Given a text prompt $t$, the CLIP textual encoder $CLIP_t$ projects it into a shared embedding space as $E_t = CLIP_t(t)$, where $E_t$ represents the textual embedding of the prompt $t$. For a clean face image $x_s$, a StyleGAN inverter network $Inv(\cdot)$ converts it into the corresponding style latent code, denoted as $\omega_s = Inv(x_s)$. To increase diversity, we apply random resizing and padding operations as data augmentation to the target face image. The augmented target face is then fed into FR to generate a softmax vector $v$, which guides the generation of the adversarial face image. Subsequently, the textual embedding $E_t$, latent code $\omega_s$, and target face representation $v$ are concatenated and fused. This process can be formalized as follows:
\begin{equation}
    \omega_s^{*}=M_{\Theta_M}([\omega_s,E_t,v])
\end{equation}
where $\mathcal{M}_{\Theta_M}$ is a Multi-Level Fusion Network with learnable parameter $\Theta_M$. Then the adversarial image is generated by $\hat{x}_s=G_L(\omega_s^{*} )$.

\begin{figure*}[!t]
     \centering
     \begin{subfigure}[b]{0.14\textwidth}
         \centering
         \caption{\textcolor{DarkGreen}{Source}}
         \includegraphics[width=\textwidth]{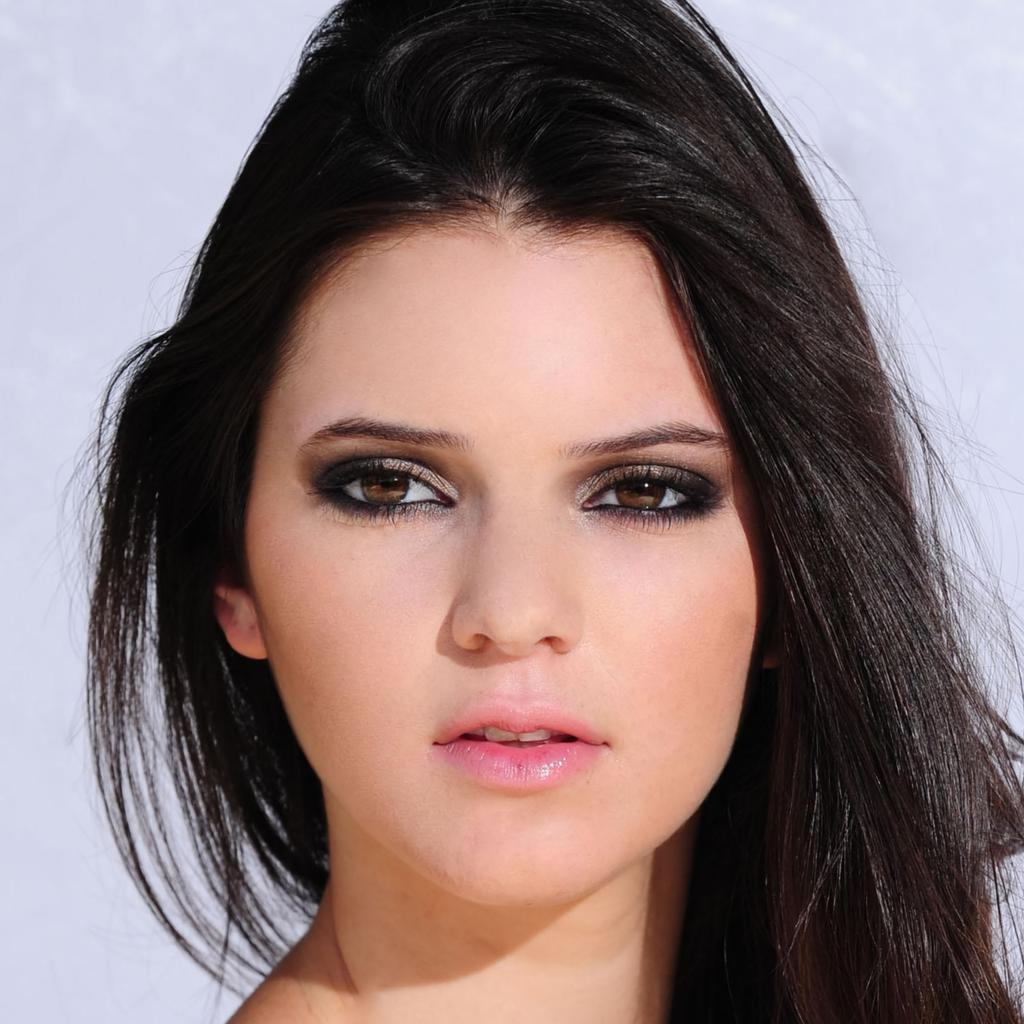}
     \end{subfigure}
     \hspace{.08in}
     \begin{subfigure}[b]{0.14\textwidth}
         \centering
         \caption{MI-FGSM}
         \includegraphics[width=\textwidth]{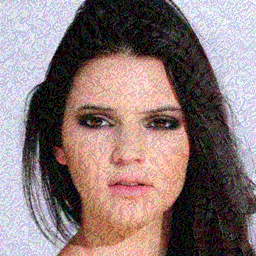}
     \end{subfigure}
     \hspace{.08in}
     \begin{subfigure}[b]{0.14\textwidth}
         \centering
         \caption{PGD}
         \includegraphics[width=\textwidth]{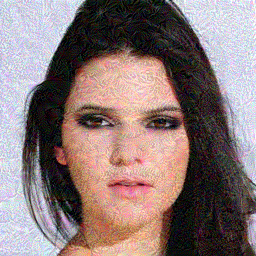}
     \end{subfigure}
     \hspace{.08in}
     \begin{subfigure}[b]{0.14\textwidth}
         \centering
         \caption{AdvHat}
         \includegraphics[width=\textwidth]{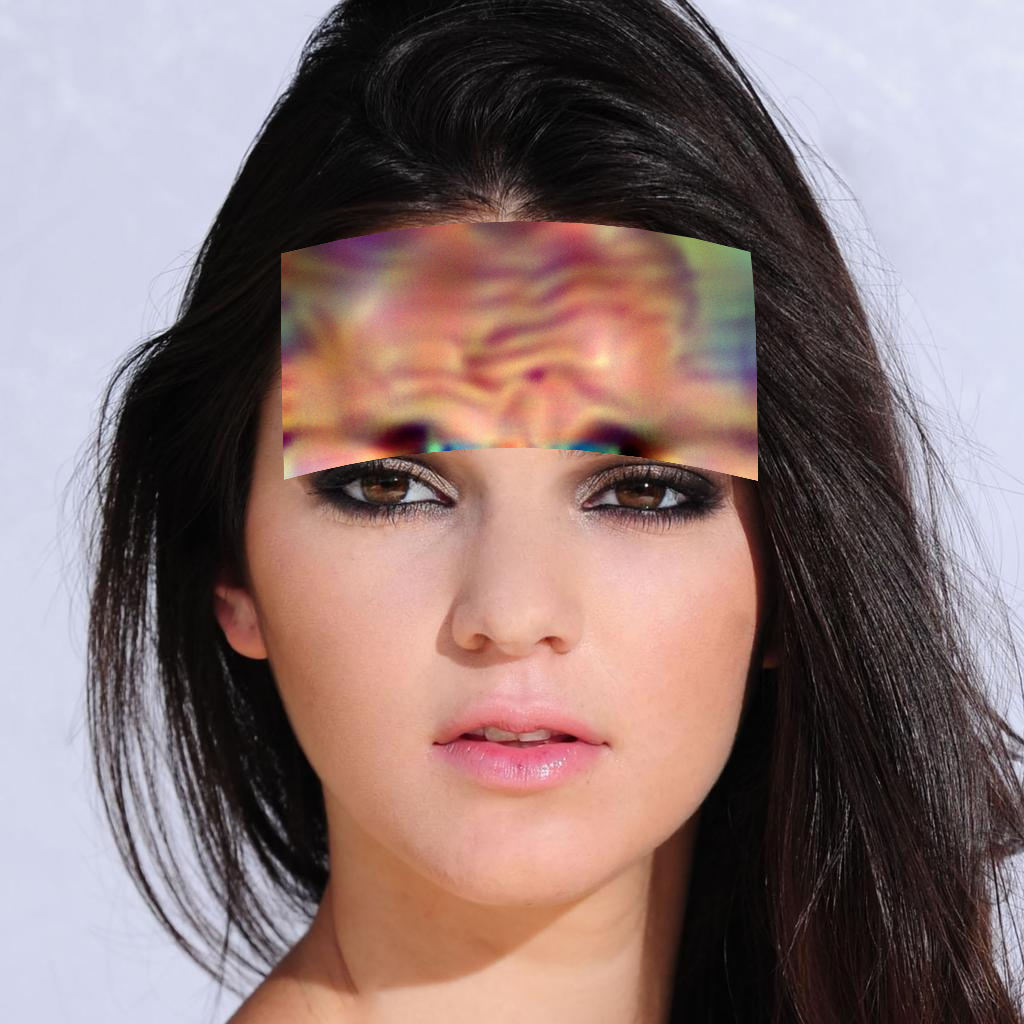}
     \end{subfigure}
     \hspace{.08in}
     \begin{subfigure}[b]{0.14\textwidth}
         \centering
         \caption{TCA$^2$}
         \includegraphics[width=\textwidth]{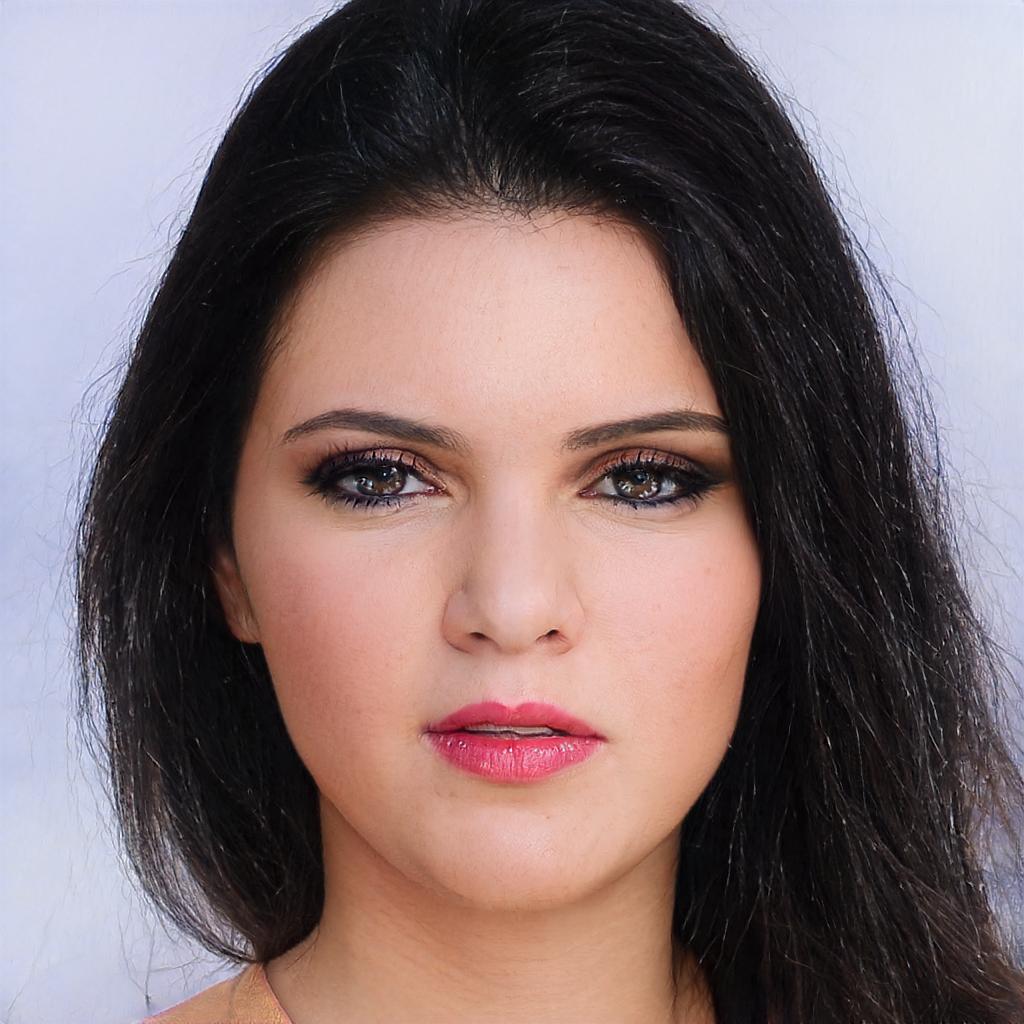}
     \end{subfigure} 
     \hspace{.08in}
     \begin{subfigure}[b]{0.14\textwidth}
         \centering
         \caption{\textcolor{red}{Target}}
         \includegraphics[width=\textwidth]{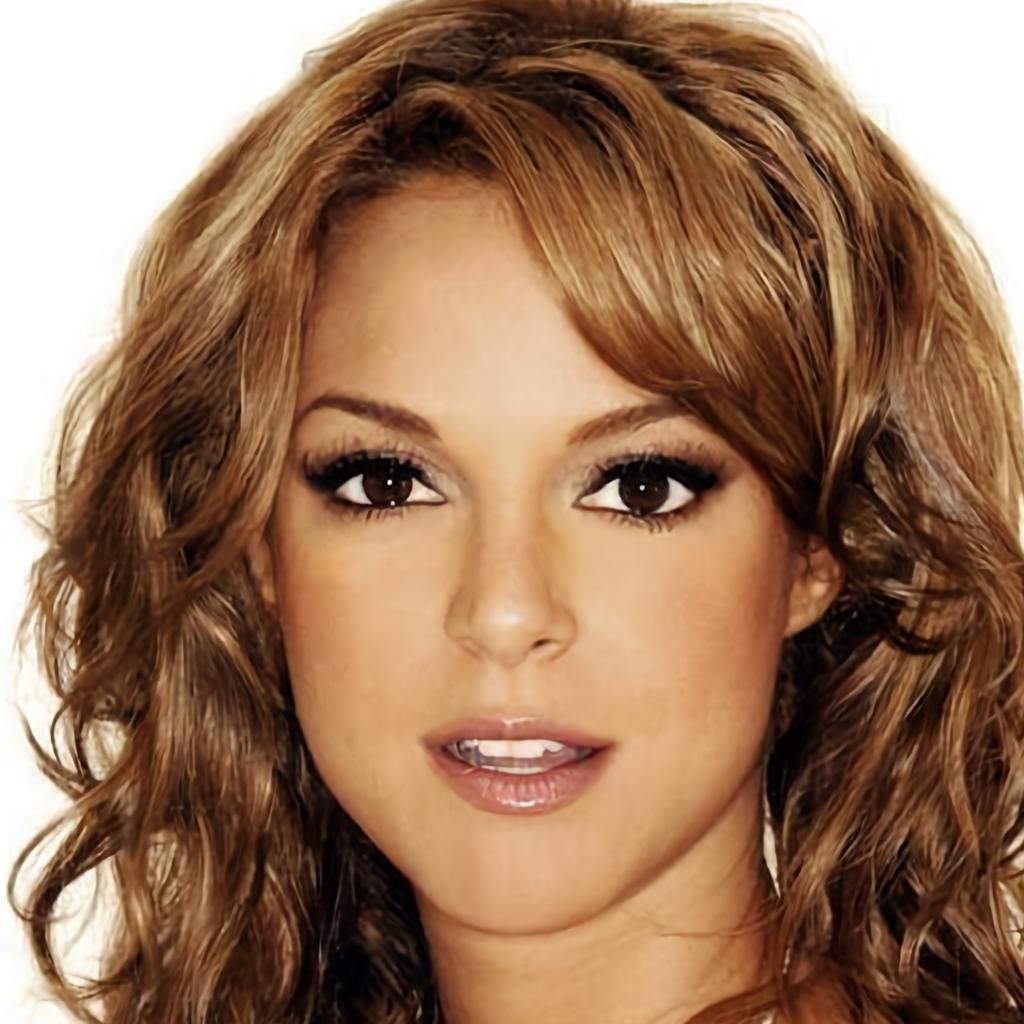}
     \end{subfigure}
     \\
     \begin{subfigure}[b]{0.14\textwidth}
         \centering
         \includegraphics[width=\textwidth]{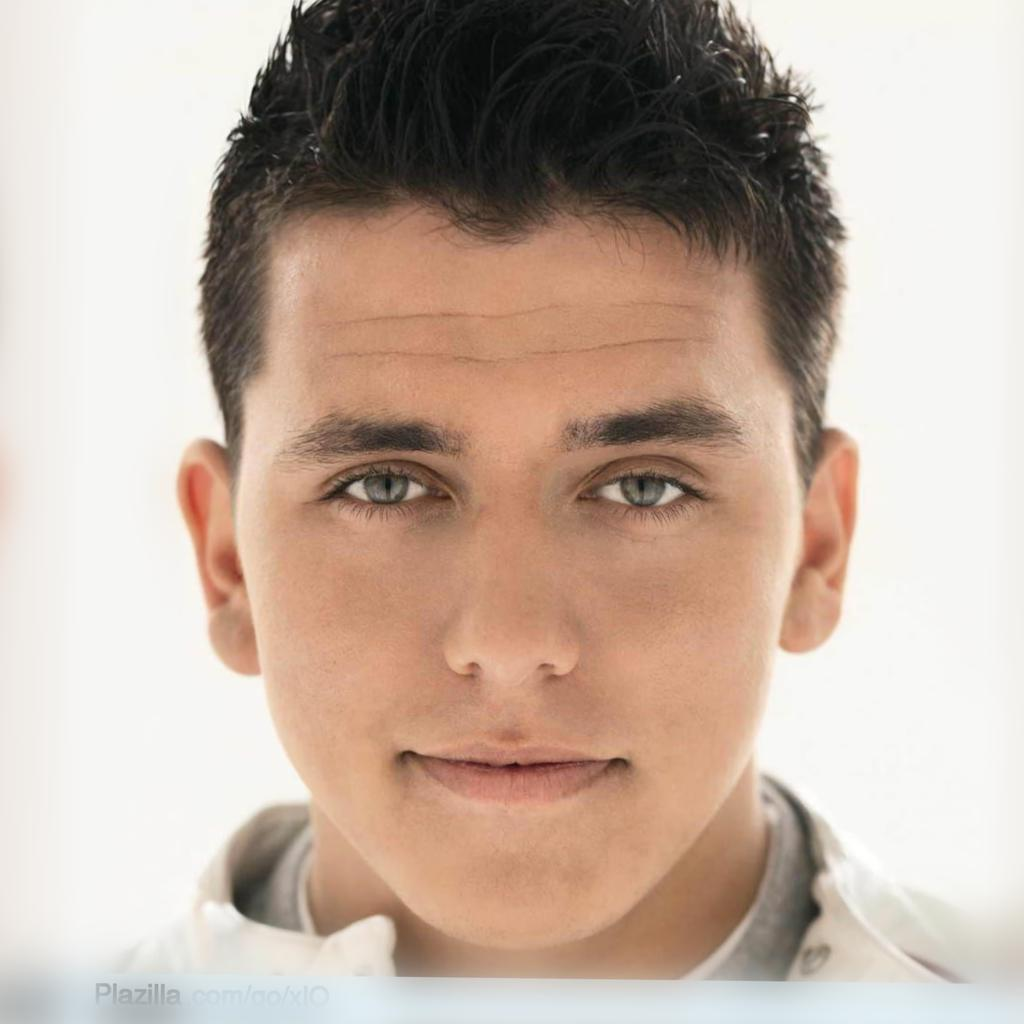}
     \end{subfigure}
     \hspace{.08in}
     \begin{subfigure}[b]{0.14\textwidth}
         \centering
         \includegraphics[width=\textwidth]{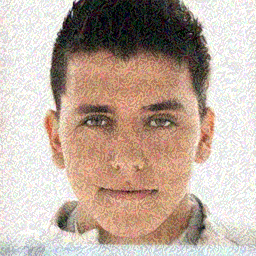}
     \end{subfigure}
     \hspace{.08in}
     \begin{subfigure}[b]{0.14\textwidth}
         \centering
         \includegraphics[width=\textwidth]{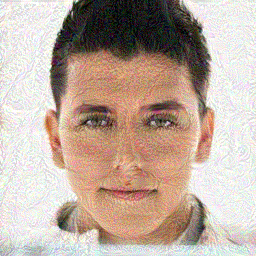}
     \end{subfigure}
     \hspace{.08in}
     \begin{subfigure}[b]{0.14\textwidth}
         \centering
         \includegraphics[width=\textwidth]{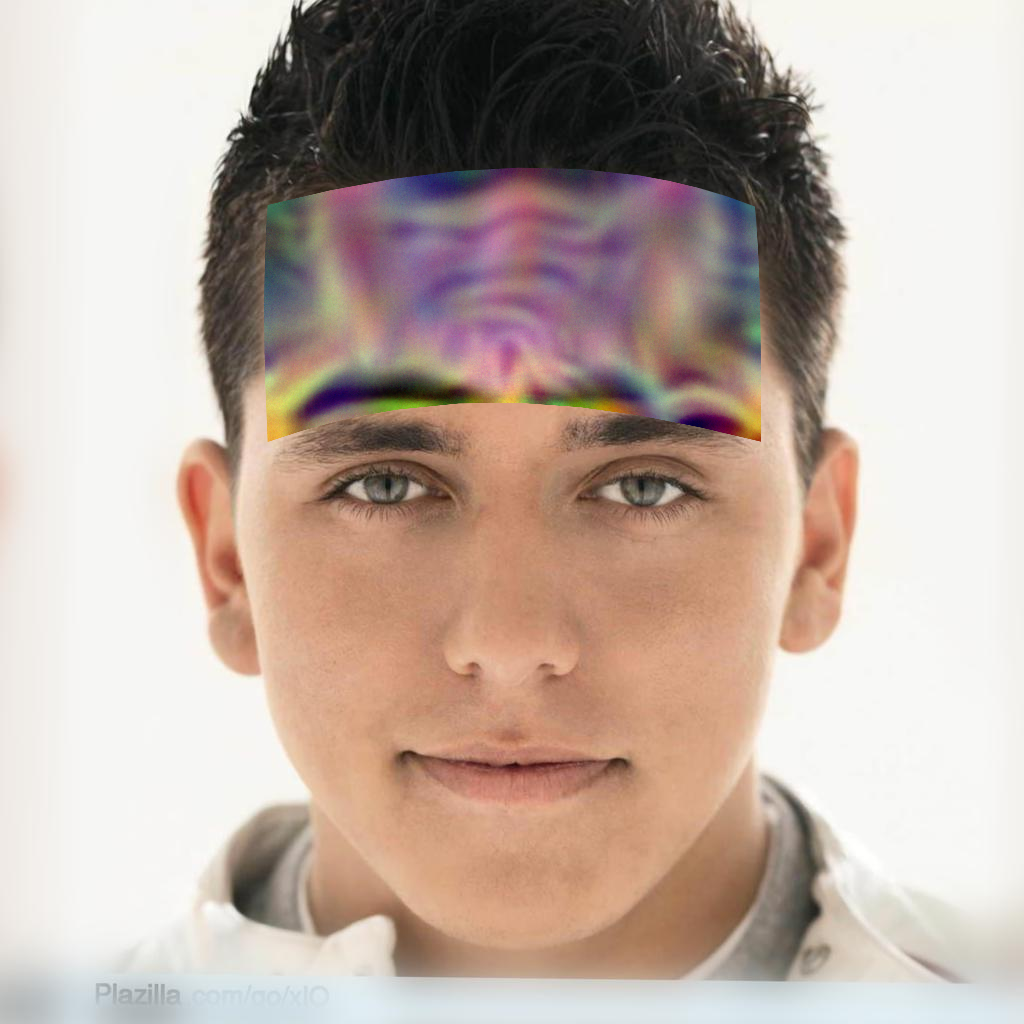}
     \end{subfigure}
     \hspace{.08in}
     \begin{subfigure}[b]{0.14\textwidth}
         \centering
         \includegraphics[width=\textwidth]{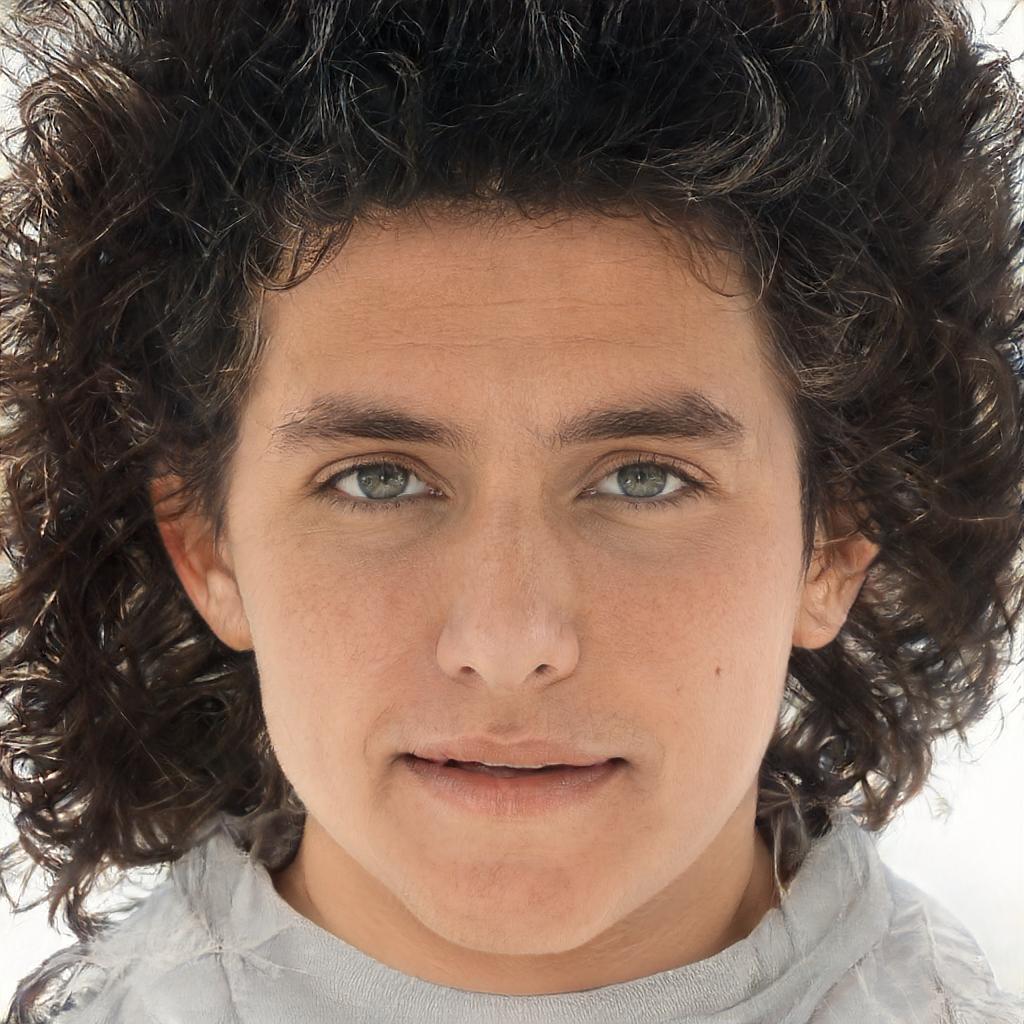}
     \end{subfigure}
     \hspace{.08in}
     \begin{subfigure}[b]{0.14\textwidth}
         \centering
         \includegraphics[width=\textwidth]{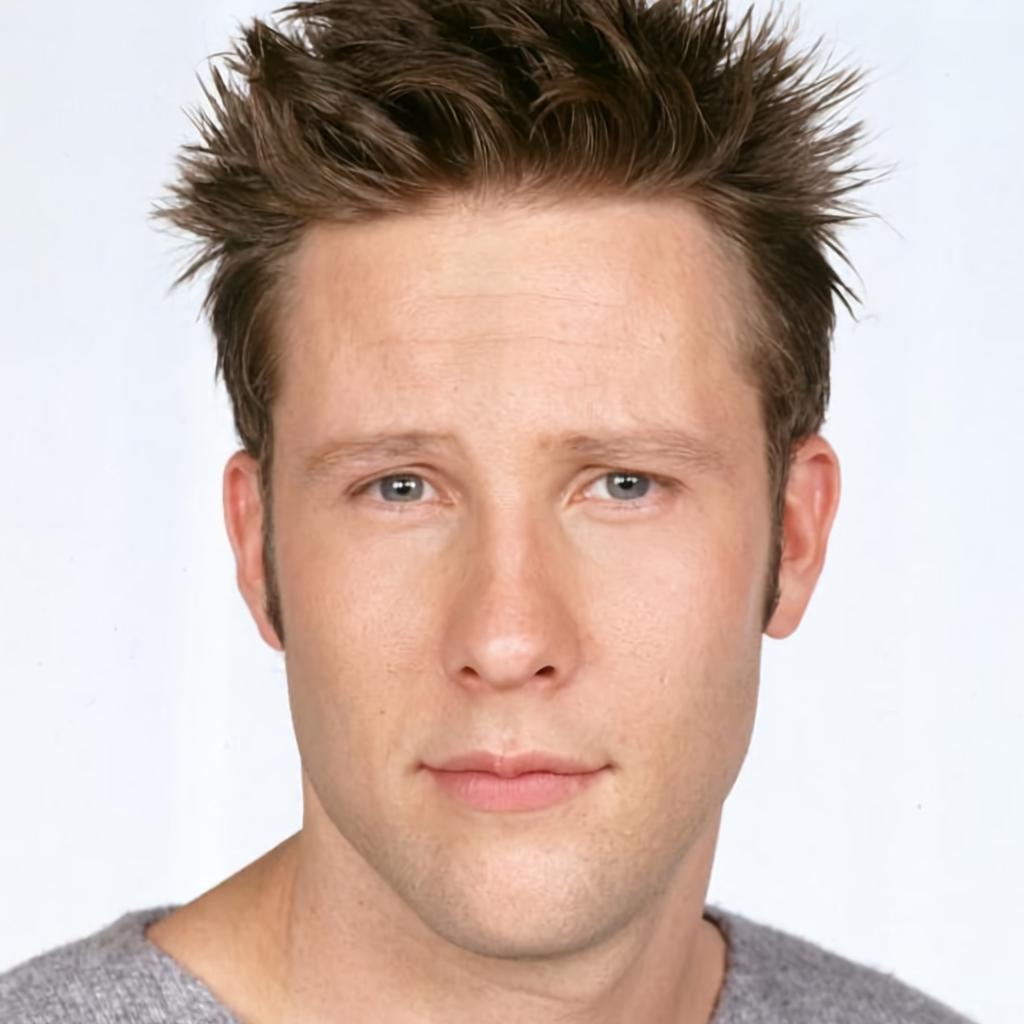}
     \end{subfigure}
     \\
     \begin{subfigure}[b]{0.14\textwidth}
         \centering
         \includegraphics[width=\textwidth]{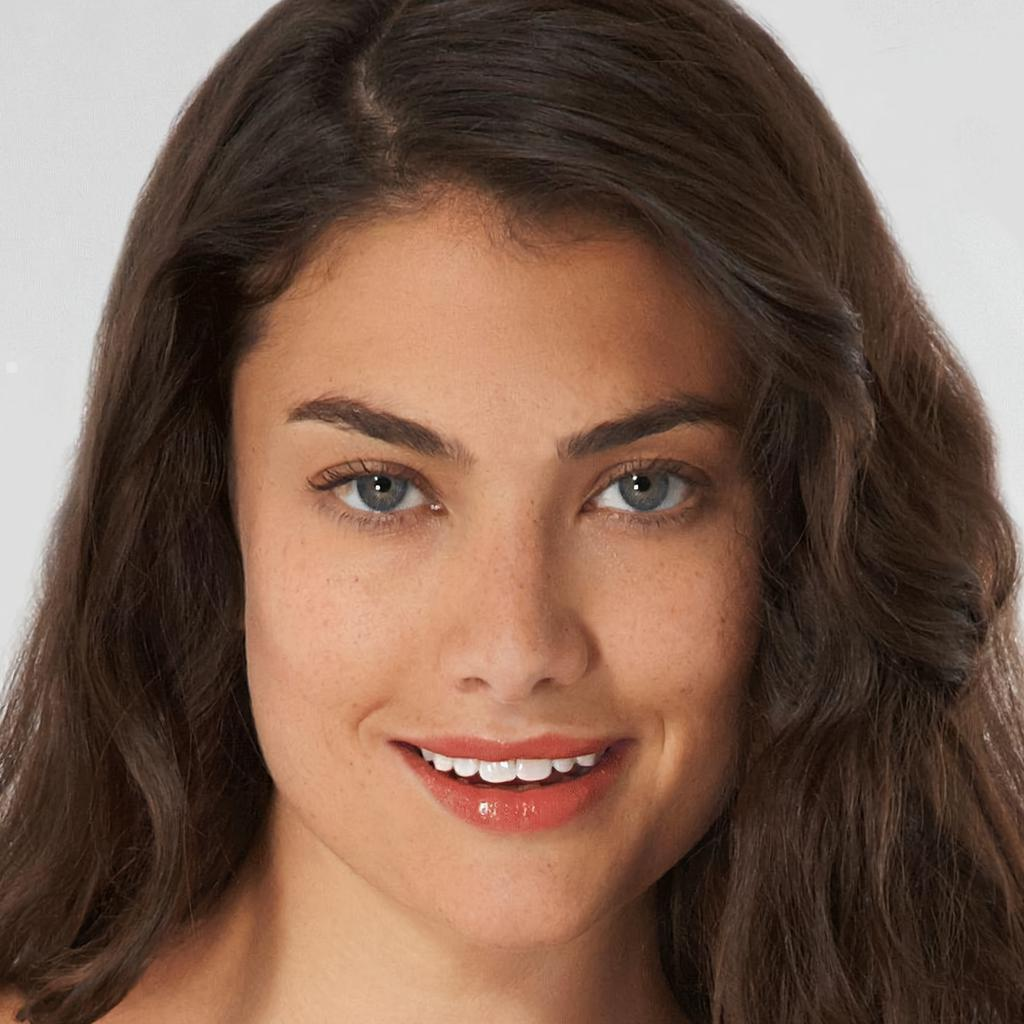}
     \end{subfigure}
     \hspace{.08in}
     \begin{subfigure}[b]{0.14\textwidth}
         \centering
         \includegraphics[width=\textwidth]{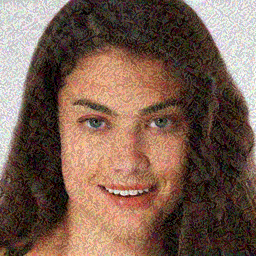}
     \end{subfigure}
     \hspace{.08in}
     \begin{subfigure}[b]{0.14\textwidth}
         \centering
         \includegraphics[width=\textwidth]{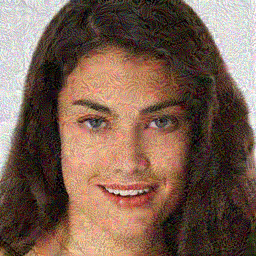}
     \end{subfigure}
     \hspace{.08in}
     \begin{subfigure}[b]{0.14\textwidth}
         \centering
         \includegraphics[width=\textwidth]{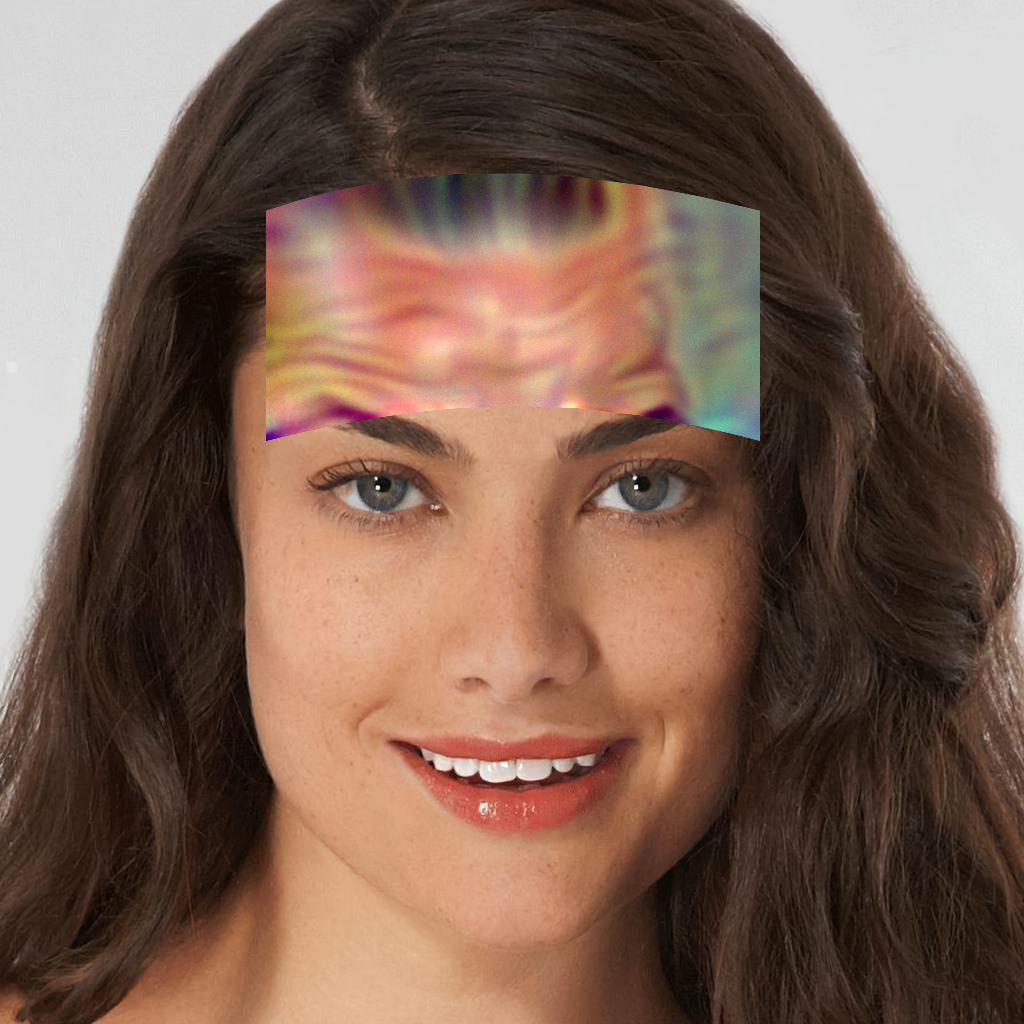}
     \end{subfigure}
     \hspace{.08in}
     \begin{subfigure}[b]{0.14\textwidth}
         \centering
         \includegraphics[width=\textwidth]{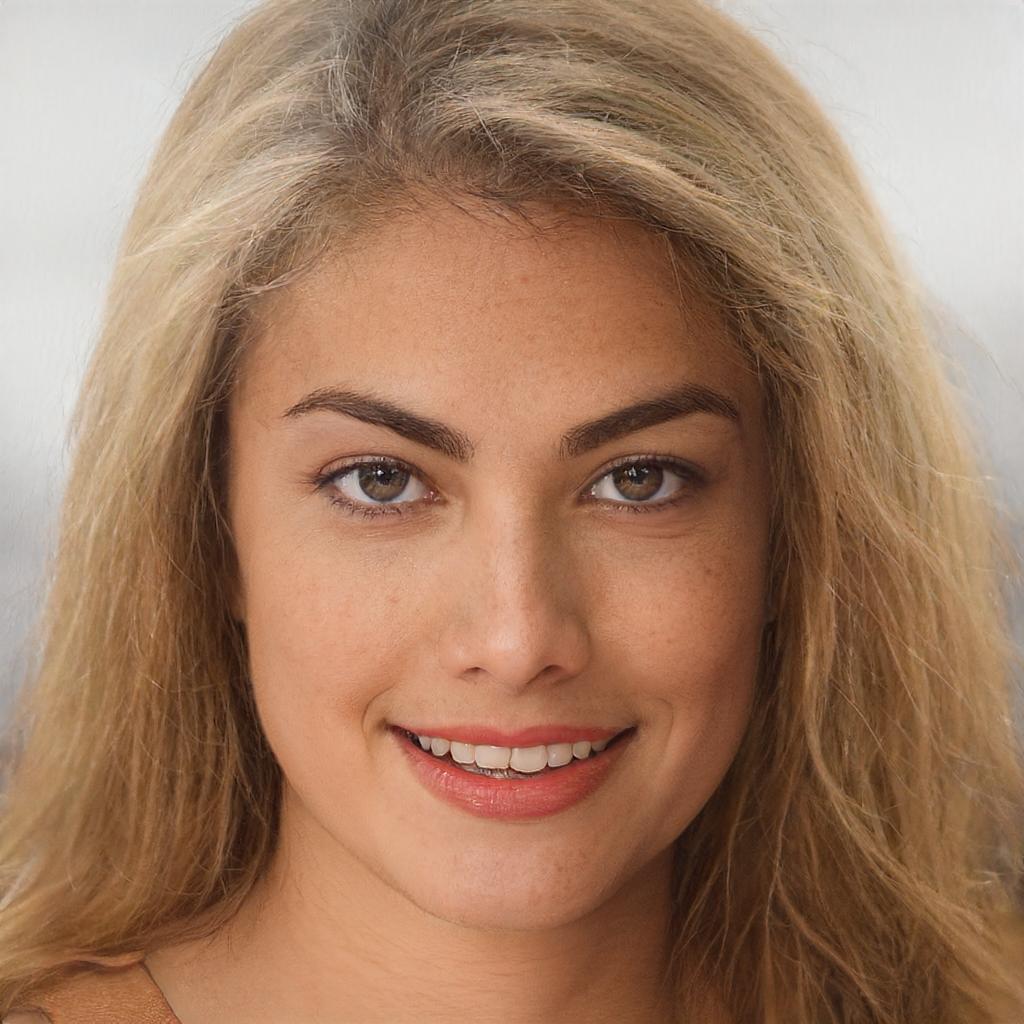}
         
     \end{subfigure}
     \hspace{.08in}
     \begin{subfigure}[b]{0.14\textwidth}
         \centering
         \includegraphics[width=\textwidth]{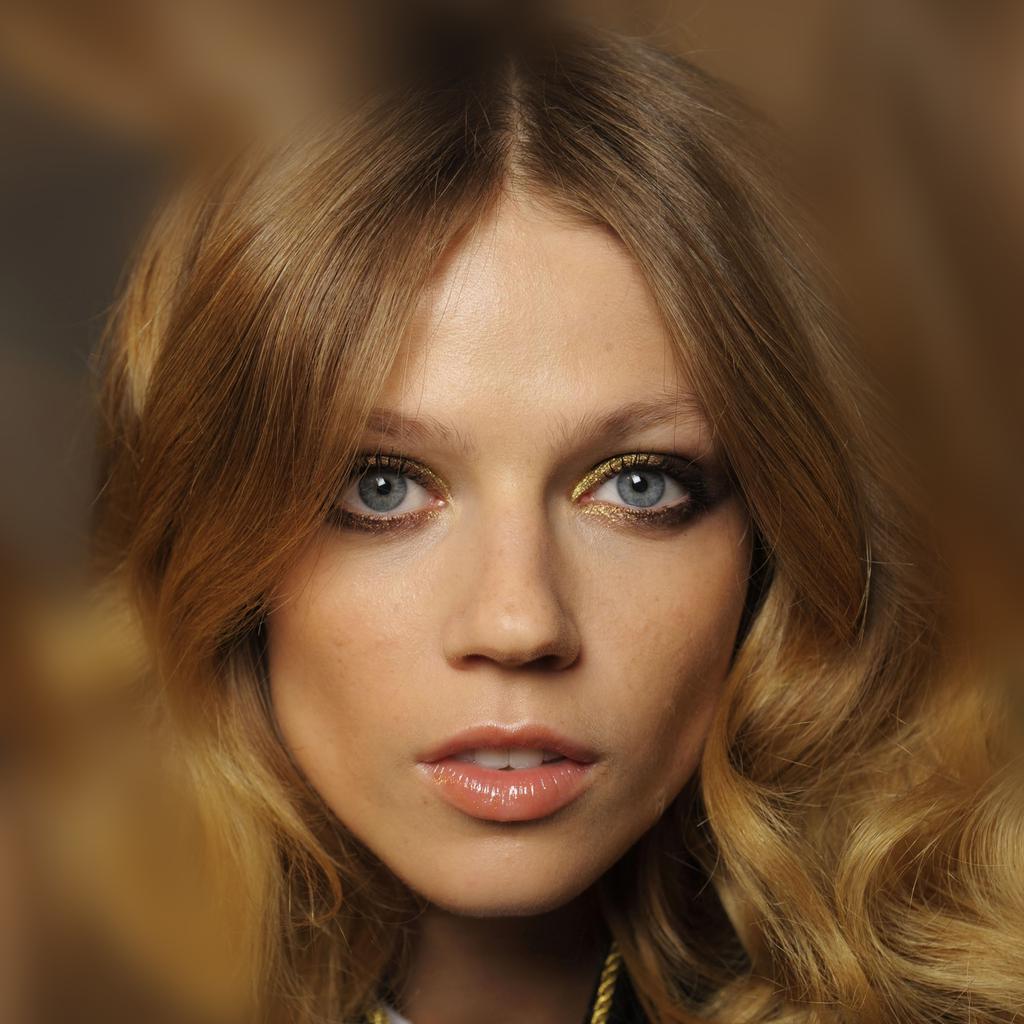}
        
     \end{subfigure}
     \\
     \begin{subfigure}[b]{0.14\textwidth}
         \centering
         \includegraphics[width=\textwidth]{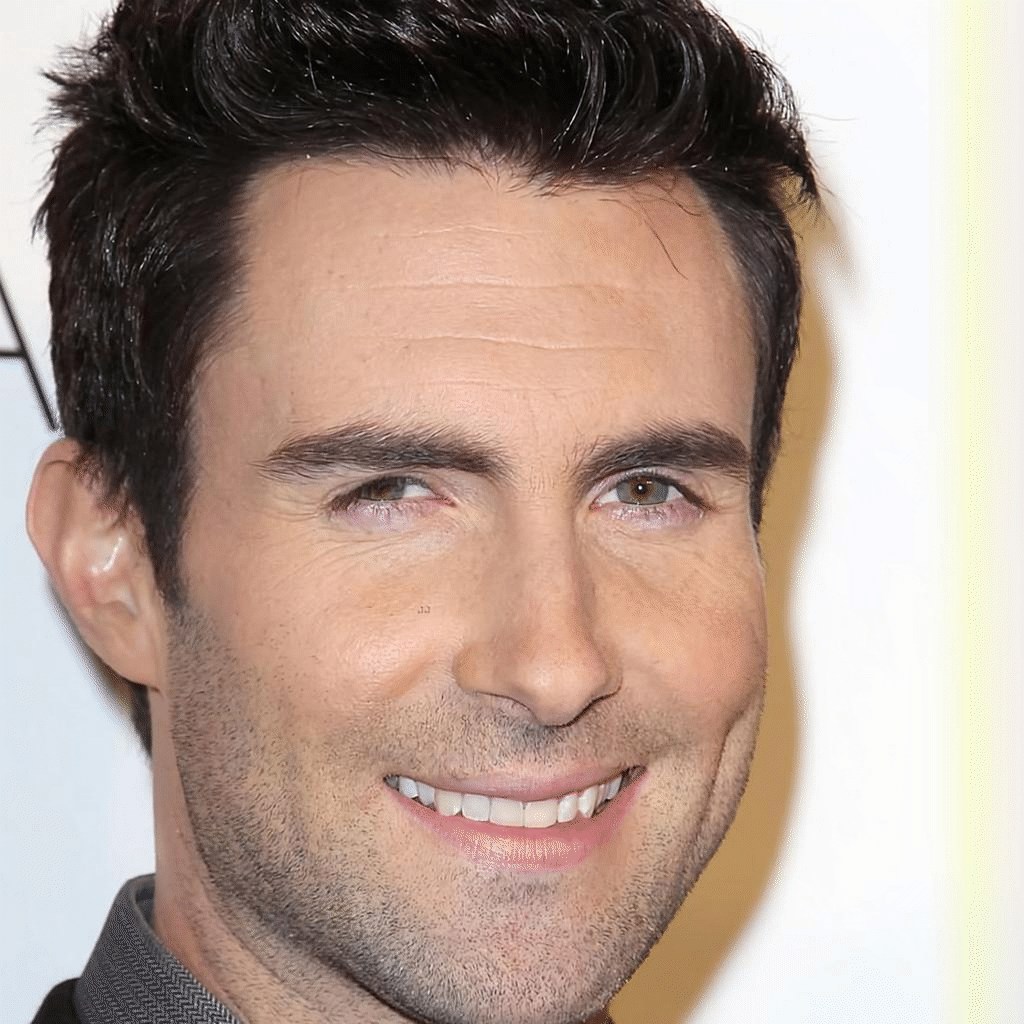}
         
     \end{subfigure}
     \hspace{.08in}
     \begin{subfigure}[b]{0.14\textwidth}
         \centering
         \includegraphics[width=\textwidth]{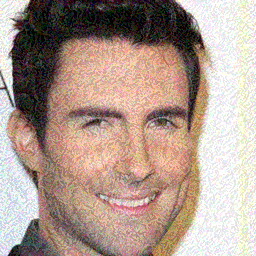}
         
     \end{subfigure}
     \hspace{.08in}
     \begin{subfigure}[b]{0.14\textwidth}
         \centering
         \includegraphics[width=\textwidth]{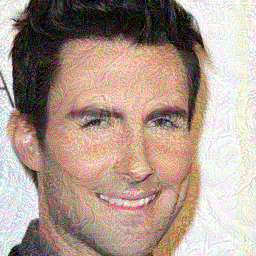}
         
     \end{subfigure}
     \hspace{.08in}
     \begin{subfigure}[b]{0.14\textwidth}
         \centering
         \includegraphics[width=\textwidth]{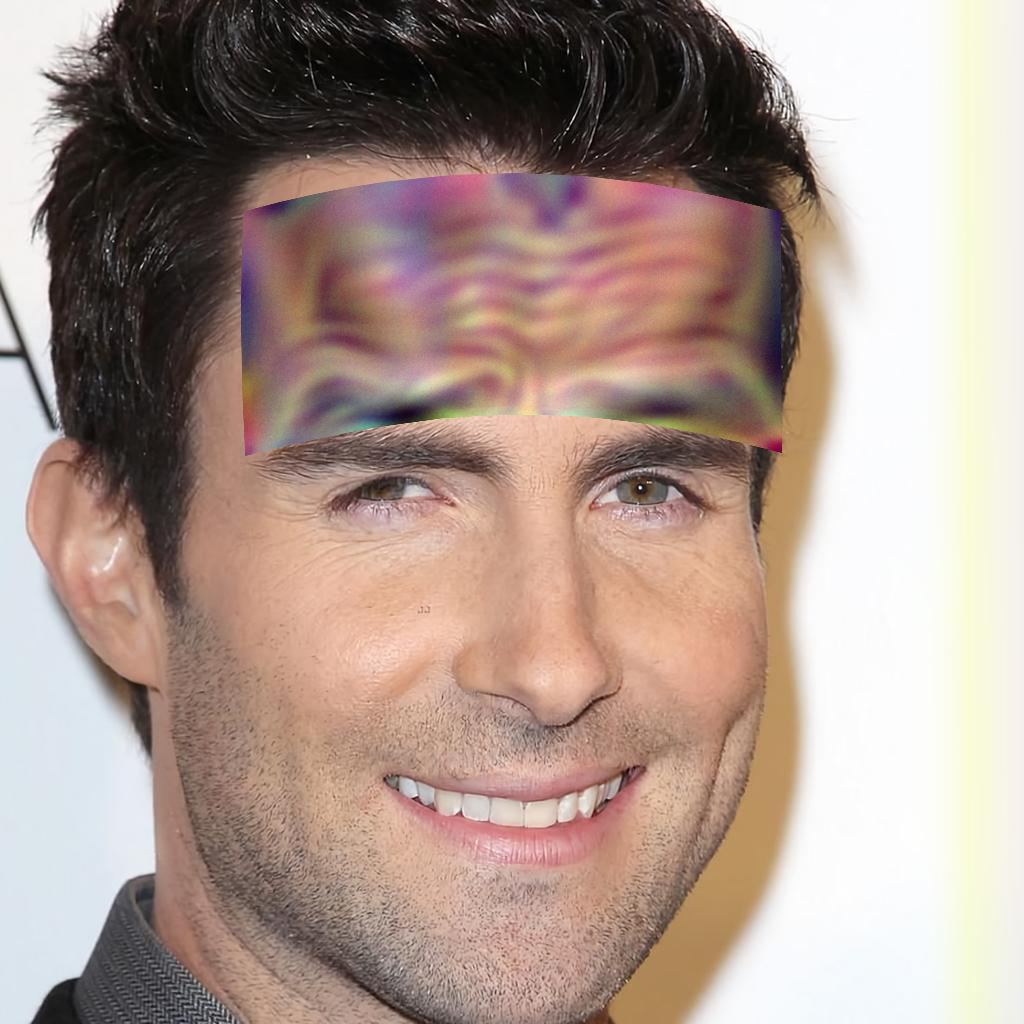}
         
     \end{subfigure}
     \hspace{.08in}
     \begin{subfigure}[b]{0.14\textwidth}
         \centering
         \includegraphics[width=\textwidth]{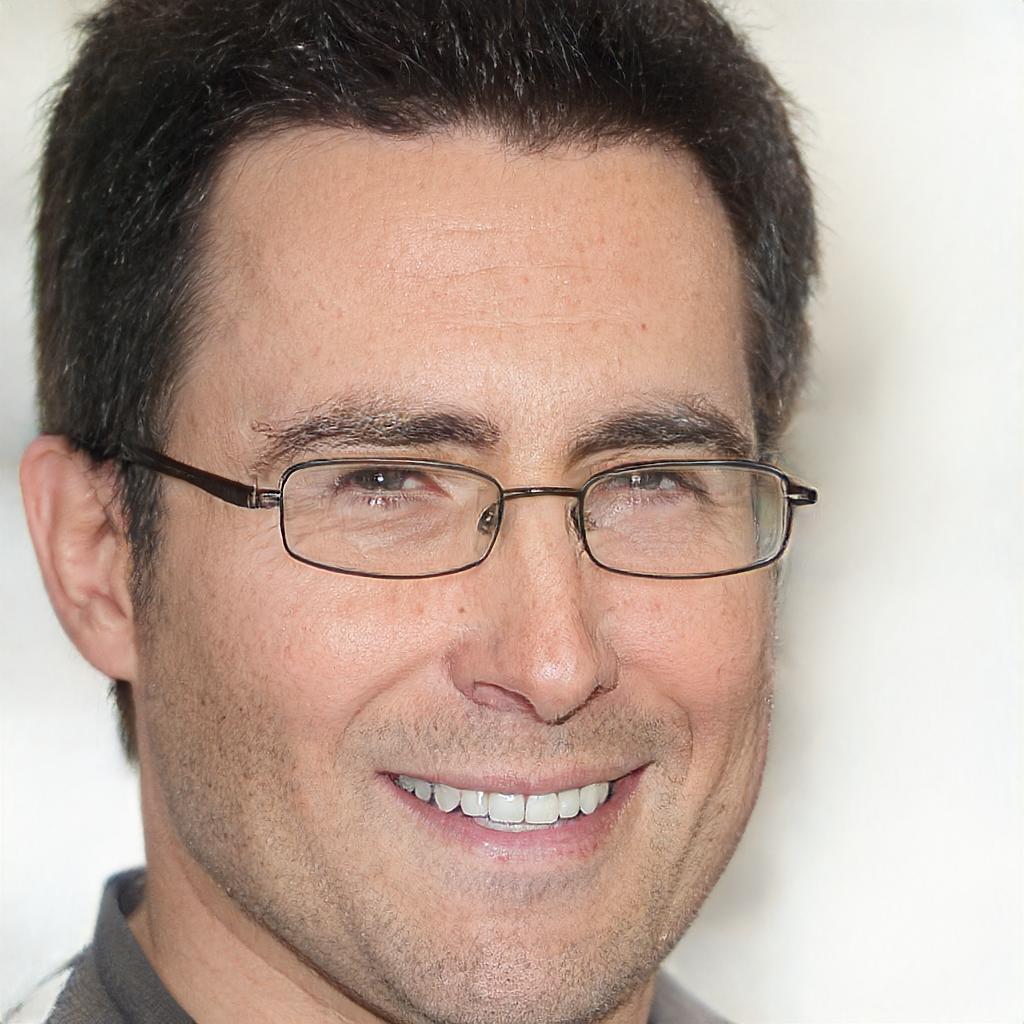}
         
     \end{subfigure}
     \hspace{.08in}
     \begin{subfigure}[b]{0.14\textwidth}
         \centering
         \includegraphics[width=\textwidth]{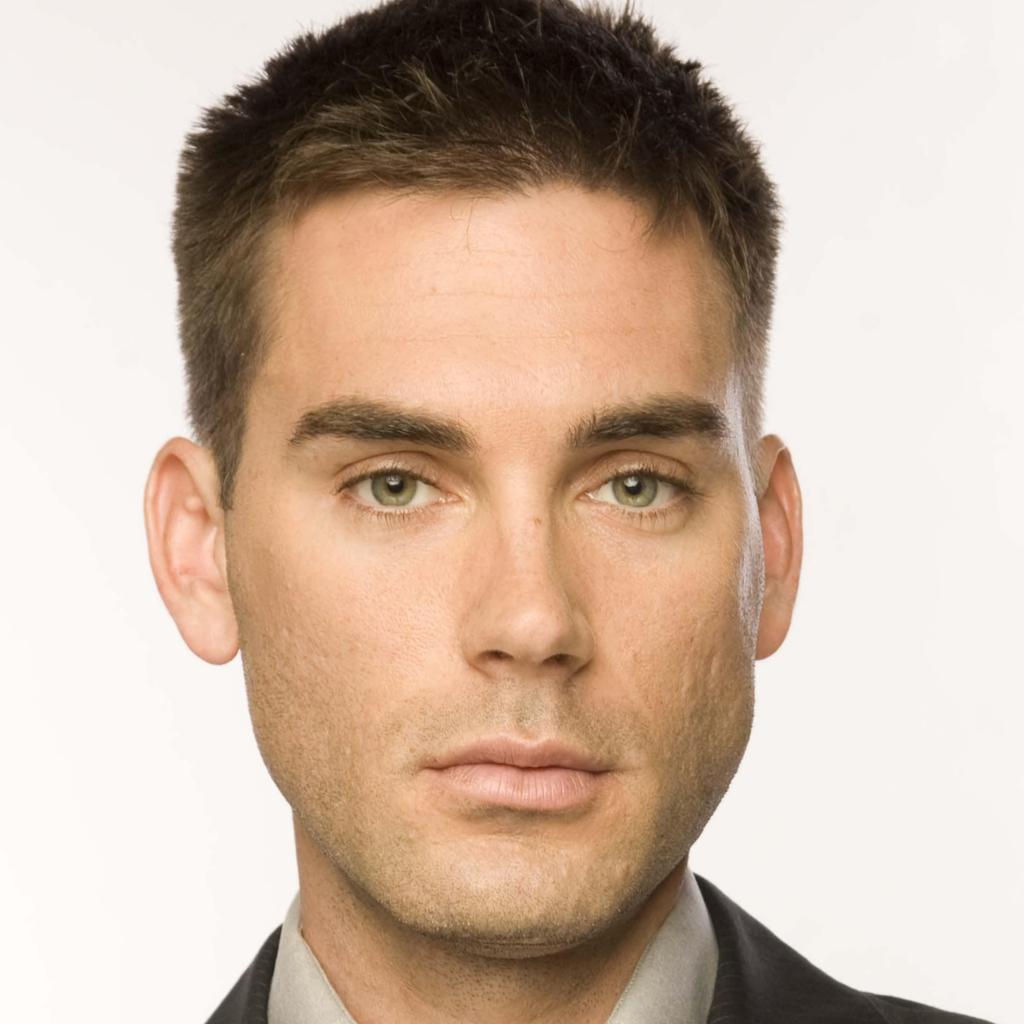}
        
     \end{subfigure}
     \\
        \caption{The visualization of source face, other different adversarial face and target face. Each image row is source, MI-FGSM, PGD, AdvHat,  TCA$^2$, and target face, respectively.}
        \label{fig:quality_assessment}
\end{figure*}

Additionally, we aim to align the adversarial face image with a controlling natural language prompt. For an image $\hat{x}_s$ guided by the text prompt $t$, our goal is for the image $\hat{x}_s$ to exhibit the attributes described by the text prompt $t$. Specifically, we use CLIP to bridge the gap between the text prompt $t$ and the image $\hat{x}_s$. The textual guidance loss is defined as follows:
\begin{equation}
\label{guide}
    \mathcal{L} _{guide} =CLIP(\hat{x}_s,t)
\end{equation}
where $CLIP(\cdot, \cdot)$ represents a pre-trained vision-language model. Additionally, apart from the attribute specified in the text prompt $t$, we aim to preserve all other attributes in the adversarial face image. This is similar to norm-based attacks, where the goal is to minimize the pixel-level differences between the clean image and the adversarial one. We apply the same principle to maintain minimal perceptual changes. The perception preservation loss is defined as follows:
\begin{equation}
\label{perc}
    \mathcal{L} _{perc} =D(\hat{x}_s,x_s)
\end{equation}
where $D$ represents a perceptual network pretrained using LPIPS. Our ultimate goal is for the adversarial face image $\hat{x}_s$ to effectively deceive the face recognition model $\mathcal{F}$. Specifically, in the context of an impersonation attack, we aim for the similarity scores between $\hat{x}_s$ and the target image $x_t$ to be higher than those of other pairs. In our approach, we use the cosine similarity loss as our adversarial impersonation loss, defined as follows:
\begin{equation}
\label{adv}
    \mathcal{L} _{adv}=cos(\mathcal{F}(\hat{x}_s),\mathcal{F}(x_t))
\end{equation}
where $cos(\cdot)$ is the cosine similarity function.
Finally, combining the three loss functions, we have
\begin{equation}
\label{impe}
    \mathcal{L} _{impe} = \lambda _{guide} \mathcal{L} _{guide} + \lambda _{perc} \mathcal{L} _{perc} + \mathcal{L} _{adv}
\end{equation}
where $\lambda _{guide}$ and $\lambda _{perc}$ are hyperparameters that balance the contributions of the respective loss terms. Here, $\mathcal{L} _{adv}$ represents the adversarial objective as defined in \refeq{adv}. Meanwhile, $\mathcal{L} _{guide}$ and $\mathcal{L} _{perc}$ correspond to the text-guided control and the preservation of other attributes in \refeq{guide} and \refeq{perc} respectively.
\subsubsection{ Enhance the Transferablity of Target Adversarial Face}
Previous impersonation attacks\cite{jia2022adv,DBLP:conf/aaai/LiuWP0H024} generate adversarial face images by optimizing a specific white-box surrogate face recognition (FR) model. This approach inevitably leads to overfitting to the white-box model, resulting in poor attack performance when targeting a black-box FR model. To address this issue, we adopt both data and model augmentation techniques to enhance the transferability of our TCA$^2$ method: 1)  \textbf{Data Augmentation}. Previous works \cite{xie2019improving, xiong2022stochastic} have demonstrated that data augmentation strategies help prevent adversarial examples from overfitting to specific data patterns. We employ this strategy to improve the generalization of the target identity. Specifically, we apply an input transformation $T(\cdot)$ with a stochastic dropout component, following the approach used in DIM \cite{xie2019improving}. In our TCA$^2$, this involves applying random resizing and padding operations on the target face $x_t$;  2) \textbf{Meta Learning}. To mitigate overfitting to a specific surrogate model, some works \cite{DBLP:conf/iclr/LiuCLS17,gubri2022lgv} have proposed enhancing the white-box model to improve generalization. Notably, model ensemble methods \cite{DBLP:conf/iclr/LiuCLS17} simply aggregate losses from multiple models. However, obtaining multiple models can be challenging, and ensemble methods still tend to overfit to the combined white-box models. Inspired by recent adversarial research \cite{fang2022learning}, we employ meta-learning to simulate both white-box and black-box environments. Specifically, given a total of $T+1$ FR models, we randomly select $T$ models for the meta-train set and 1 model for the meta-test set. In each iteration, the meta-train and meta-test sets are reshuffled from the $T+1$ FR models. We first evaluate \refeq{adv} on the meta-test set, then jointly optimize the current loss using the ensemble losses from both the meta-train and meta-test sets. The meta-learning details of our TCA$^2$ approach are provided in the Supplementary Materials.

\section{Experiments}
\label{sec:Experiment}
\subsection{Experiment Settings}
\subsubsection{Implementation details}
In our experiments, we utilize StyleGAN2, pretrained on the FFHQ face dataset, as our generative model. For adversarial text guidance, we employ the CLIP model, which is pretrained on the WIT dataset. For GAN inversion, we adopt the BDInvert method \cite{kang2021gan}. We collect 18 text prompts representing diverse facial styles for the text guidance (details provided in the Supplementary Material). For training optimization, we use the Adam optimizer with $\beta_1$ set to 0.9, $\beta_2$ set to 0.999, and a learning rate of 0.01. The training process is run for 50 epochs. We set the values of $\lambda_{guide}$ and $\lambda_{perc}$ to 0.5 and 0.05, respectively. All experiments are conducted using PyTorch on a V100 GPU with 32 GB of memory.
\subsubsection{Dataset}
We conducted experiments using two publicly available facial datasets: (1) the CelebA-Identity dataset \cite{na2022unrestricted}, which is a subset of the CelebA-HQ dataset \cite{huang2018introvae} comprising 307 identities. Each identity includes at least 15 facial images, totaling 5,478 images, all at a resolution of 1024×1024 pixels. (2) The KID-F dataset\footnote{https://github.com/PCEO-AI-CLUB/KID-F}, also known as the K-pop Idol Dataset - Female (KID-F), consists of approximately 6,000 high-quality facial images of Korean female idols. For our experiments, we selected about 2,000 images representing 100 identities from the KID-F dataset. We randomly chose 1,000 images from different identities as source images from both datasets. Additionally, five images were selected as target facial images in each dataset.
\subsubsection{Attacked Threaten Model}
To validate our proposed impersonation attack against face recognition, we trained models on the two aforementioned datasets. Specifically, well-pretrained MobileFace \cite{chinaev2018mobileface}, IRSE50 \cite{8578843}, IR152 \cite{8953658}, and FaceNet \cite{schroff2015facenet} were fine-tuned on the CelebA-Identity and KID-F datasets. All FR models aligned the input face images via MTCNN \cite{DBLP:journals/spl/ZhangZLQ16} during the preprocessing step.
\subsubsection{Evaluation Metrics}
Following \cite{deb2020advfaces}, we used the \textit{attack success rate} (ASR) to evaluate our proposed TCA$^2$. The ASR is defined as the proportion of adversarial faces that are misclassified by the face recognition model. The ASR for an impersonation attack is formulated as follows:
\begin{equation}
    A S R=\frac{\sum_{i}^{N} 1_{\tau}\left(\cos \left[\mathcal{F}\left(\hat{x}_s^{i}\right), \mathcal{F}\left(x_t^{i}\right)\right]>\tau\right)}{N} \times 100 \%
    \label{asr_impe}
\end{equation}
where $1_{\tau}$ denotes the indicator function, $\hat{x}_s$ and $x_t$ represent the generated adversarial face and the target face, respectively. 
Threshold $\tau$ is set as 0.01 $FAR$ (False Acceptance Rate), and $N$ is the number of images. $A S R$ measures the proportion of source-target pairs whose similarity scores exceed $\tau$ out of all source-target pairs. More details can be found in the Supplementary Material.

Additionally, we report the PSNR, SSIM \cite{wang2004image}, and FID \cite{heusel2017gans} scores to evaluate the imperceptibility of TCA$^2$. Higher PSNR and SSIM scores indicate greater similarity to the original images, while a lower FID score suggests more realistic images. 
\subsubsection{Baseline methods}
We compare our TCA$^2$ approach with recent noise-based and unrestricted adversarial attacks against face recognition. The noise-based methods include MI-FGSM \cite{dong2018boosting}, and PGD \cite{madry2017towards}. Unrestricted adversarial attacks include Adv-Makeup \cite{yin2021adv}, Adv-Hat \cite{komkov2021advhat}, Adv-Attribute \cite{jia2022adv}, \cite{li2021exploring}, Latent-HSJA \cite{na2022unrestricted}, and Adv-Diffusion \cite{DBLP:conf/aaai/LiuWP0H024}. Unlike traditional norm-based methods, which strictly ensure that perturbations do not exceed a set boundary, unrestricted adversarial attacks do not provide a strict guarantee that the perturbations will stay within the attribute bounds. All experimental settings closely follow those described in the original papers. Additional information is provided in the Supplementary Materials.
\subsection{Experimental Results}
\subsubsection{Baseline comparison}
In this section, we present the experimental results of TCA$^2$ on both datasets against four different pretrained face recognition models in black-box attack scenarios. To ensure a fair comparison, all TCA$^2$ results are averaged over five text style prompts.

We evaluate the black-box attack performance of TCA$^2$ on four face recognition models, namely MobileFace \cite{chinaev2018mobileface}, IRSE50 \cite{8578843}, IR152 \cite{8953658}, and FaceNet \cite{schroff2015facenet}. Notably, we adopt a leave-one-out strategy, where three face recognition (FR) models are treated as available white-box models to train our TCA$^2$ framework, with the remaining model used as the target black-box model. 
\reftbl{table:verification_impersonation} reports the ASR results of TCA$^2$ in impersonation attacks against the target models on the CelebA-Identity and KID-F datasets. In most cases, TCA$^2$ achieves a higher ASR than other traditional norm-based adversarial attacks \cite{dong2018boosting,madry2017towards} and unrestricted adversarial attacks \cite{yin2021adv,komkov2021advhat,li2021exploring,jia2022adv,DBLP:conf/aaai/LiuWP0H024}.

\begin{figure}[h]
\centerline{\includegraphics[width=18pc]{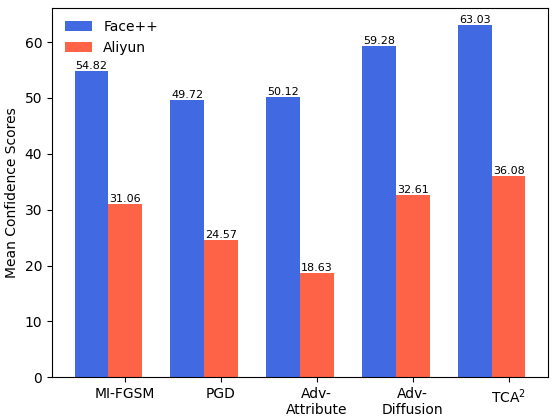}}
\caption{Mean confidence scores returned from commercial APIs, \textit{i.e}, Face++ and Aliyun.
}
\label{APIs}
\end{figure}
\begin{table*}[]
\small
	\centering
 \caption{Attack success rate ($A S R$ \%) of impersonation attack against the face recognition task on the CelebA-Identity and KID-F datasets. We choose four FR models (\textit{i.e.}, MobileFace,  IRSE50, IR152 and FaceNet) to evaluate the attack methods. \label{table:verification_impersonation}}
	\begin{tabular}{l || c c c c || c c c c  }
\toprule[0.15em]
\rowcolor{mygray} \textbf{Dataset} & \multicolumn{4}{c||}{\textbf{CelebA-Identity}}&\multicolumn{4}{c}{\textbf{KID-F}} \\
\rowcolor{mygray} \textbf{Target Model} & MobileFace & IRSE50 & IR152& FaceNet & MobileFace & IRSE50 & IR152& FaceNet \\
\midrule[0.15em]
Clean & 12.68  & 3.80  & 1.09  & 2.65  & 2.70  & 3.66  & 0.69  & 4.17     \\
Inverted & 13.57 & 2.74 & 0.66 & 3.32 & 1.81 & 2.52 & 1.45 & 6.66   \\
$\text{MI-FGSM}_{\text{(CVPR'18)}}$ & 59.89  & 70.37 & 39.00     & 34.90      & 56.15    & 65.58  & 37.74     & 32.47  \\
$\text{PGD}_{\text{(ICLR'18)}}$ & 50.10&61.73& 45.26&38.85& 43.05&62.80& 40.01&33.33 \\
Adv-Makeup & 15.59  & 54.48  & 36.37    & 32.00      & 17.64   & 50.03  & 40.97     & 32.38  \\
$\text{Adv-Hat}_{\text{(ICPR'21)}}$ ~ & 8.40 & 7.23 & 2.74& 5.27 & 5.99 & 7.77 & 10.09 & 6.44  \\
$\text{Adv-Attribute}_{\text{(NeurIPS'21)}}$  ~ & 45.59 & 56.81 & 38.67& 30.81 & 42.59 & 53.77 & 39.90 & 34.81  \\
\cite{li2021exploring}$_{\text{(CVPR'21)}}$  ~ & 28.60 & 52.74 & 30.73& 33.78 & 28.63 & 47.04 & 28.70 & 33.46  \\
$\text{Latent-HSJA}_{\text{(ECCV'22)}}$ ~ & 14.40 & 37.15 & 14.77& 12.64 & 15.99 & 33.51 & 16.71 & 15.82  \\
$\text{Adv-Diffusion}_{\text{(AAAI'24)}}$ ~ & 72.27 & \textbf{80.08} & 52.88& 36.12 & 65.55 & \textbf{83.79} & 52.01 & 35.99  \\
\midrule
\rowcolor{orange!6} TCA$^2$ & \textbf{73.10} &78.42 & \textbf{53.72} & \textbf{42.26} & \textbf{65.57} & 82.31 & \textbf{53.51} & \textbf{39.64}   \\
\bottomrule[0.1em]
\end{tabular}
\end{table*}

\subsubsection{Image Quality Assessment}
In addition to the effectiveness of adversarial examples, their concealment is a crucial quality. Ideally, an adversarial image should be indistinguishable from a non-adversarial image. The FID scores of TCA$^2$ on two datasets, reported in Supplementary Materials, assess the naturalness of the generated images. Leveraging the capabilities of StyleGAN, TCA$^2$ is able to produce high-quality, photorealistic adversarial face images. Additionally, the PSNR and SSIM results are presented in Supplementary Materials.  Visualizations of the adversarial faces generated by different attack methods are shown in \refig{fig:quality_assessment}.

\subsubsection{Attacks on commercial model via APIs}
To further validate the effectiveness and transferability of TCA$^2$ in real-world scenarios, we evaluated our algorithm using two well-known commercial face verification APIs: Face++ and Aliyun. The experimental results on the CelebA-Identity dataset are presented in \refig{APIs}. As shown in \refig{APIs}, TCA$^2$ outperforms the state-of-the-art method, Adv-Diffusion, on both commercial models.

\subsubsection{Ablation Study}
In this section, we will report some ablation results to evaluate the contributions of our TCA$^2$ components.

\begin{figure}
    \centering
    
     \begin{subfigure}[b]{0.14\textwidth}
         \centering
         \includegraphics[width=\textwidth]{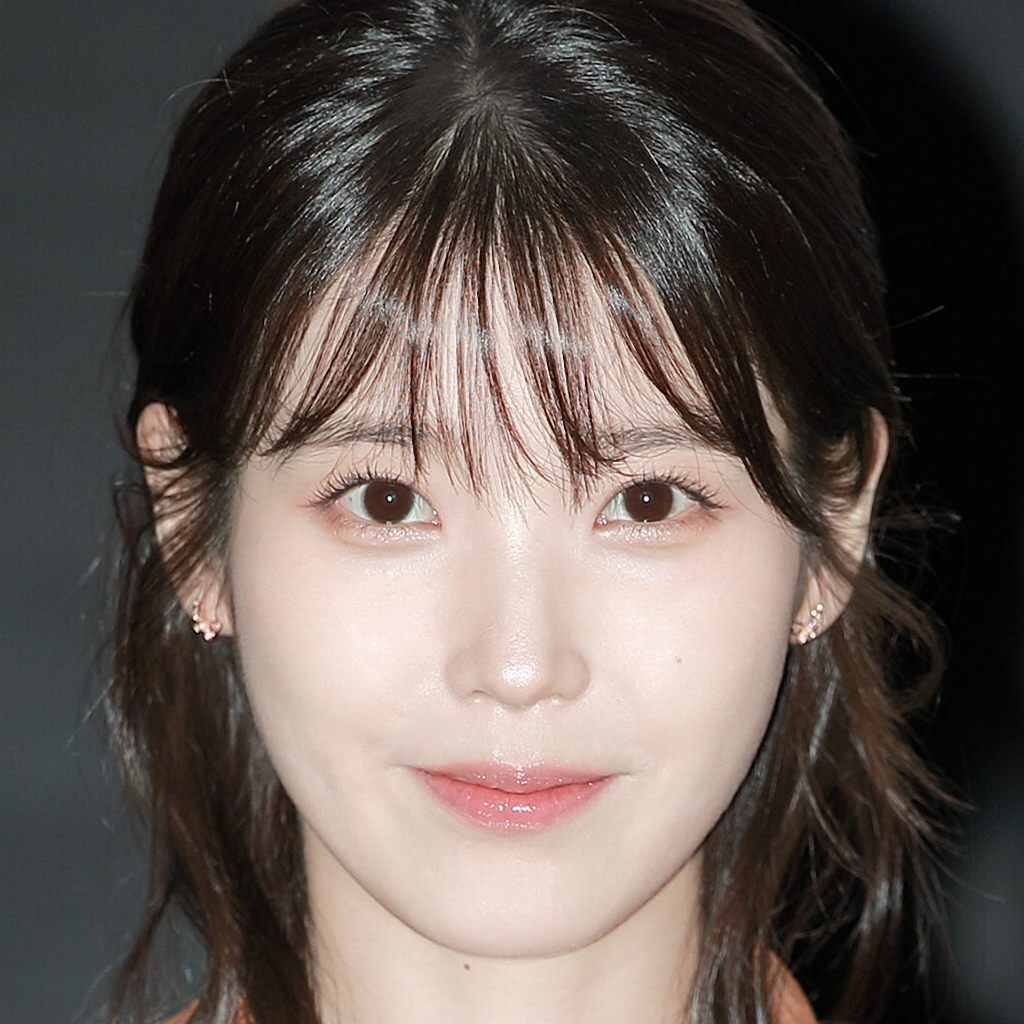}
         \caption{Original}
     \end{subfigure}
     \hspace{.08in}
     \begin{subfigure}[b]{0.14\textwidth}
         \centering
         \includegraphics[width=\textwidth]{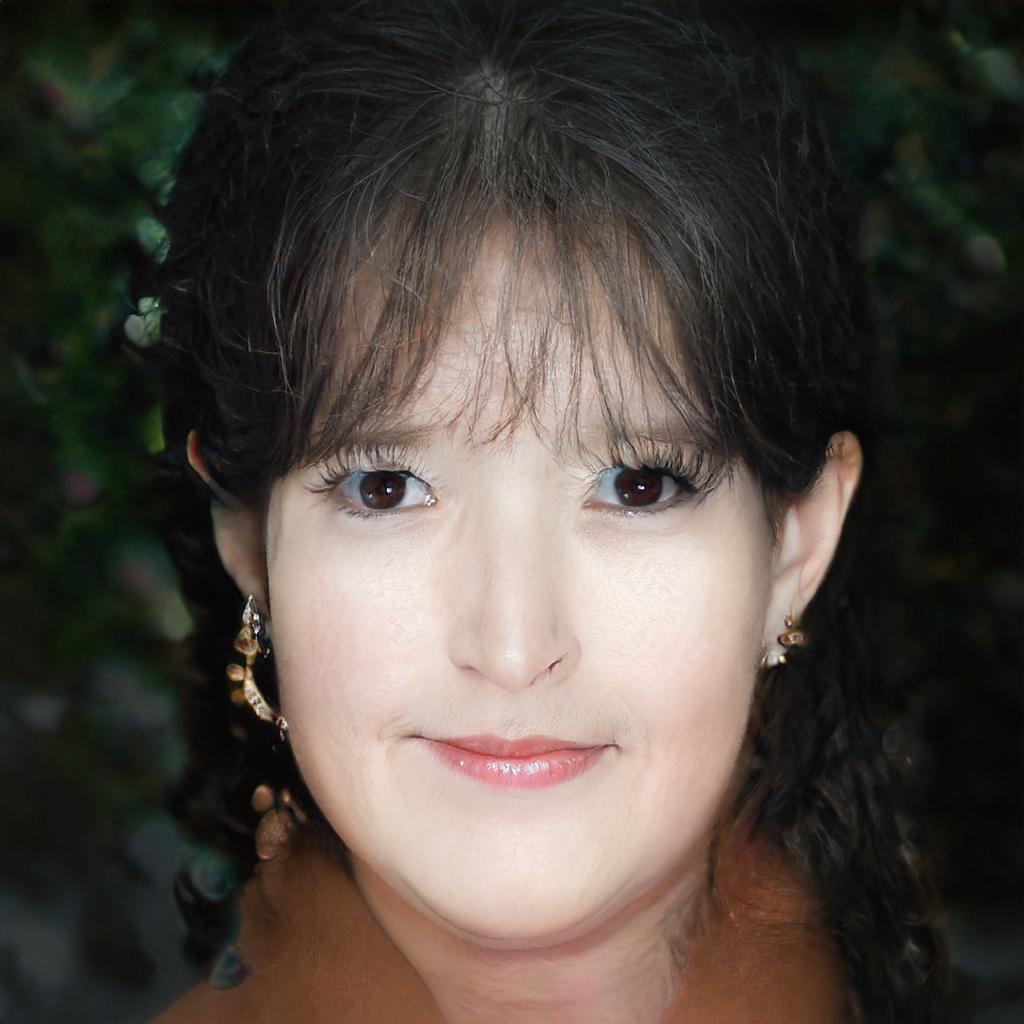}
         \caption{w/o text guide}
         \end{subfigure}
    \hspace{.08in}
    \begin{subfigure}[b]{0.14\textwidth}
         \centering
         \includegraphics[width=\textwidth]{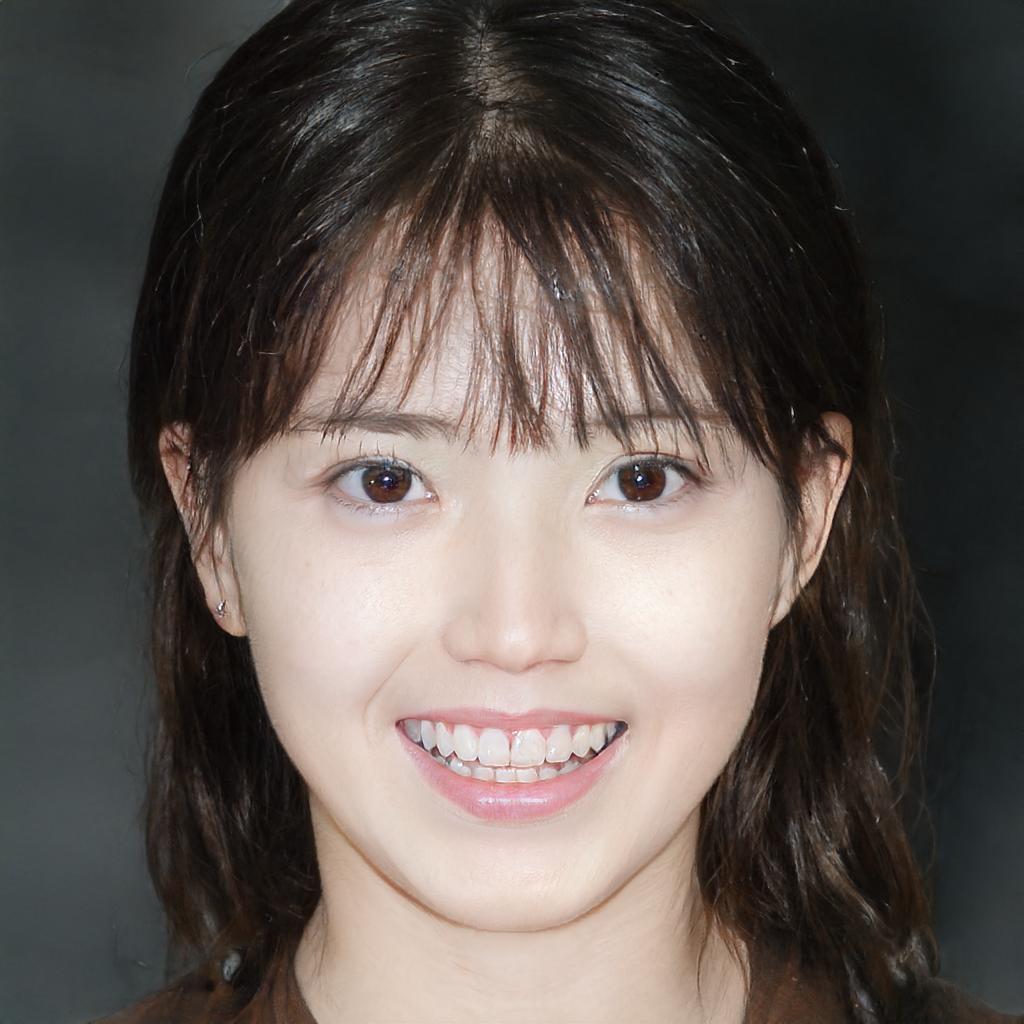}
         \caption{w text guide}
         \end{subfigure}
         \caption{The visualizations of the text prompt's style on the visual quality of the output images. These images successfully deceive the facial recognition (FR) model. The text prompt used was "\textit{A female face with open mouth}".\label{fig:txt_guide}}
\end{figure}

\begin{itemize}
    \item \textbf{Style text prompt:} As illustrated in \refig{fig:txt_guide}, without the guidance of a text prompt, the generated image tends to introduce globally visible perturbations aimed at optimizing the adversarial objective. These global perturbations have two main consequences: first, they can result in the generated image appearing unnatural; second, they limit the ability to offer users the option to select a desired style attribute compared to the clean image.

    \begin{figure}[!ht]
     \centering
     \begin{subfigure}[b]{0.14\textwidth}
         \centering
         \includegraphics[width=\textwidth]{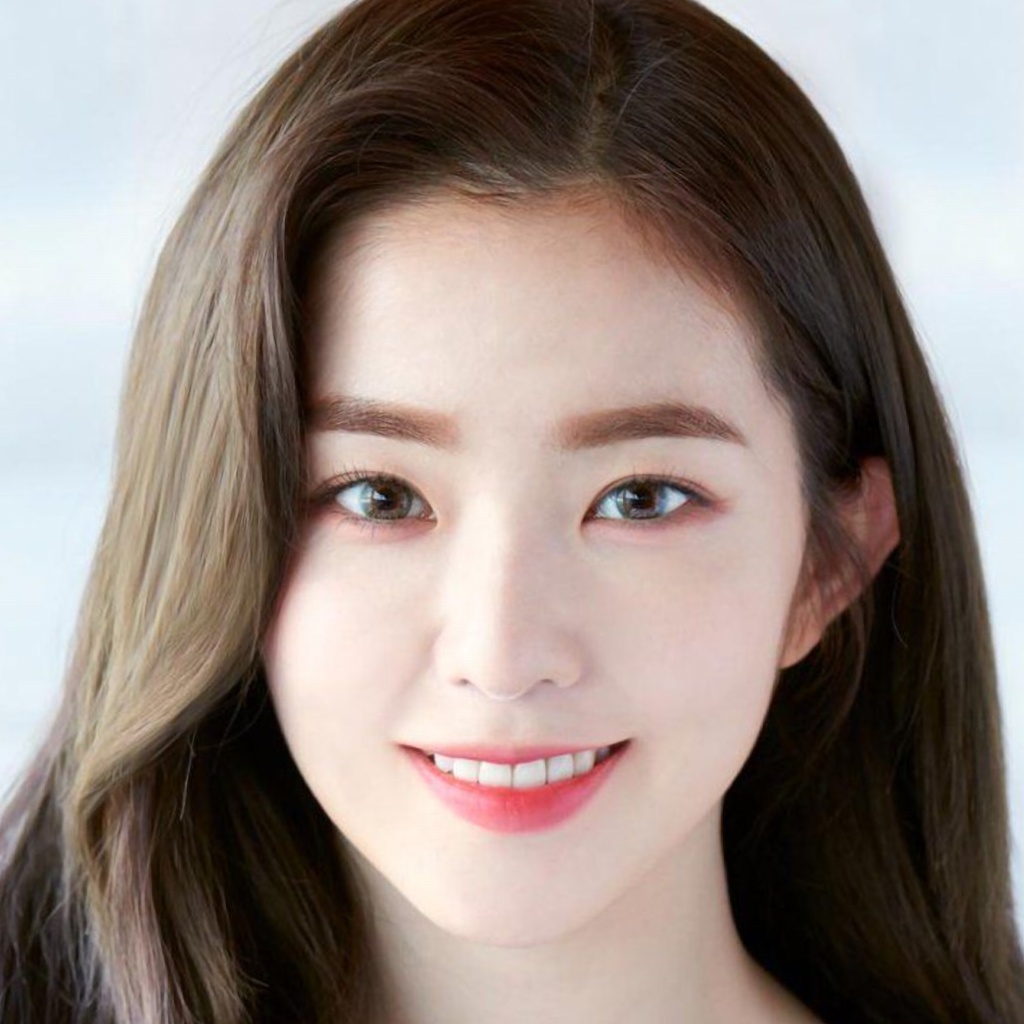}
         \caption{Original Face Image}
     \end{subfigure}
     \hfill
     \begin{subfigure}[b]{0.14\textwidth}
         \centering
         \includegraphics[width=\textwidth]{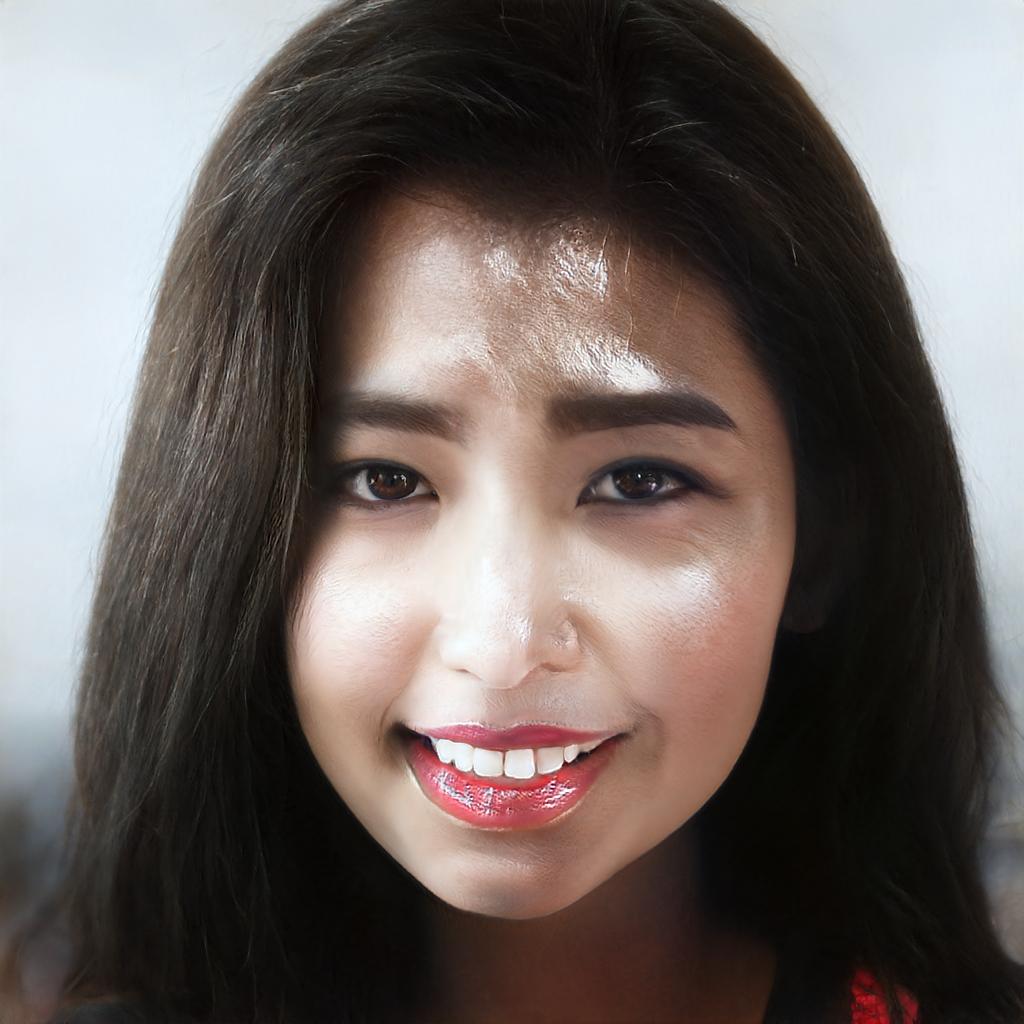}
         \caption{w/o restriction}
         \end{subfigure}
     \hfill
    \begin{subfigure}[b]{0.14\textwidth}
         \centering
         \includegraphics[width=\textwidth]{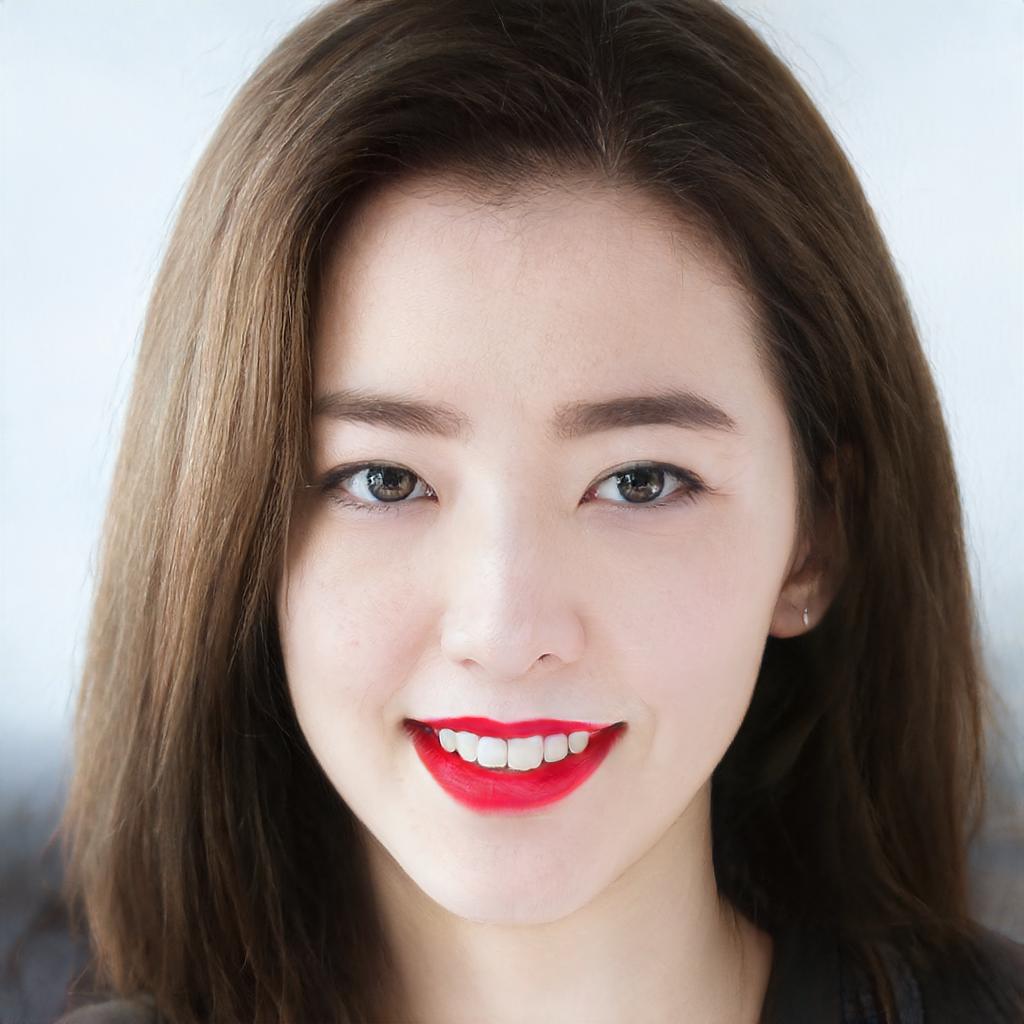}
         \caption{w text restriction}
         \end{subfigure}
     
        \caption{The visualizations of impact of perception preservation on the visual quality of the output images. The same text prompt, '\textit{A female face with red lipstick},' is applied to both images.}
        \label{fig:perp_guide}
\end{figure}
    
    \item \textbf{Perception preservation:} To evaluate the impact of perception preservation in TCA$^2$, we removed all perception preservation constraints. The visualization is presented in 
\refig{fig:perp_guide}. Our perception preservation constraints guide the adversarial optimization to explore the vicinity of the original latent code.
\end{itemize}
Additionally, we analyze the impact of other variables, specifically the hyperparameters $\lambda_{guide}$ and $\lambda_{perc}$. Furthermore, we perform an ablation study to assess the impact on transferability. These results are provided in the Supplementary Materials.

\section{Conclusion}

In this paper, we proposed a novel approach that leverages natural language to guide the style latent code of StyleGAN2 in generating photorealistic face images capable of conducting impersonation attacks against face recognition systems. Additionally, TCA$^2$ demonstrates superior attack generalization across different face recognition models, making the generated adversarial images highly transferable to unknown black-box systems. Extensive experiments show that the faces generated using our method not only embody the desired attributes specified in the controlled text but also successfully deceive state-of-the-art face recognition systems, including commercial APIs, with a high success rate. However, defense mechanisms against unrestricted adversarial attacks remain underexplored. In future work, we plan to investigate more generalized defense strategies to enhance the robustness of face recognition systems against both norm-based and unrestricted adversarial attacks.
\bibliographystyle{aaai}
\bibliography{bibliography}

\begin{thebibliography}{}

\bibitem[\protect\citeauthoryear{Ali \bgroup et al\mbox.\egroup }{2021}]{ali2021classical}
Ali, W.; Tian, W.; Din, S.~U.; Iradukunda, D.; and Khan, A.~A.
\newblock 2021.
\newblock Classical and modern face recognition approaches: a complete review.
\newblock {\em Multimedia Tools and Applications} 80:4825--4880.

\bibitem[\protect\citeauthoryear{Brown \bgroup et al\mbox.\egroup }{2017}]{brown2017adversarial}
Brown, T.~B.; Man{\'e}, D.; Roy, A.; Abadi, M.; and Gilmer, J.
\newblock 2017.
\newblock Adversarial patch.
\newblock {\em arXiv preprint arXiv:1712.09665}.

\bibitem[\protect\citeauthoryear{Carlini and Wagner}{2017}]{carlini2017towards}
Carlini, N., and Wagner, D.
\newblock 2017.
\newblock Towards evaluating the robustness of neural networks.
\newblock In {\em IEEE SP},  39--57.
\newblock Ieee.

\bibitem[\protect\citeauthoryear{Chinaev, Chigorin, and Laptev}{2018}]{chinaev2018mobileface}
Chinaev, N.; Chigorin, A.; and Laptev, I.
\newblock 2018.
\newblock Mobileface: 3d face reconstruction with efficient cnn regression.

\bibitem[\protect\citeauthoryear{Deb, Zhang, and Jain}{2020}]{deb2020advfaces}
Deb, D.; Zhang, J.; and Jain, A.~K.
\newblock 2020.
\newblock Advfaces: Adversarial face synthesis.
\newblock In {\em 2020 IEEE International Joint Conference on Biometrics (IJCB)},  1--10.
\newblock IEEE.

\bibitem[\protect\citeauthoryear{Deng \bgroup et al\mbox.\egroup }{2019}]{8953658}
Deng, J.; Guo, J.; Xue, N.; and Zafeiriou, S.
\newblock 2019.
\newblock Arcface: Additive angular margin loss for deep face recognition.
\newblock In {\em CVPR},  4685--4694.

\bibitem[\protect\citeauthoryear{Dong \bgroup et al\mbox.\egroup }{2018}]{dong2018boosting}
Dong, Y.; Liao, F.; Pang, T.; Su, H.; Zhu, J.; Hu, X.; and Li, J.
\newblock 2018.
\newblock Boosting adversarial attacks with momentum.
\newblock In {\em CVPR},  9185--9193.

\bibitem[\protect\citeauthoryear{Dong \bgroup et al\mbox.\egroup }{2019}]{dong2019efficient}
Dong, Y.; Su, H.; Wu, B.; Li, Z.; Liu, W.; Zhang, T.; and Zhu, J.
\newblock 2019.
\newblock Efficient decision-based black-box adversarial attacks on face recognition.
\newblock In {\em CVPR},  7714--7722.

\bibitem[\protect\citeauthoryear{Fang \bgroup et al\mbox.\egroup }{2022}]{fang2022learning}
Fang, S.; Li, J.; Lin, X.; and Ji, R.
\newblock 2022.
\newblock Learning to learn transferable attack.
\newblock In {\em AAAI}, volume~36,  571--579.

\bibitem[\protect\citeauthoryear{Goodfellow, Shlens, and Szegedy}{2014}]{goodfellow2014explaining}
Goodfellow, I.~J.; Shlens, J.; and Szegedy, C.
\newblock 2014.
\newblock Explaining and harnessing adversarial examples.
\newblock {\em arXiv preprint arXiv:1412.6572}.

\bibitem[\protect\citeauthoryear{Gubri \bgroup et al\mbox.\egroup }{2022}]{gubri2022lgv}
Gubri, M.; Cordy, M.; Papadakis, M.; Traon, Y.~L.; and Sen, K.
\newblock 2022.
\newblock Lgv: Boosting adversarial example transferability from large geometric vicinity.
\newblock In {\em ECCV},  603--618.
\newblock Springer.

\bibitem[\protect\citeauthoryear{Guetta \bgroup et al\mbox.\egroup }{2021}]{guetta2021dodging}
Guetta, N.; Shabtai, A.; Singh, I.; Momiyama, S.; and Elovici, Y.
\newblock 2021.
\newblock Dodging attack using carefully crafted natural makeup.
\newblock {\em arXiv preprint arXiv:2109.06467}.

\bibitem[\protect\citeauthoryear{Heusel \bgroup et al\mbox.\egroup }{2017}]{heusel2017gans}
Heusel, M.; Ramsauer, H.; Unterthiner, T.; Nessler, B.; and Hochreiter, S.
\newblock 2017.
\newblock Gans trained by a two time-scale update rule converge to a local nash equilibrium.
\newblock {\em NeurIPS} 30.

\bibitem[\protect\citeauthoryear{Hu, Shen, and Sun}{2018}]{8578843}
Hu, J.; Shen, L.; and Sun, G.
\newblock 2018.
\newblock Squeeze-and-excitation networks.
\newblock In {\em CVPR},  7132--7141.

\bibitem[\protect\citeauthoryear{Huang \bgroup et al\mbox.\egroup }{2018}]{huang2018introvae}
Huang, H.; He, R.; Sun, Z.; Tan, T.; et~al.
\newblock 2018.
\newblock Introvae: Introspective variational autoencoders for photographic image synthesis.
\newblock {\em NeurIPS} 31.

\bibitem[\protect\citeauthoryear{Ilyas \bgroup et al\mbox.\egroup }{2018}]{ilyas2018black}
Ilyas, A.; Engstrom, L.; Athalye, A.; and Lin, J.
\newblock 2018.
\newblock Black-box adversarial attacks with limited queries and information.
\newblock In {\em ICML},  2137--2146.
\newblock PMLR.

\bibitem[\protect\citeauthoryear{Jia \bgroup et al\mbox.\egroup }{2022}]{jia2022adv}
Jia, S.; Yin, B.; Yao, T.; Ding, S.; Shen, C.; Yang, X.; and Ma, C.
\newblock 2022.
\newblock Adv-attribute: Inconspicuous and transferable adversarial attack on face recognition.
\newblock {\em NeurIPS} 35:34136--34147.

\bibitem[\protect\citeauthoryear{Kang, Kim, and Cho}{2021}]{kang2021gan}
Kang, K.; Kim, S.; and Cho, S.
\newblock 2021.
\newblock Gan inversion for out-of-range images with geometric transformations.
\newblock In {\em ICCV},  13941--13949.

\bibitem[\protect\citeauthoryear{Karmon, Zoran, and Goldberg}{2018}]{karmon2018lavan}
Karmon, D.; Zoran, D.; and Goldberg, Y.
\newblock 2018.
\newblock Lavan: Localized and visible adversarial noise.
\newblock In {\em ICML},  2507--2515.
\newblock PMLR.

\bibitem[\protect\citeauthoryear{Karras \bgroup et al\mbox.\egroup }{2020}]{karras2020analyzing}
Karras, T.; Laine, S.; Aittala, M.; Hellsten, J.; Lehtinen, J.; and Aila, T.
\newblock 2020.
\newblock Analyzing and improving the image quality of stylegan.
\newblock In {\em CVPR},  8110--8119.

\bibitem[\protect\citeauthoryear{Komkov and Petiushko}{2021}]{komkov2021advhat}
Komkov, S., and Petiushko, A.
\newblock 2021.
\newblock Advhat: Real-world adversarial attack on arcface face id system.
\newblock In {\em 2020 International Conference on Pattern Recognition (ICPR)},  819--826.
\newblock IEEE.

\bibitem[\protect\citeauthoryear{Li \bgroup et al\mbox.\egroup }{2021}]{li2021exploring}
Li, D.; Wang, W.; Fan, H.; and Dong, J.
\newblock 2021.
\newblock Exploring adversarial fake images on face manifold.
\newblock In {\em CVPR},  5789--5798.

\bibitem[\protect\citeauthoryear{Liu \bgroup et al\mbox.\egroup }{2017}]{DBLP:conf/iclr/LiuCLS17}
Liu, Y.; Chen, X.; Liu, C.; and Song, D.
\newblock 2017.
\newblock Delving into transferable adversarial examples and black-box attacks.
\newblock In {\em ICLR}.

\bibitem[\protect\citeauthoryear{Liu \bgroup et al\mbox.\egroup }{2024}]{DBLP:conf/aaai/LiuWP0H024}
Liu, D.; Wang, X.; Peng, C.; Wang, N.; Hu, R.; and Gao, X.
\newblock 2024.
\newblock Adv-diffusion: imperceptible adversarial face identity attack via latent diffusion model.
\newblock In {\em AAAI}, volume~38,  3585--3593.

\bibitem[\protect\citeauthoryear{Madry \bgroup et al\mbox.\egroup }{2017}]{madry2017towards}
Madry, A.; Makelov, A.; Schmidt, L.; Tsipras, D.; and Vladu, A.
\newblock 2017.
\newblock Towards deep learning models resistant to adversarial attacks.
\newblock {\em arXiv preprint arXiv:1706.06083}.

\bibitem[\protect\citeauthoryear{Massoli \bgroup et al\mbox.\egroup }{2021}]{massoli2021detection}
Massoli, F.~V.; Carrara, F.; Amato, G.; and Falchi, F.
\newblock 2021.
\newblock Detection of face recognition adversarial attacks.
\newblock {\em Computer Vision and Image Understanding} 202:103103.

\bibitem[\protect\citeauthoryear{Na, Ji, and Kim}{2022}]{na2022unrestricted}
Na, D.; Ji, S.; and Kim, J.
\newblock 2022.
\newblock Unrestricted black-box adversarial attack using gan with limited queries.
\newblock In {\em ECCV},  467--482.
\newblock Springer.

\bibitem[\protect\citeauthoryear{Qiu \bgroup et al\mbox.\egroup }{2020}]{qiu2020semanticadv}
Qiu, H.; Xiao, C.; Yang, L.; Yan, X.; Lee, H.; and Li, B.
\newblock 2020.
\newblock Semanticadv: Generating adversarial examples via attribute-conditioned image editing.
\newblock In {\em ECCV},  19--37.
\newblock Springer.

\bibitem[\protect\citeauthoryear{Radford \bgroup et al\mbox.\egroup }{2021}]{radford2021learning}
Radford, A.; Kim, J.~W.; Hallacy, C.; Ramesh, A.; Goh, G.; Agarwal, S.; Sastry, G.; Askell, A.; Mishkin, P.; Clark, J.; et~al.
\newblock 2021.
\newblock Learning transferable visual models from natural language supervision.
\newblock In {\em ICML},  8748--8763.
\newblock PMLR.

\bibitem[\protect\citeauthoryear{Ryu, Park, and Choi}{2021}]{DBLP:journals/istr/RyuPC21}
Ryu, G.; Park, H.; and Choi, D.
\newblock 2021.
\newblock Adversarial attacks by attaching noise markers on the face against deep face recognition.
\newblock {\em Journal of Information Security and Applications} 60:102874.

\bibitem[\protect\citeauthoryear{Schroff, Kalenichenko, and Philbin}{2015}]{schroff2015facenet}
Schroff, F.; Kalenichenko, D.; and Philbin, J.
\newblock 2015.
\newblock Facenet: A unified embedding for face recognition and clustering.
\newblock In {\em CVPR},  815--823.

\bibitem[\protect\citeauthoryear{Sharif \bgroup et al\mbox.\egroup }{2016}]{sharif2016accessorize}
Sharif, M.; Bhagavatula, S.; Bauer, L.; and Reiter, M.~K.
\newblock 2016.
\newblock Accessorize to a crime: Real and stealthy attacks on state-of-the-art face recognition.
\newblock In {\em Proceedings of the 2016 ACM Sigsac Conference on Computer and Communications Security},  1528--1540.

\bibitem[\protect\citeauthoryear{Szegedy \bgroup et al\mbox.\egroup }{2013}]{szegedy2013intriguing}
Szegedy, C.; Zaremba, W.; Sutskever, I.; Bruna, J.; Erhan, D.; Goodfellow, I.; and Fergus, R.
\newblock 2013.
\newblock Intriguing properties of neural networks.
\newblock {\em arXiv preprint arXiv:1312.6199}.

\bibitem[\protect\citeauthoryear{Vakhshiteh, Nickabadi, and Ramachandra}{2021}]{vakhshiteh2021adversarial}
Vakhshiteh, F.; Nickabadi, A.; and Ramachandra, R.
\newblock 2021.
\newblock Adversarial attacks against face recognition: A comprehensive study.
\newblock {\em IEEE Access} 9:92735--92756.

\bibitem[\protect\citeauthoryear{Vanschoren}{2019}]{vanschoren2019meta}
Vanschoren, J.
\newblock 2019.
\newblock Meta-learning.
\newblock {\em Automated machine learning: methods, systems, challenges}  35--61.

\bibitem[\protect\citeauthoryear{Wang \bgroup et al\mbox.\egroup }{2004}]{wang2004image}
Wang, Z.; Bovik, A.~C.; Sheikh, H.~R.; and Simoncelli, E.~P.
\newblock 2004.
\newblock Image quality assessment: from error visibility to structural similarity.
\newblock {\em IEEE TIP} 13(4):600--612.

\bibitem[\protect\citeauthoryear{Wang \bgroup et al\mbox.\egroup }{2021}]{Wang0WSL21}
Wang, X.; Zhang, Z.; Wu, B.; Shen, F.; and Lu, G.
\newblock 2021.
\newblock Prototype-supervised adversarial network for targeted attack of deep hashing.
\newblock In {\em CVPR},  16357--16366.

\bibitem[\protect\citeauthoryear{Xiao \bgroup et al\mbox.\egroup }{2018}]{xiao2019generating}
Xiao, C.; Li, B.; Zhu, J.-Y.; He, W.; Liu, M.; and Song, D.
\newblock 2018.
\newblock Generating adversarial examples with adversarial networks.
\newblock In {\em IJCAI},  3905--3911.

\bibitem[\protect\citeauthoryear{Xie \bgroup et al\mbox.\egroup }{2019}]{xie2019improving}
Xie, C.; Zhang, Z.; Zhou, Y.; Bai, S.; Wang, J.; Ren, Z.; and Yuille, A.~L.
\newblock 2019.
\newblock Improving transferability of adversarial examples with input diversity.
\newblock In {\em CVPR},  2730--2739.

\bibitem[\protect\citeauthoryear{Xiong \bgroup et al\mbox.\egroup }{2022}]{xiong2022stochastic}
Xiong, Y.; Lin, J.; Zhang, M.; Hopcroft, J.~E.; and He, K.
\newblock 2022.
\newblock Stochastic variance reduced ensemble adversarial attack for boosting the adversarial transferability.
\newblock In {\em CVPR},  14983--14992.

\bibitem[\protect\citeauthoryear{Yin \bgroup et al\mbox.\egroup }{2021}]{yin2021adv}
Yin, B.; Wang, W.; Yao, T.; Guo, J.; Kong, Z.; Ding, S.; Li, J.; and Liu, C.
\newblock 2021.
\newblock Adv-makeup: A new imperceptible and transferable attack on face recognition.
\newblock {\em arXiv preprint arXiv:2105.03162}.

\bibitem[\protect\citeauthoryear{Zhang \bgroup et al\mbox.\egroup }{2016}]{DBLP:journals/spl/ZhangZLQ16}
Zhang, K.; Zhang, Z.; Li, Z.; and Qiao, Y.
\newblock 2016.
\newblock Joint face detection and alignment using multitask cascaded convolutional networks.
\newblock {\em {IEEE} Signal Process. Lett.} 23(10):1499--1503.

\bibitem[\protect\citeauthoryear{Zhang \bgroup et al\mbox.\egroup }{2018}]{zhang2018unreasonable}
Zhang, R.; Isola, P.; Efros, A.~A.; Shechtman, E.; and Wang, O.
\newblock 2018.
\newblock The unreasonable effectiveness of deep features as a perceptual metric.
\newblock In {\em CVPR},  586--595.

\bibitem[\protect\citeauthoryear{Zhang \bgroup et al\mbox.\egroup }{2022}]{ZhangWLSZ22}
Zhang, Z.; Wang, X.; Lu, G.; Shen, F.; and Zhu, L.
\newblock 2022.
\newblock Targeted attack of deep hashing via prototype-supervised adversarial networks.
\newblock {\em {IEEE} Trans. Multim.} 24:3392--3404.

\end{thebibliography}
\newpage
\section*{Appendix}
\section{Impact of different level $\omega$ in StyleGan}
\label{StyleGan}
\begin{figure*}[!t]
\centering
\includegraphics[scale=0.65]{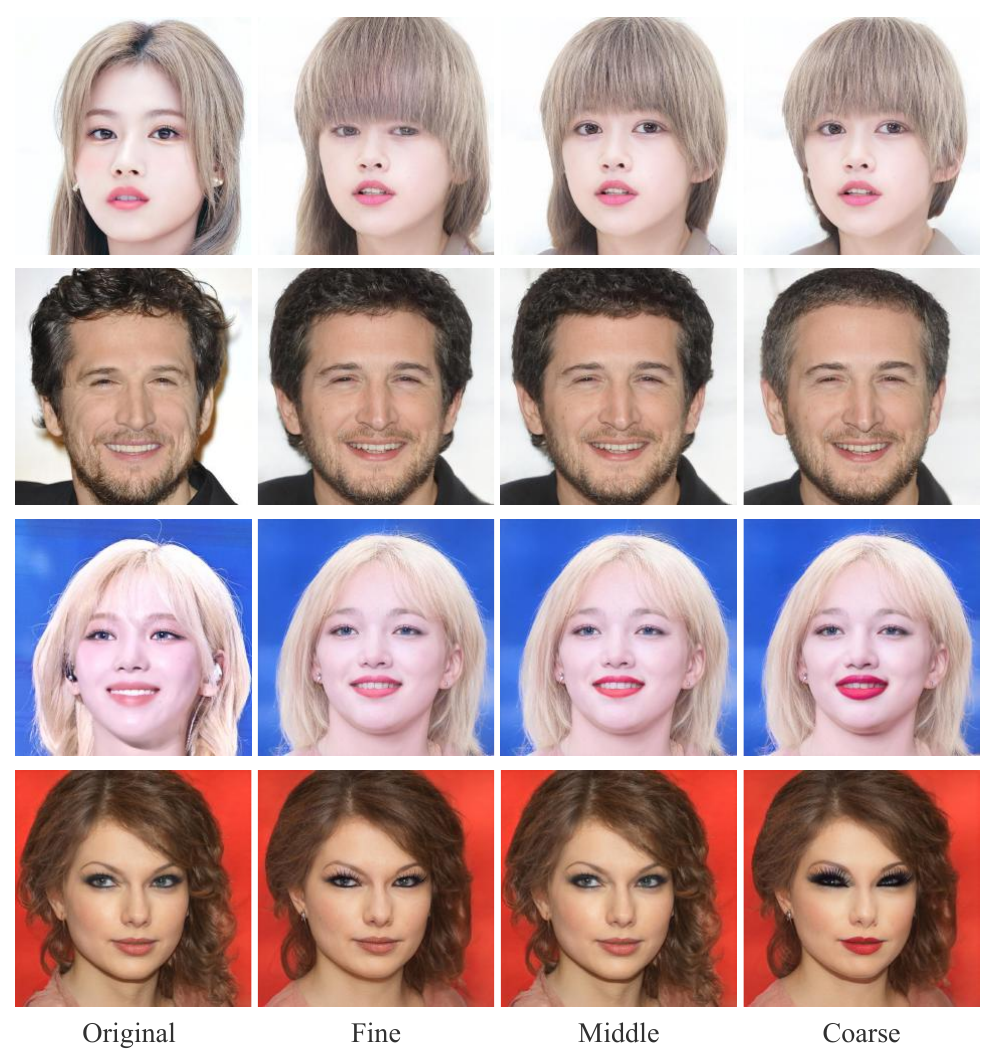}
\caption{The visualizations of impact of style levels on controlling attributes of varying granularity in images. Each row of images is generated using the same text prompt. The latent codes from shallower layers produce coarser attributes, while those from deeper layers yield finer details. \label{different_depth}}
\end{figure*}

Sec.3.2 provides an introduction to the preliminaries of StyleGAN. In StyleGAN, the latent code $\omega$ at different depths influences the generated attributes to varying degrees: shallow layers control coarse attributes, middle layers control intermediate attributes, and deep layers control fine attributes. This impact is illustrated in \refig{different_depth}.

\section{Meta learning based adversarial attack}
\label{Meta_learning}
In this section, we detail the meta-learning-based strategy \cite{vanschoren2019meta} employed to enhance the transferability of attacks against unknown face recognition (FR) models (refer to Section 3.3.2 of the main paper). Our approach is structured into two phases: meta-training and meta-testing. The overall objective function is presented in the main paper as follows:
\begin{equation}
\label{impe1}
    \mathcal{L} _{impe} = \lambda _{guide} \mathcal{L} _{guide} + \lambda _{perc} \mathcal{L} _{perc} + \mathcal{L} _{adv}
\end{equation}

\textbf{Meta-Train}: Given a total of $T+1$ face recognition (FR) models, we randomly select $T$ models for the meta-training set and one model for the meta-testing set. The meta-training and meta-testing sets are then reshuffled at each iteration. The learnable model parameters of the Meta-Learning Framework Network (MLFN) are denoted by $\Theta_M$.

The meta-training loss for the $i^{th}$ face recognition (FR) model is defined as:
\begin{equation}
\label{Meta-Train}
    \mathcal{L}_i^{tr}\left ( \Theta_M \right ) =cos\left ( \mathcal{F}_i\left ( G_L\left ( M_{\Theta_M}([\omega_s ,E_t, v ]) \right )  \right ) ,\mathcal{F}_i(x_t) \right ) 
\end{equation}
where $i \in \left \{ 1,\dots T \right \} $ , $\omega_s$ represents the inverted latent code corresponding to $x_s$, $E_t$ denotes the textual CLIP embedding, and the softmax vector $v$ from Sec.3.3.1 is expressed as follows:
\begin{equation}
        v=\left\{\begin{array}{ll}
\mathcal{F}_i^{tr}(\operatorname{Resize}(\operatorname{RandPad}(\operatorname{RandResize}(\boldsymbol{x}_t)))), & p \leq 0.5 \\
\mathcal{F}_i^{tr}(\boldsymbol{x}_t), & p>0.5
\end{array}\right.
\end{equation} 
where $p \sim  \mathcal{U}(0,1)$ represents the probability of applying a transformation to the image. The MLFN integrates these variables into an adversarial latent code $\omega_s^{*}$. Subsequently, the StyleGAN generator $G_L$ synthesizes an adversarial face image intended to be misidentified as $x_t$. For $i^{th}$ model, the update to $\Theta_M$ is performed as follows:
\begin{equation}
\label{Theta_M}
    {\Theta_{M}}_i'\gets \Theta_M-\alpha _1\triangledown _{\Theta_M}\mathcal{L}_i^{tr}\left ( \Theta_M \right )
\end{equation}
where $\alpha_1$ is the learning rate for the meta update.

\textbf{Meta-Test}: The face recognition model that is excluded from the meta-training phase is employed as the meta-test model. The updated parameters $\Theta_{M}$, derived from each of the meta-train models, are then used to attack the meta-test model. This process is formulated as follows:
\begin{equation}
\label{meta_test}
    \mathcal{L}_i^{te}\left ( \Theta_M' \right ) =cos\left ( \mathcal{F}_i\left ( G_L\left ( M_{\Theta_M'}([\omega_s ,E_t, v ]) \right )  \right ) ,\mathcal{F}_i(x_t) \right ) 
\end{equation}

\textbf{Meta-Optimization}: The parameter $\Theta_M$ is ultimately updated during both the meta-train and meta-test stages as follows:
\begin{align}
\label{meta_fun}
{\Theta_M}^\ast =\underset{\Theta_M}{argmin} \lambda _{guide} \mathcal{L} _{guide} + 
 \lambda _{perc} \mathcal{L} _{perc} + \nonumber \\ 
\sum_{i=1}^{T}\left ( \mathcal{L}_i^{tr}\left ( \Theta_M \right )+\mathcal{L}_i^{te}\left ( \Theta_M' \right )  \right )  
\end{align}
where the last term denotes the aggregation of losses from the meta-train and meta-test stages.

The \refeq{meta_fun} can be optimized using the gradient descent method, as described in \refalgo{alg:alg1}.

\begin{algorithm} 
\caption{The algorithm of TCA$^2$ against FR in impersonation setting.} 
\label{alg:alg1} 
\begin{algorithmic} 
    \State \textbf{Input:} Benign source face image $x_s$ and impersonated target face $x_t$, desired style text prompt $t$, generative model $G_L$, FR models $\mathcal{F}=\mathcal{F}_1,\dots,\mathcal{F}_T$, CLIP textual encoder $CLIP_t$, GAN inverter $Inv$ and maximum epoch numbers $K$.
    \State \textbf{Output:} Adversarial face image $\hat{x}$ with attribute in prompt $t$.
    \State \textbf{Initialization:} Initialize model parameters $\Theta_M$ MLFE and trade-off weights $\lambda _{guide}$ , $\lambda _{perc}$.
    
    \For{every $k$ in $K$} 
    \State Compute $\mathcal{L} _{guide}$ and $\mathcal{L} _{perc}$;
    
    \State \textbf{Meta-Train}:
    \State  Random select $T$ models from $\mathcal{F}$ as meta-train models;
    \State Compute $\mathcal{L}_i^{tr}$ with \refeq{Meta-Train} for the $i^{th}$ model and then update ${\Theta_{M}}_i'$ with \refeq{Theta_M};
    \State \textbf{Meta-Test}:
    \State Use the left FR as meta-test model;
    \State Compute $\mathcal{L}_i^{te}$ with \refeq{meta_test};
    \State \textbf{Meta-Optimization}:
    \State with \refeq{meta_fun};
    \EndFor
    \State \Return ${\Theta_M}^\ast$.
\end{algorithmic}
\end{algorithm}

\section{Implementation details}
\label{Implementation_details}
\subsection{Text prompts}
\begin{figure*}[!t]

\centering
\includegraphics[scale=1]{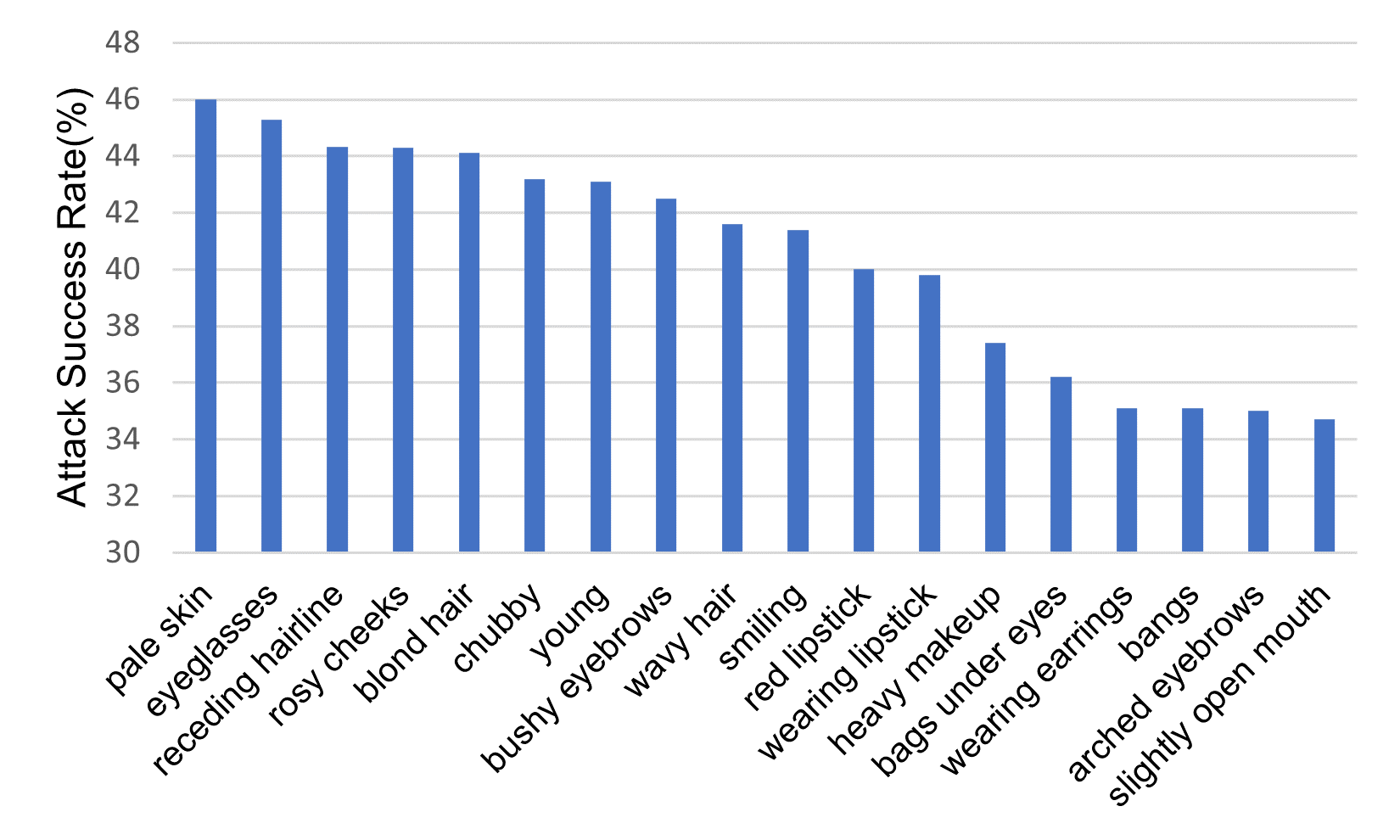}
\caption{The visualizations of the impact on the attack success rate with different facial attributes. The attacked FR model is FaceNet. All experiment results are benchmarked on the KID-F dataset. \label{different_attribute}}

\end{figure*}

We have collected 18 style text prompts from the Internet to guide the adversarial example generation. Details and the text prompts we used are provided in the \reftbl{tab:style_prompt}. We also report the result of the attack success rate with different facial attributes in \refig{different_attribute}.
\begin{table*}[]
\centering
\caption{Style prompt used in our experiments.\label{tab:style_prompt}}
\begin{tabular}{c | c | l}
\toprule[0.15em]
\rowcolor{mygray}  & \textbf{Style attribute description}  & \textbf{Text Prompts} \\ \midrule
\textbf{1} & red lipstick & a face with red lipstick. \\ 
\textbf{2} & blond hair & a face with blond hair. \\ 
\textbf{3} & wavy hair & a face with wavy hair. \\ 
\textbf{4} & young & a young face. \\
\textbf{5} & eyeglasses & a face with eyeglasses. \\
\textbf{6} & heavy makeup & a face with heavy makeup. \\
\textbf{7} & rosy cheeks & a face with rosy cheeks. \\
\textbf{8} & chubby & a chubby face. \\
\textbf{9} & slightly open mouth & a face with slightly open mouth. \\
\textbf{10} & bushy eyebrows & a face with bushy eyebrows. \\ 
\textbf{11} & wearing lipstick & a face wearing lipstick. \\ 
\textbf{12} & smiling & a smiling face. \\
\textbf{13} & arched eyebrows & a face with arched eyebrows. \\
\textbf{14} & bangs & a face with bangs. \\
\textbf{15} & wearing earrings & a face wearing earrings. \\
\textbf{16} & bags under eyes & a face with bags under eyes. \\
\textbf{17} & receding hairline & a face with receding hairline. \\
\textbf{18} & pale skin & a face with pale skin. \\
\midrule[0.15em]
\end{tabular}
\end{table*}
\subsection{Threshold $\tau$}
In Sec.4.1.4, $\tau$ indicates the threshold, $A S R$ measures the proportion of source-target pairs whose similarity scores exceed $\tau$ out of all source-target pairs. In our experiment, 
threshold value $\tau$ at 0.01 false acceptance rate for four FR models \textit{i.e}, MobileFace(0.302) \cite{chinaev2018mobileface}, IRSE50(0.241) \cite{8578843}, IR152(0.167) \cite{8953658}, and FaceNet(0.409) \cite{schroff2015facenet}.

\section{Experimental results}
\label{Experimental_res}
\subsection{Image quality assessment}
\begin{figure*}
     \centering
     \begin{subfigure}[b]{0.23\textwidth}
         \centering
         \includegraphics[width=\textwidth]{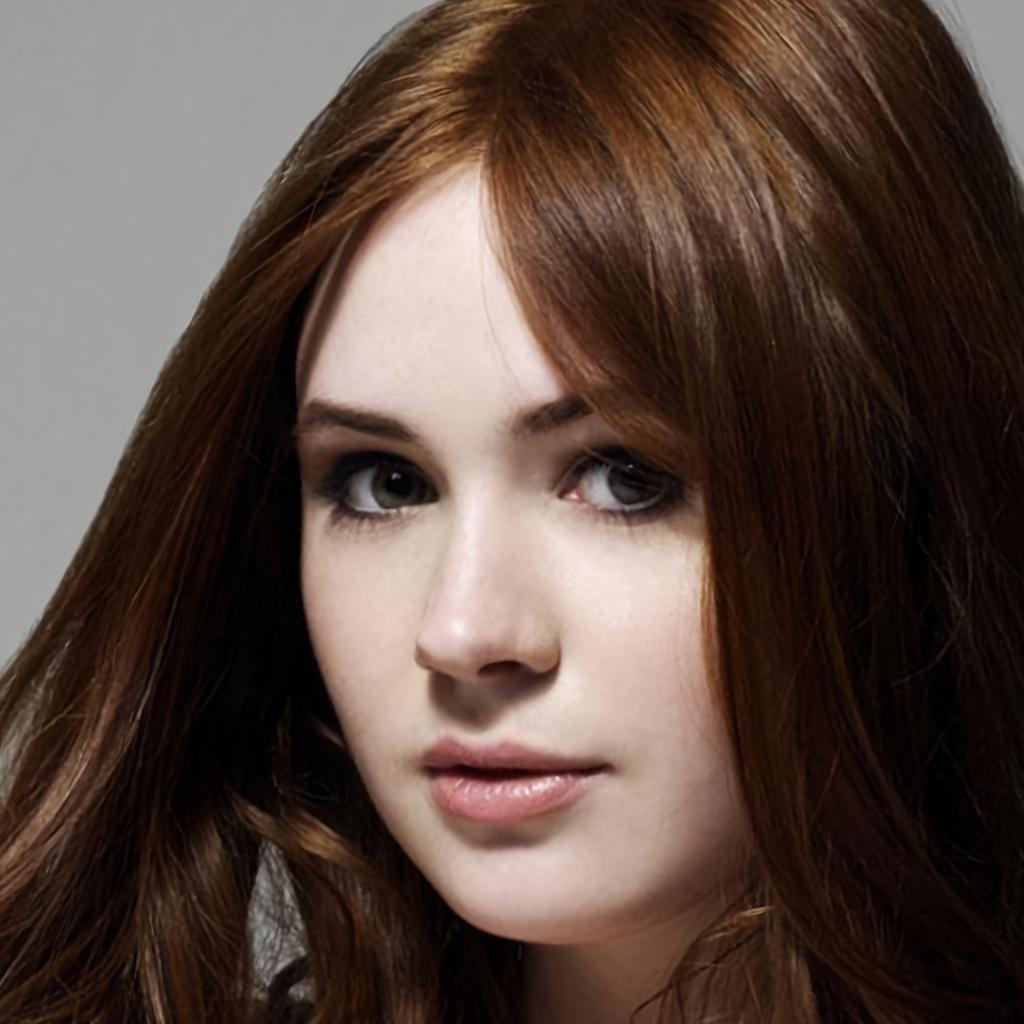}
         \caption{Original}
     \end{subfigure}
     \hfill
     \begin{subfigure}[b]{0.23\textwidth}
         \centering
         \includegraphics[width=\textwidth]{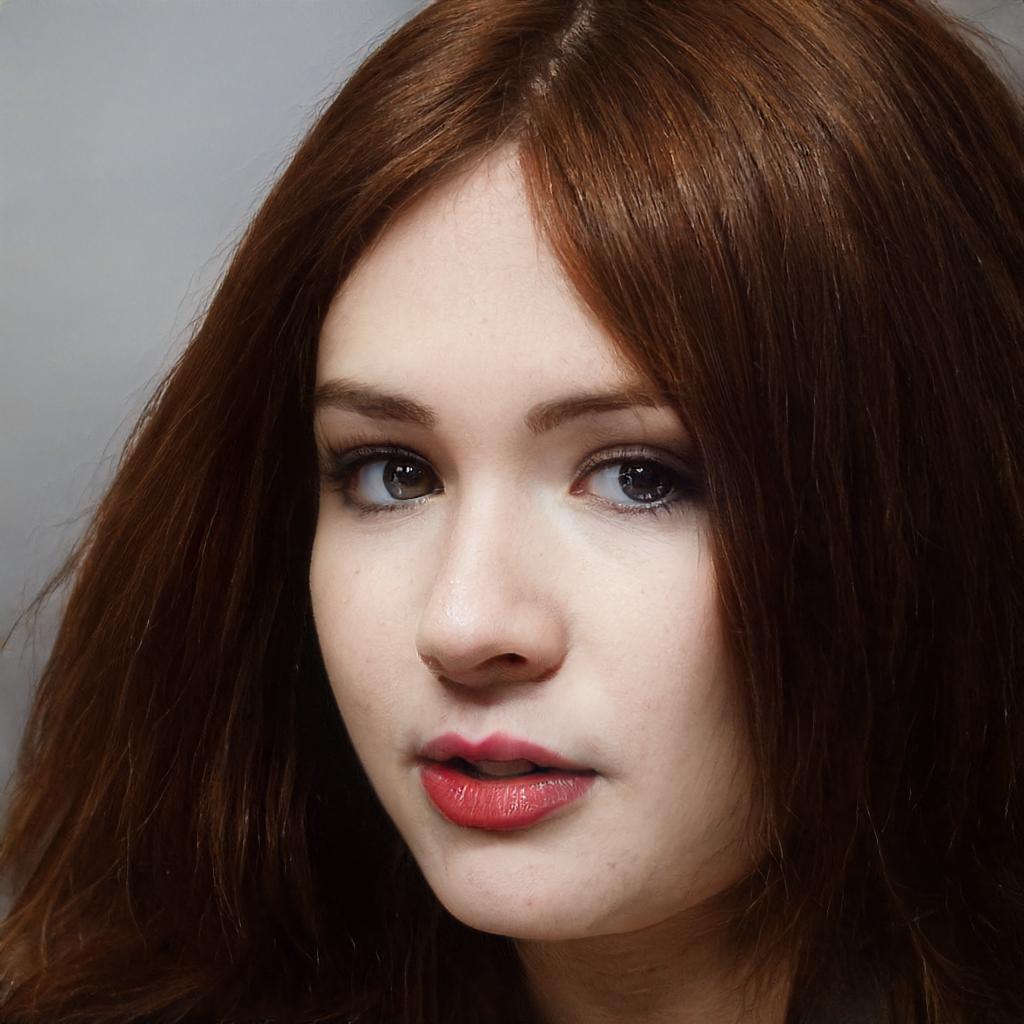}
         \caption{TCA$^2$}
     \end{subfigure}
     \hfill
     \begin{subfigure}[b]{0.23\textwidth}
         \centering
         \includegraphics[width=\textwidth]{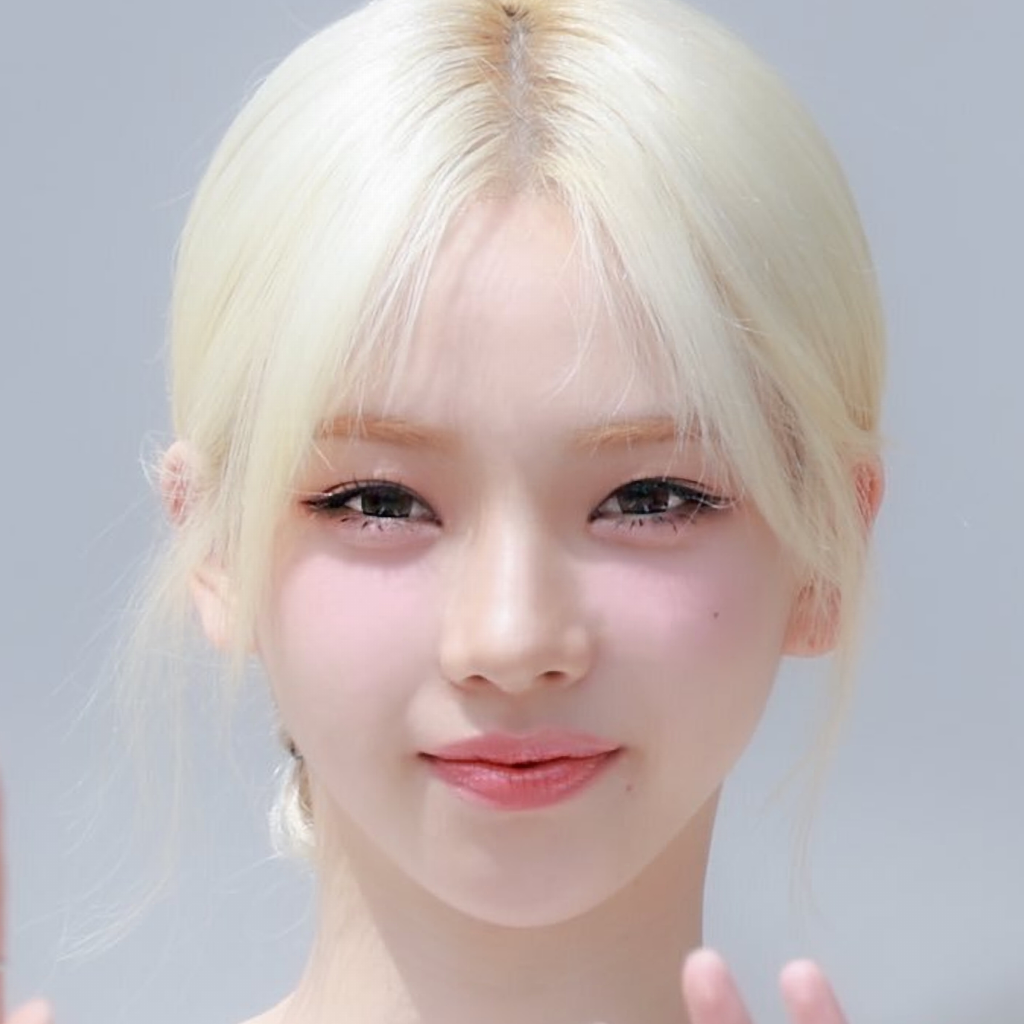}
         \caption{Original}
     \end{subfigure}
     \hfill
     \begin{subfigure}[b]{0.23\textwidth}
         \centering
         \includegraphics[width=\textwidth]{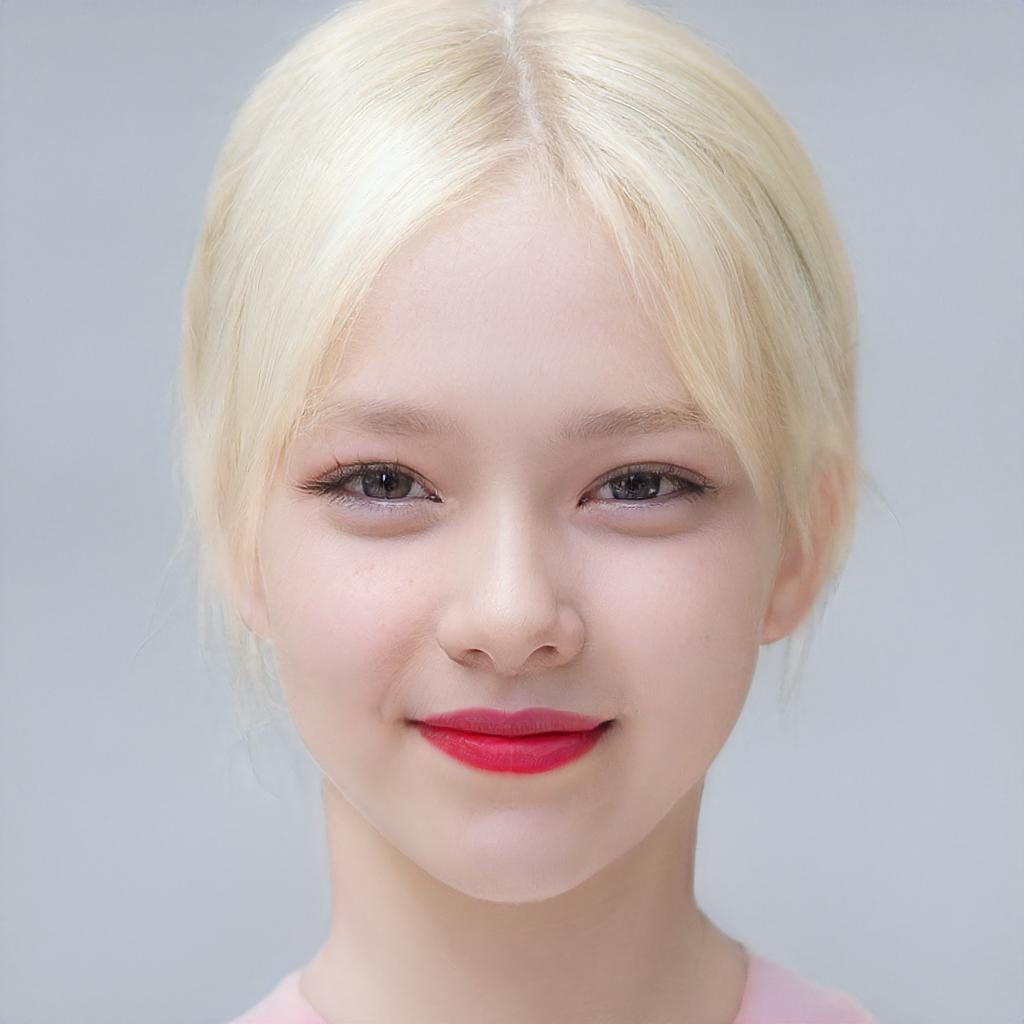}
         \caption{TCA$^2$}
     \end{subfigure}
     \vspace{-1em}
        \caption{The visualization of original face image and TCA$^2$ generated adversarial images. The given attribute is "\textit{red lipstick}" and its corresponding text prompt is "\textit{A face with red lipstick}".}
        \label{fig:red_lipstick}
\end{figure*}

\begin{figure*}
     \centering
     \begin{subfigure}[b]{0.23\textwidth}
         \centering
         \includegraphics[width=\textwidth]{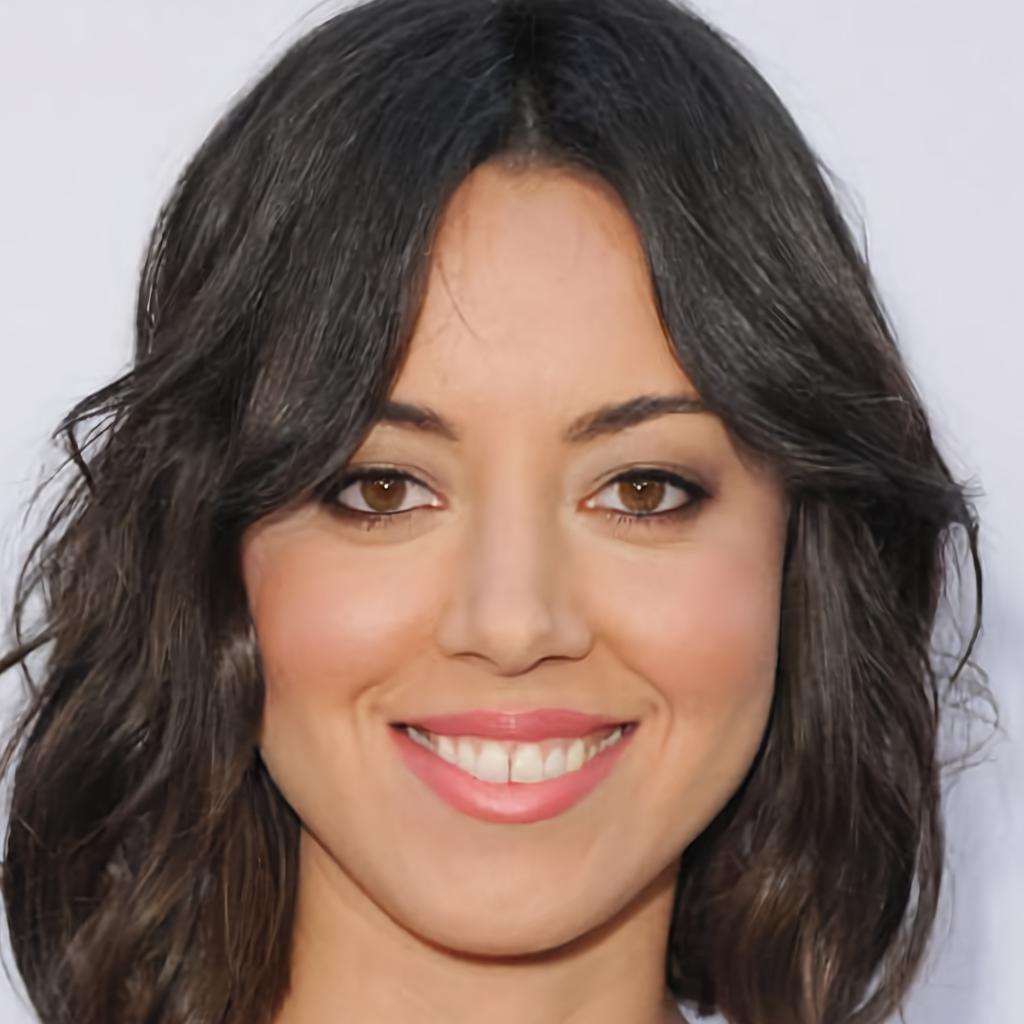}
         \caption{Original}
     \end{subfigure}
     \hfill
     \begin{subfigure}[b]{0.23\textwidth}
         \centering
         \includegraphics[width=\textwidth]{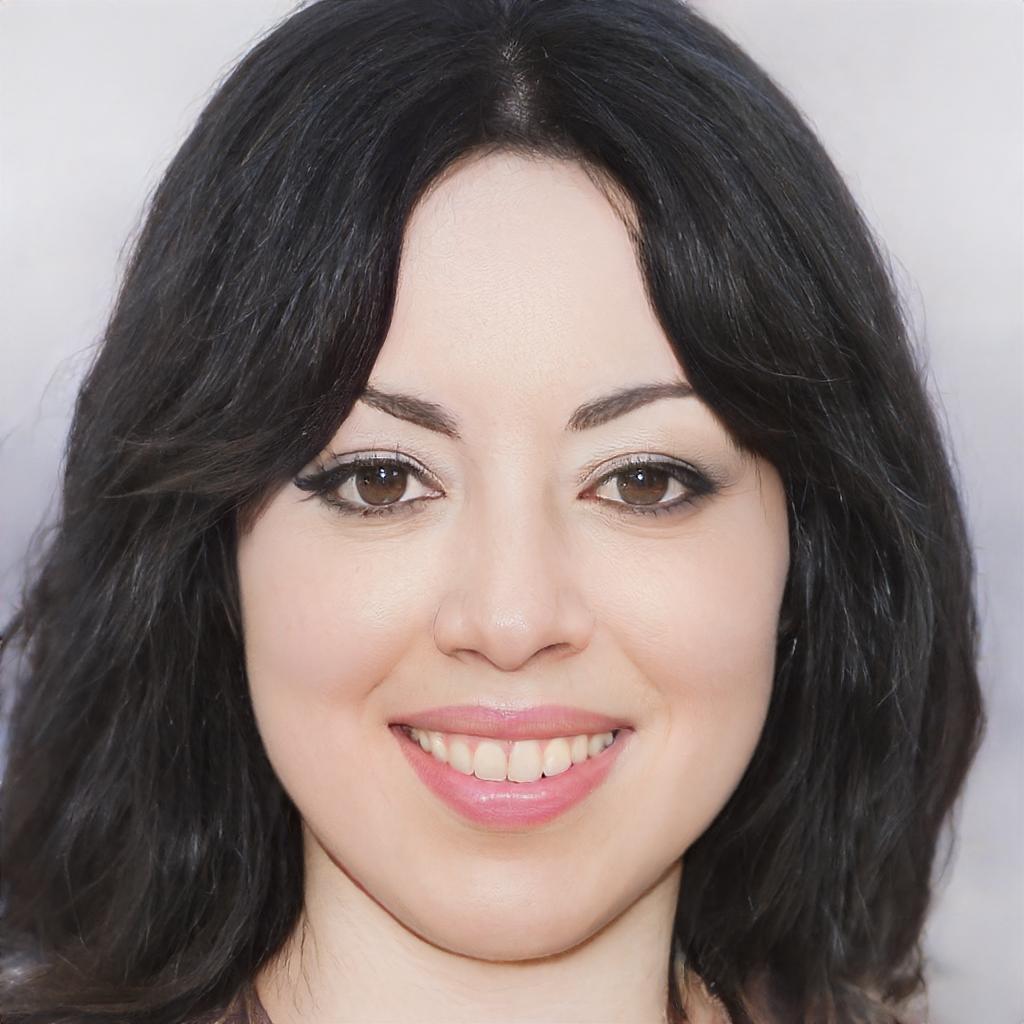}
         \caption{TCA$^2$}
     \end{subfigure}
     \hfill
     \begin{subfigure}[b]{0.23\textwidth}
         \centering
         \includegraphics[width=\textwidth]{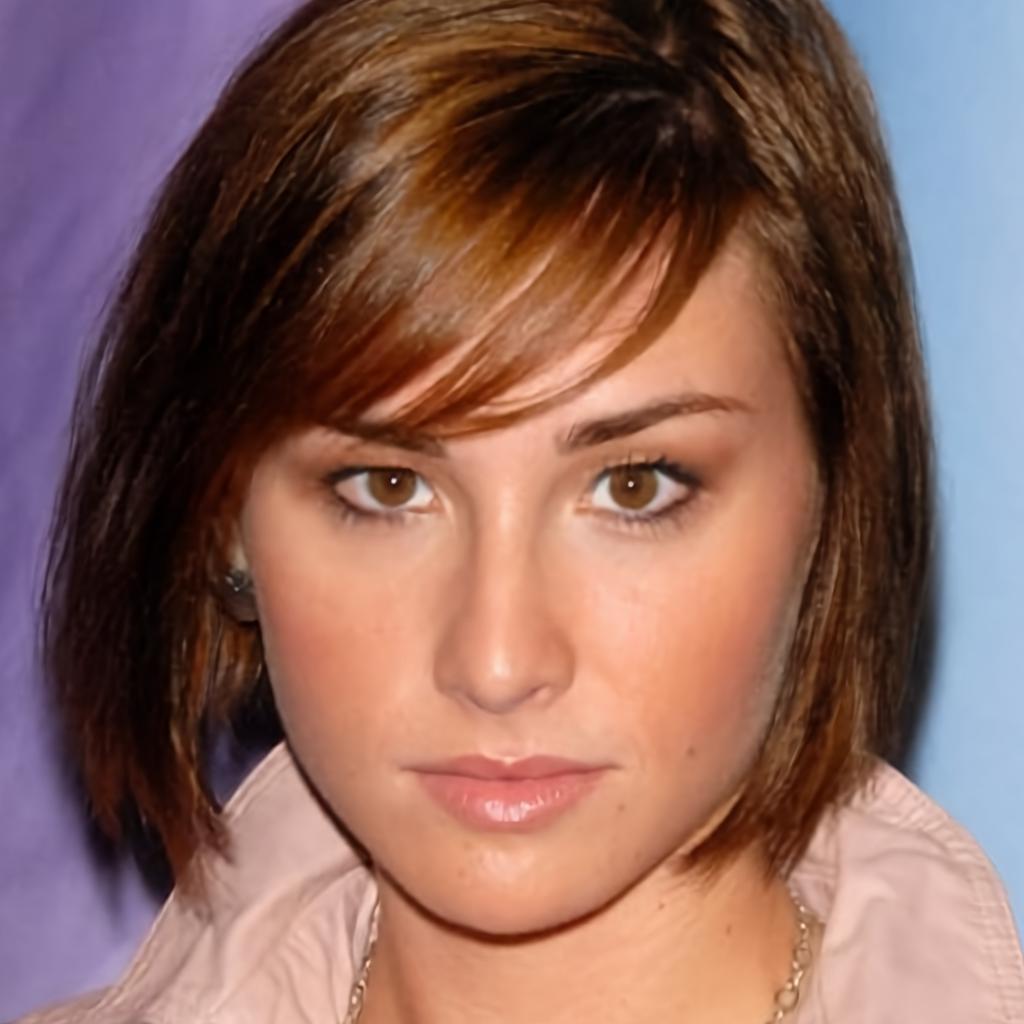}
         \caption{Original}
     \end{subfigure}
     \hfill
     \begin{subfigure}[b]{0.23\textwidth}
         \centering
         \includegraphics[width=\textwidth]{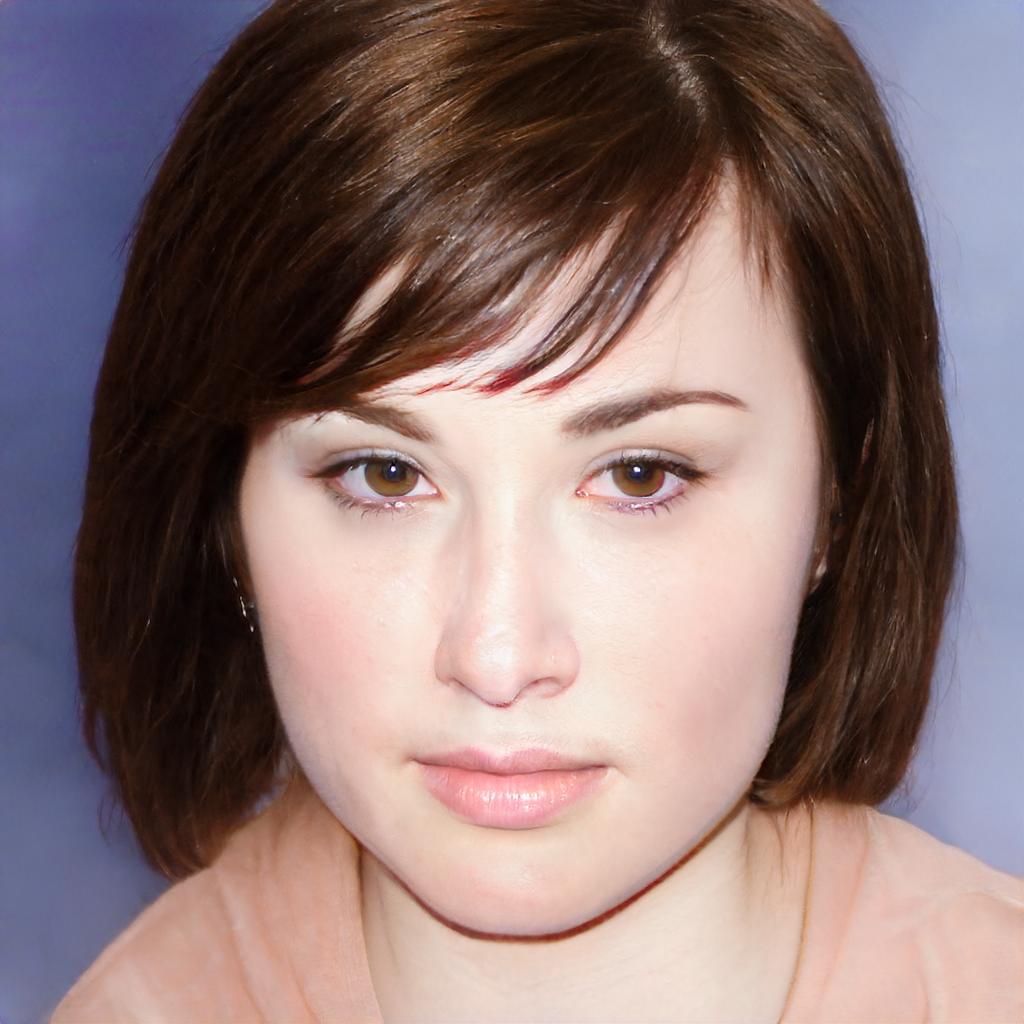}
         \caption{TCA$^2$}
     \end{subfigure}
     \vspace{-1em}
        \caption{The visualization of original face image and TCA$^2$ generated adversarial images. The given attribute is "\textit{pale skin}" and its corresponding text prompt is "\textit{A face with pale skin}".}
        \label{fig:pale_skin}
\end{figure*}

In Sec. 4.4.2, the FID scores of TCA$^2$ on two datasets is reported in \reftbl{table:fid}. While the PSNR and SSIM results are shown in \reftbl{table:ssim}. Moreover, we also provide results of TCA$^2$ in attacking the state-of-the-art FR model under the different text prompts' guidance. The visualization is shown from \refig{fig:red_lipstick} to \refig{fig:pale_skin}.
\begin{table}
\small
	\centering
 \caption{ FID scores of different attack methods against FR models on the CelebA-Identity and KID-F datasets.\label{table:fid}}
	\begin{tabular}{l | c  |c }
\toprule[0.15em]
\rowcolor{mygray} \textbf{Method} & CelebA-Identity $\downarrow$  & KID-F $\downarrow$ \\
\midrule[0.15em]
Adv-Hat~  & 113.58  & 138.22\\
Adv-Attribute~ & 44.23  &  50.82    \\
Latent-HSJA~ & 47.86 &  45.30 \\
Adv-Diffusion~  & 25.71  & 33.37\\
\midrule
\rowcolor{orange!6} TCA$^2$ & 26.62 & 32.29 \\
\bottomrule[0.1em]
\end{tabular}
\end{table}
\begin{table}[]
\small
	\centering
 \caption{PSNR and SSIM of different attack methods against FR models on the CelebA-Identity and KID-F datasets.\label{table:ssim}}
	\begin{tabular}{l || c c  || c c   }
\toprule[0.15em]
\rowcolor{mygray} \textbf{Method} & \multicolumn{2}{c||}{\textbf{CelebA-Identity}}&\multicolumn{2}{c}{\textbf{KID-F}} \\
\rowcolor{mygray}  & PSNR $\uparrow$ & SSIM $\uparrow$ & PSNR $\uparrow$ & SSIM $\uparrow$  \\
\midrule[0.15em]
Adv-Hat & 14.31  & 0.55  & 15.50  & 0.52     \\
Adv-Attribute & 28.70 & 0.66 & 20.92 & 0.68   \\
Latent-HSJA & 22.70  & 0.72  & 25.77    & 0.75       \\
Adv-Diffusion~ & 33.32  & 0.80  & 32.04   & 0.79       \\
\midrule
\rowcolor{orange!6} TCA$^2$ & 33.87 &0.81 & 34.76 & 0.82   \\
\bottomrule[0.1em]
\end{tabular}
\end{table}
\subsection{Ablation study}
\begin{figure*}[!t]
    \centering
     \begin{subfigure}[b]{0.4\textwidth}
         \centering
         \includegraphics[width=\textwidth]{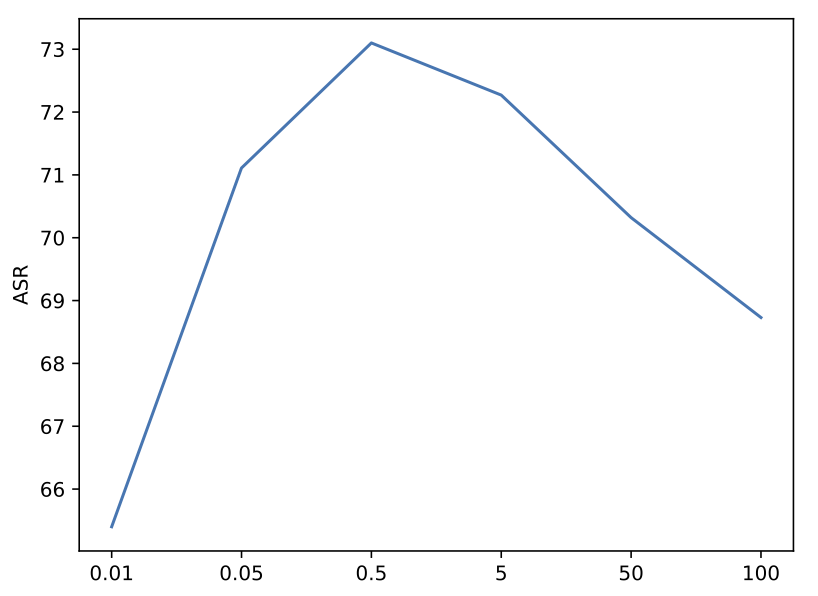}
         \caption{The impact of different $\lambda _{guide}$ on ASR.}
         \label{fig:guide}
     \end{subfigure}
     
     \begin{subfigure}[b]{0.4\textwidth}
         \centering
         \includegraphics[width=\textwidth]{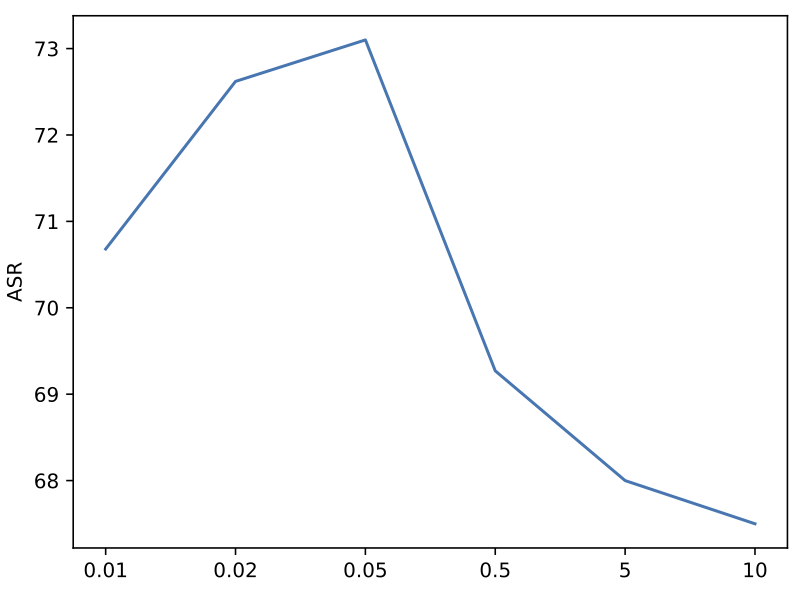}
         \caption{The impact of different $\lambda _{perc}$ on ASR.}
         \label{fig:perp}
     \end{subfigure}
\end{figure*}
\begin{table*}[!p]
\caption{The ablation study of our TCA$^2$ on transferablity. The experiments are conducted on both CelebA-Identity and KID-F datasets. \label{ablation}}
\centering
\begin{tabular}{l || c c  || c c   }
\toprule[0.15em]
\rowcolor{mygray} \textbf{Variants} & \multicolumn{2}{c||}{\textbf{CelebA-Identity}}&\multicolumn{2}{c}{\textbf{KID-F}} \\
\rowcolor{mygray}  & MobileFace & FaceNet & MobileFace &FaceNet  \\
\midrule[0.15em]
w/o data augmentation & 64.31  & 40.55  & 55.50  & 36.77     \\
w/o model augmentation & 14.31  & 20.72  & 15.76  & 10.22     \\
\midrule
\rowcolor{orange!6} TCA$^2$(Full) & 73.10 &42.26 & 65.57 & 39.64   \\
\bottomrule[0.1em]
\end{tabular}
\end{table*}
\begin{enumerate}
    \item \textbf{Effect of $\lambda _{guide}$ and $\lambda _{perc}$} Besides the experiment result in Sec.4.2.4 of the main paper, we also further conducted hyperparameter tuning experiments of the weights of text-guided controlling loss $\mathcal{L} _{guide}$ and identity preservation loss $\mathcal{L} _{perc}$. The result is shown in \refig{fig:guide} and \refig{fig:perp}. The experiments are conducted on the CelebA-Identity dataset.
    
    \item \textbf{Impact of transferablity}
    In order to support our contribution, we have conducted experiments to investigate the impact varables of transferablity. Specifically, the data augmentation and model augmentation is involved. Such result is shown in \reftbl{ablation}. From \reftbl{ablation} we can observe that data augmentation is very helpful in enhancing the transferablity of our TCA$^2$. Moreover, model augmentation has a bigger impact on the transferablity in our experiments.
    
\end{enumerate}

\begin{figure*}
     \centering
     \begin{subfigure}[b]{0.23\textwidth}
         \centering
         \includegraphics[width=\textwidth]{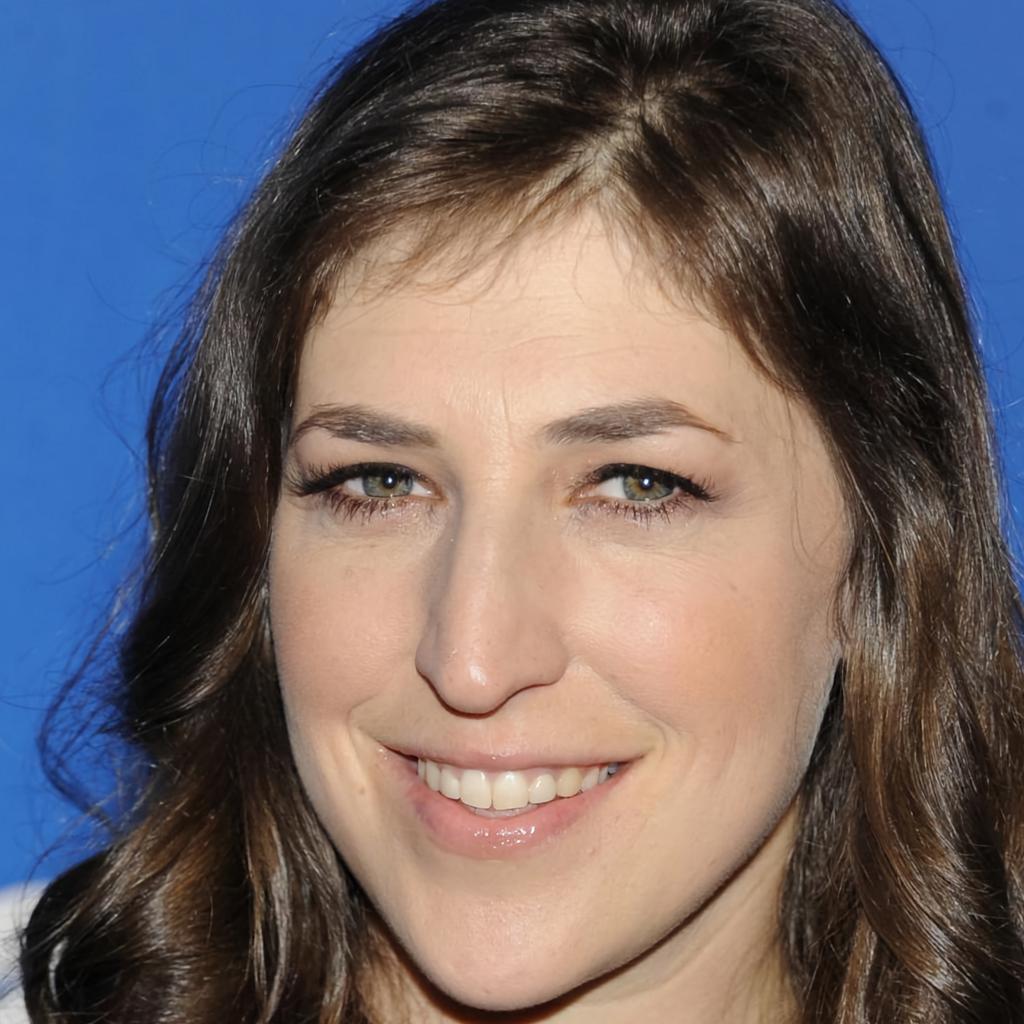}
         \caption{Original}
     \end{subfigure}
     \hfill
     \begin{subfigure}[b]{0.23\textwidth}
         \centering
         \includegraphics[width=\textwidth]{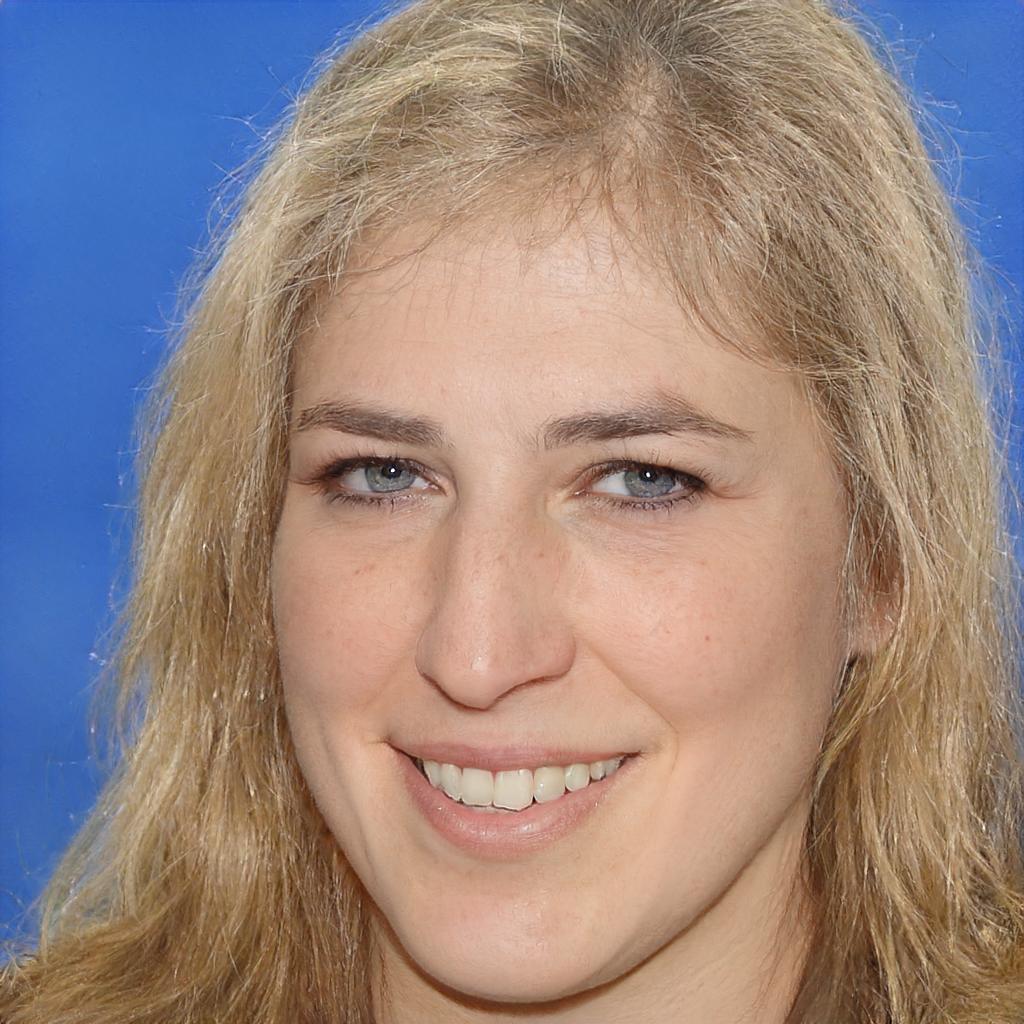}
         \caption{TCA$^2$}
     \end{subfigure}
     \hfill
     \begin{subfigure}[b]{0.23\textwidth}
         \centering
         \includegraphics[width=\textwidth]{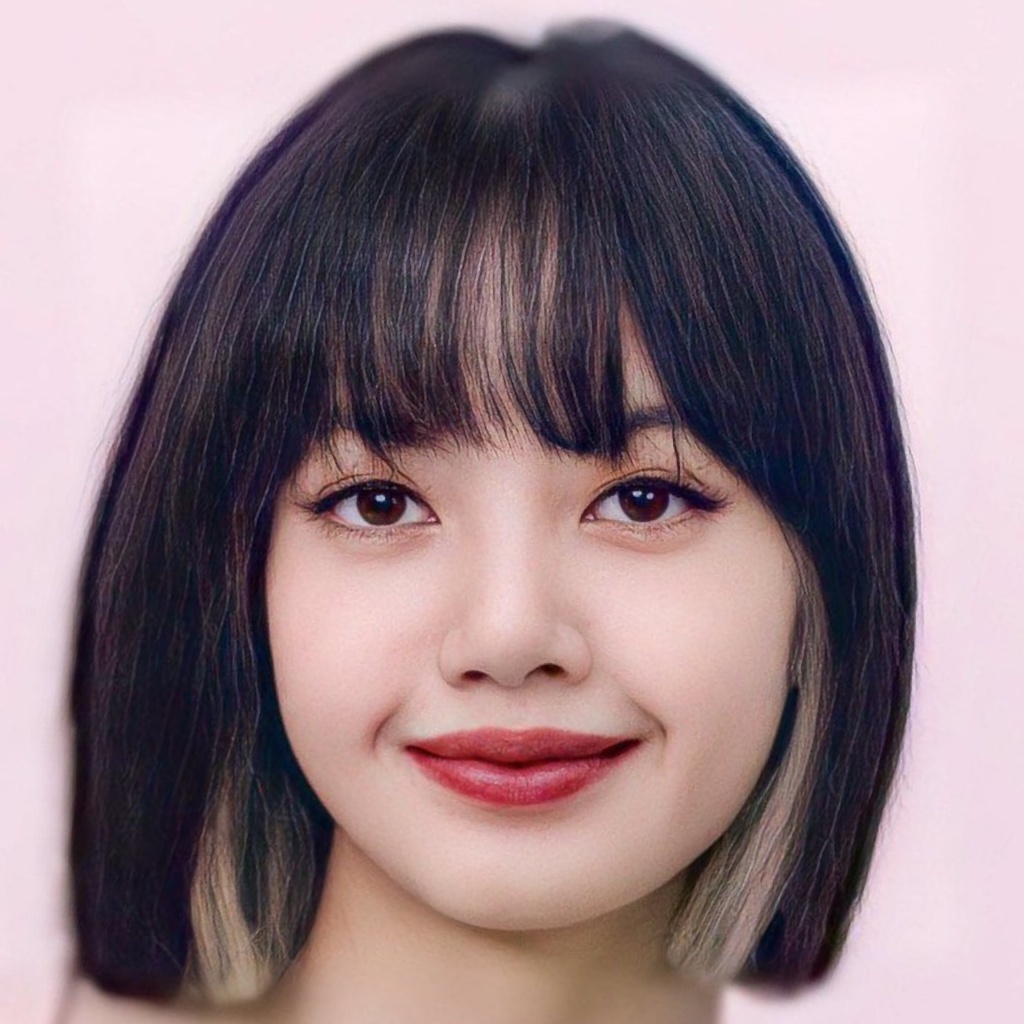}
         \caption{Original}
     \end{subfigure}
     \hfill
     \begin{subfigure}[b]{0.23\textwidth}
         \centering
         \includegraphics[width=\textwidth]{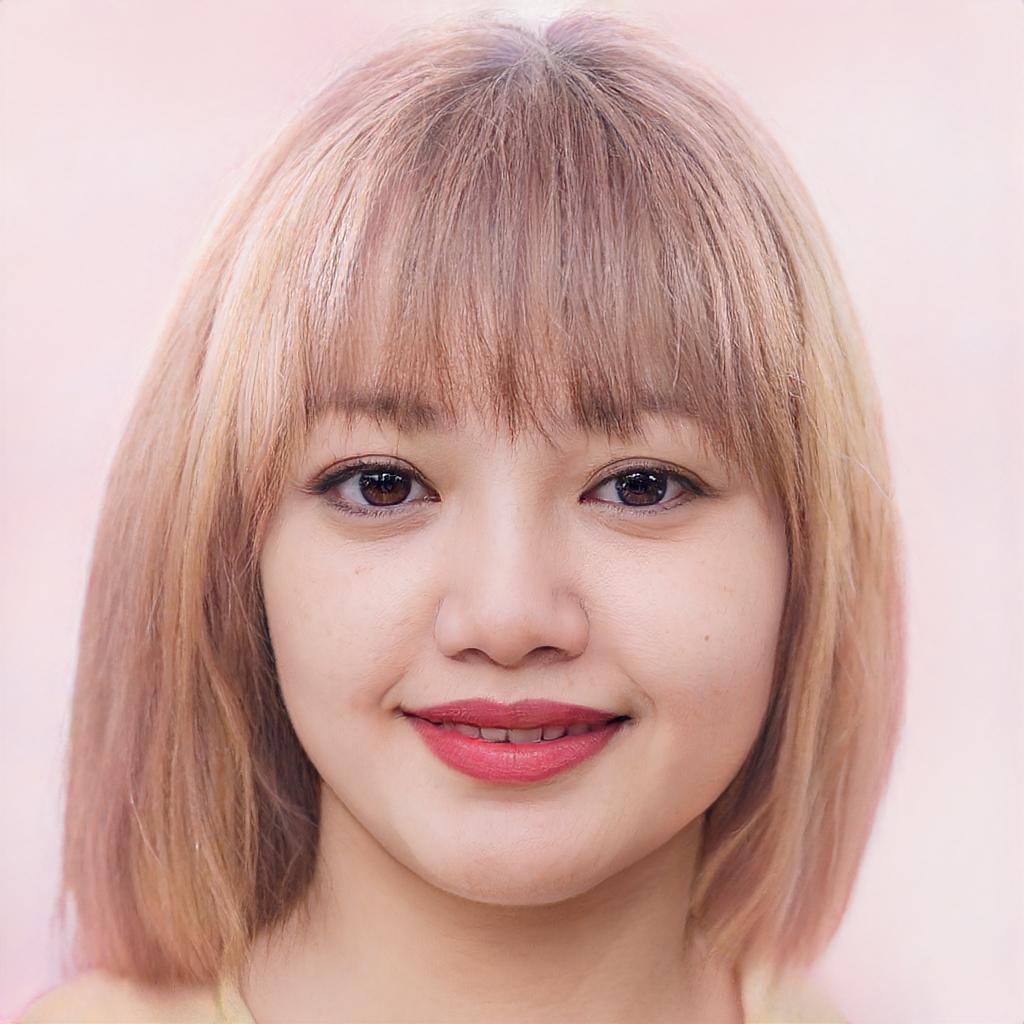}
         \caption{TCA$^2$}
     \end{subfigure}
     \vspace{-1em}
        \caption{The visualization of original face image and TCA$^2$ generated adversarial images. The given attribute is "\textit{blond hair}" and its corresponding text prompt is "\textit{A face with blond hair}".}
        
\end{figure*}
\begin{figure*}
     \centering
     \begin{subfigure}[b]{0.23\textwidth}
         \centering
         \includegraphics[width=\textwidth]{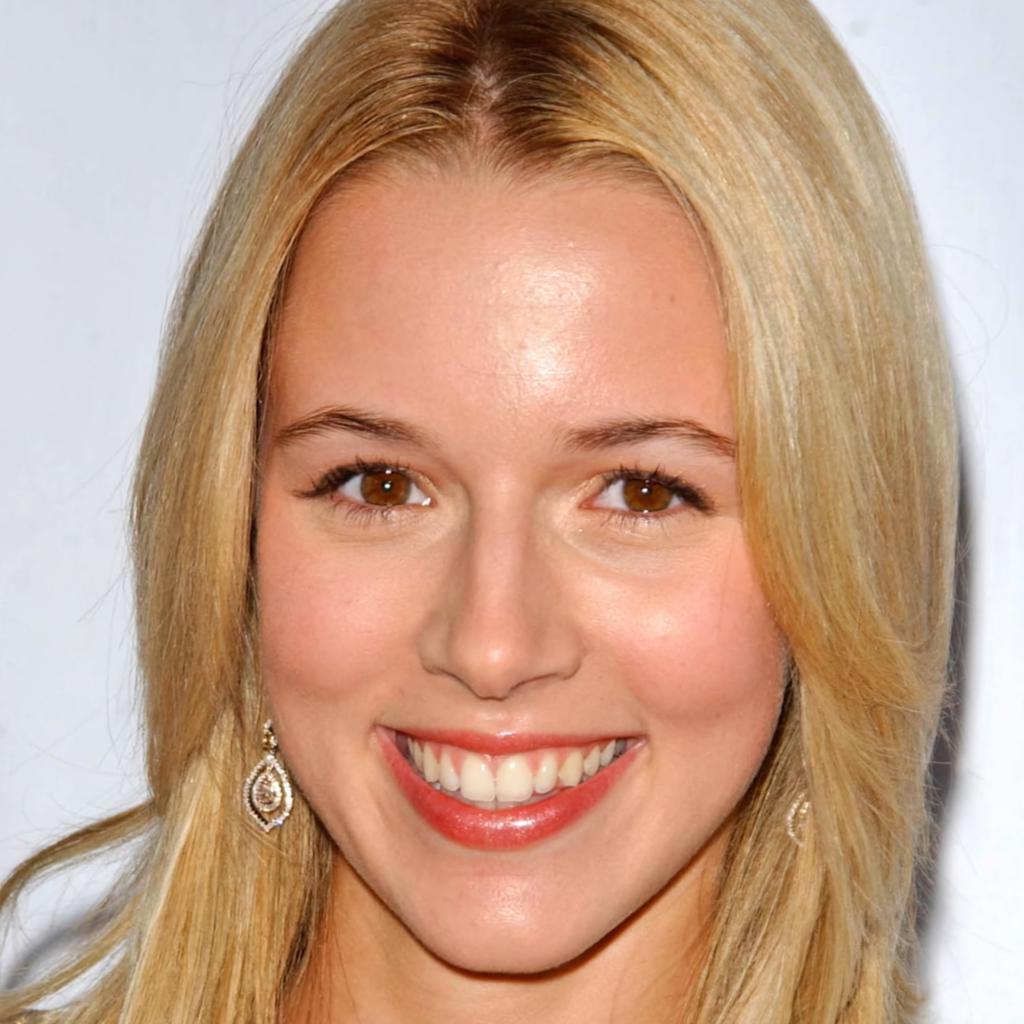}
         \caption{Original}
     \end{subfigure}
     \hfill
     \begin{subfigure}[b]{0.23\textwidth}
         \centering
         \includegraphics[width=\textwidth]{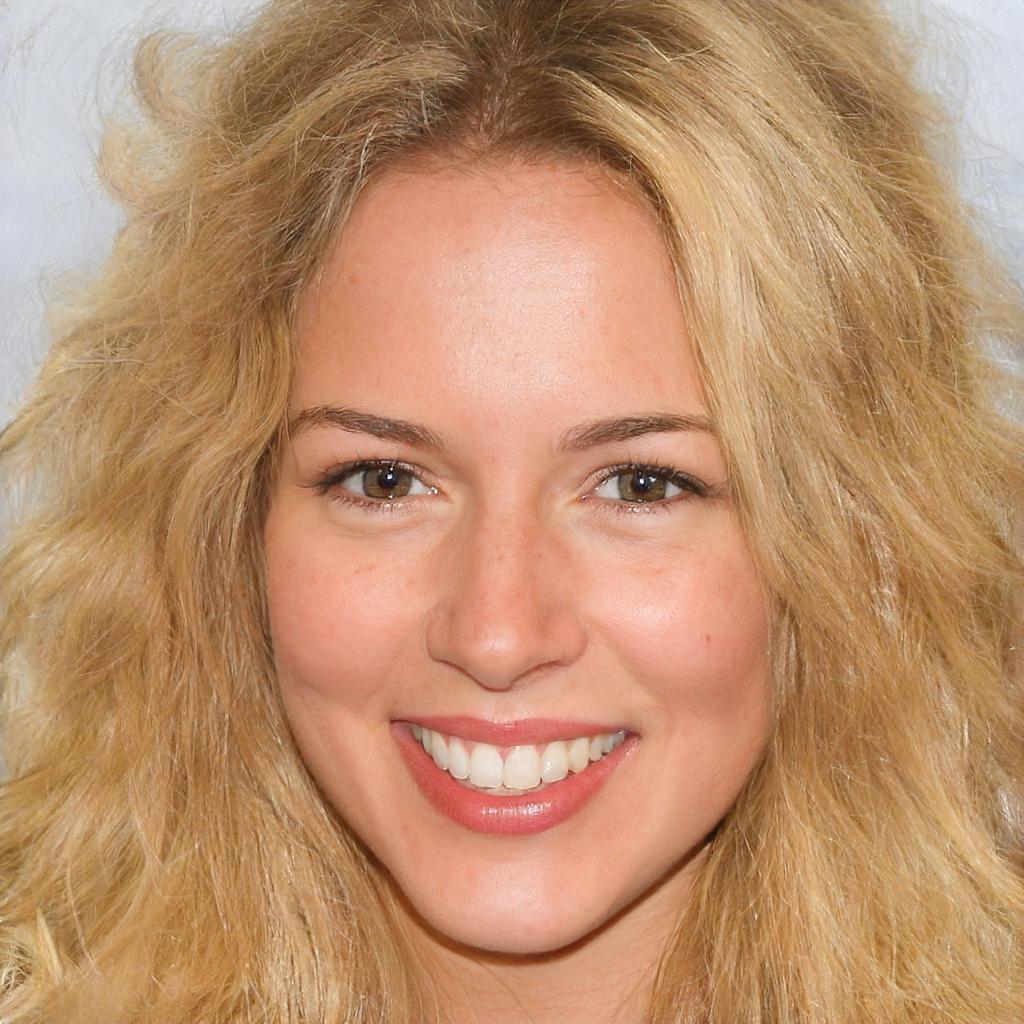}
         \caption{TCA$^2$}
     \end{subfigure}
     \hfill
     \begin{subfigure}[b]{0.23\textwidth}
         \centering
         \includegraphics[width=\textwidth]{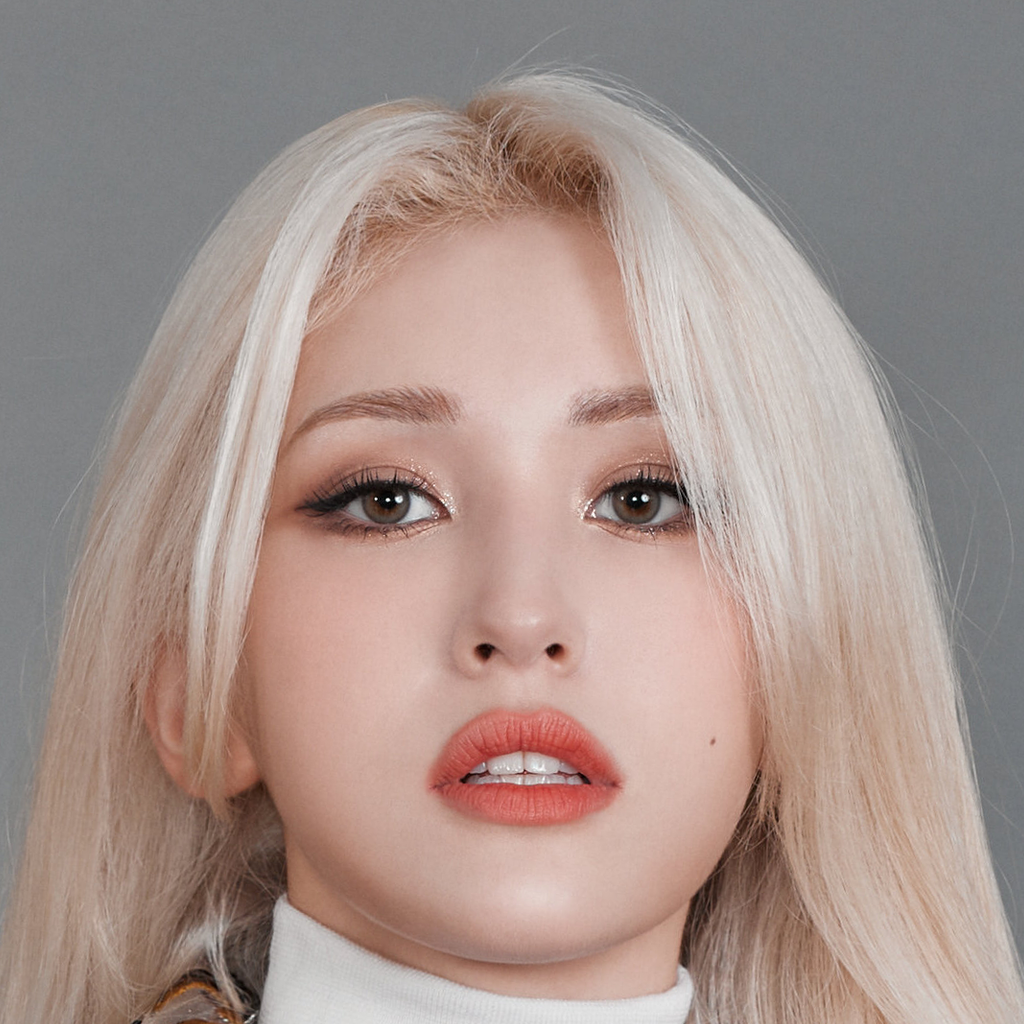}
         \caption{Original}
     \end{subfigure}
     \hfill
     \begin{subfigure}[b]{0.23\textwidth}
         \centering
         \includegraphics[width=\textwidth]{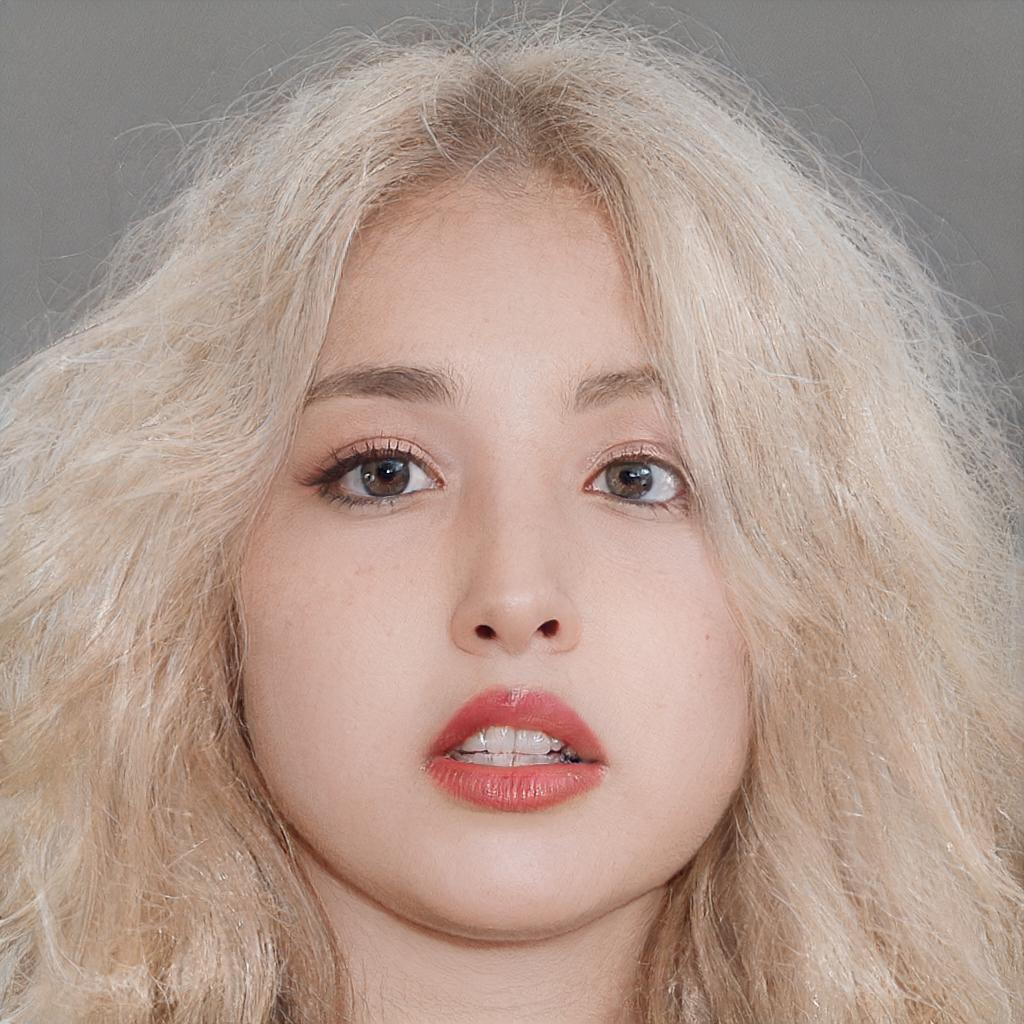}
         \caption{TCA$^2$}
     \end{subfigure}
     \vspace{-1em}
        \caption{The visualization of original face image and TCA$^2$ generated adversarial images. The given attribute is "\textit{wavy hair}" and its corresponding text prompt is "\textit{A face with wavy hair}".}
        
\end{figure*}
\begin{figure*}
     \centering
     \begin{subfigure}[b]{0.23\textwidth}
         \centering
         \includegraphics[width=\textwidth]{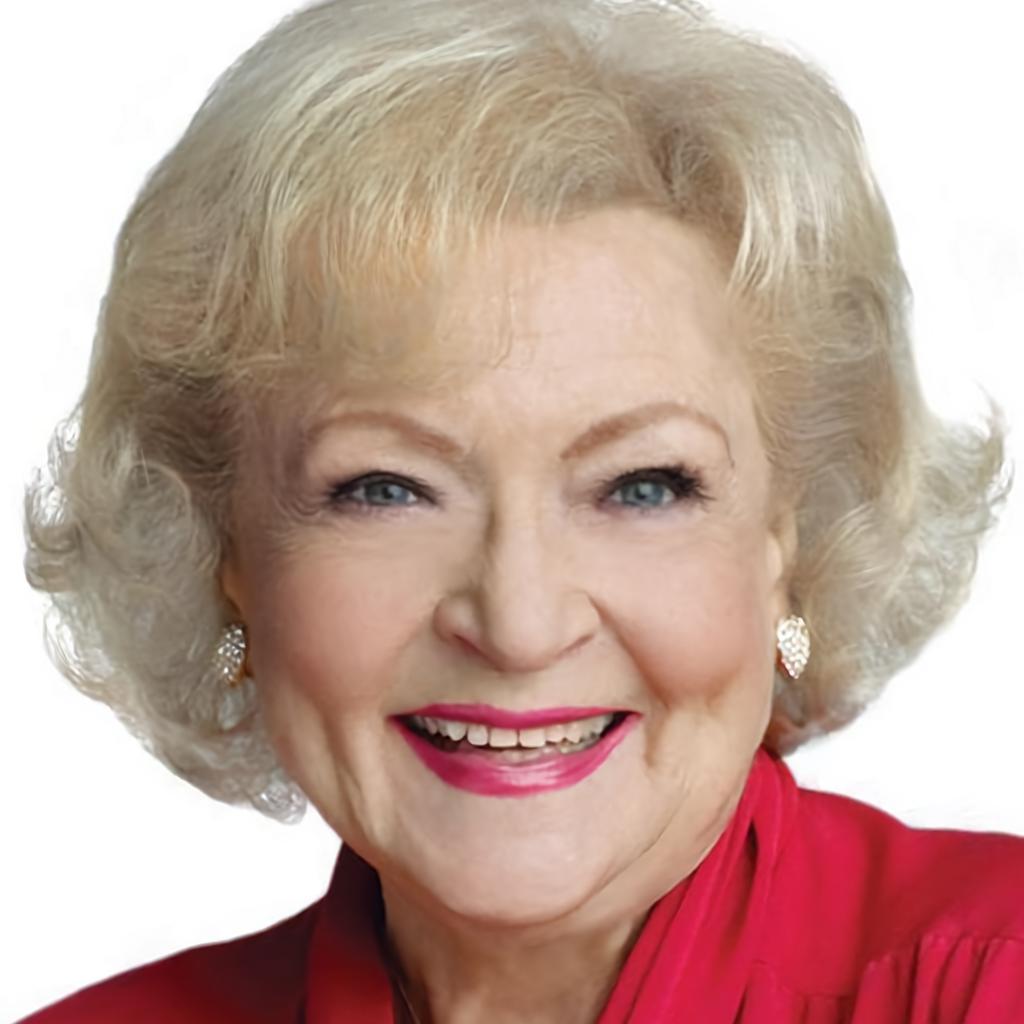}
         \caption{Original}
     \end{subfigure}
     \hfill
     \begin{subfigure}[b]{0.23\textwidth}
         \centering
         \includegraphics[width=\textwidth]{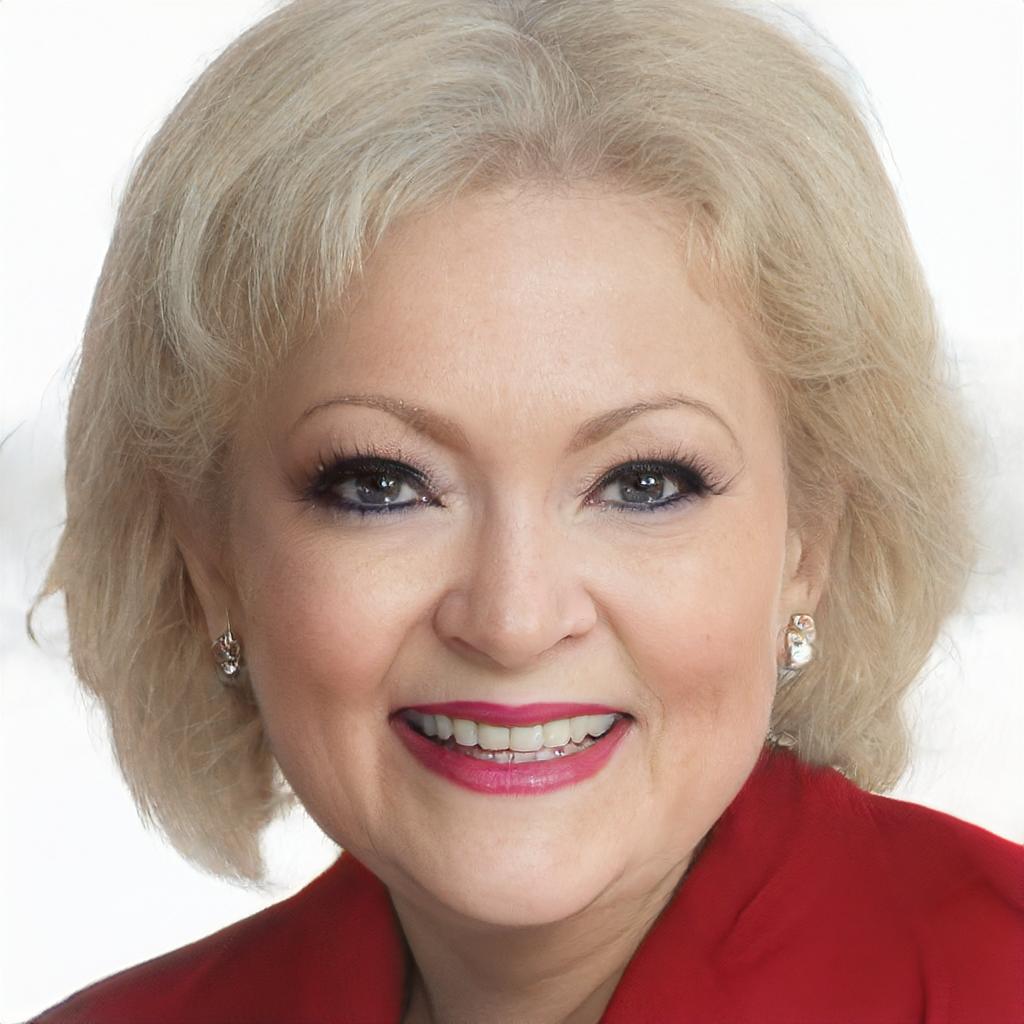}
         \caption{TCA$^2$}
     \end{subfigure}
     \hfill
     \begin{subfigure}[b]{0.23\textwidth}
         \centering
         \includegraphics[width=\textwidth]{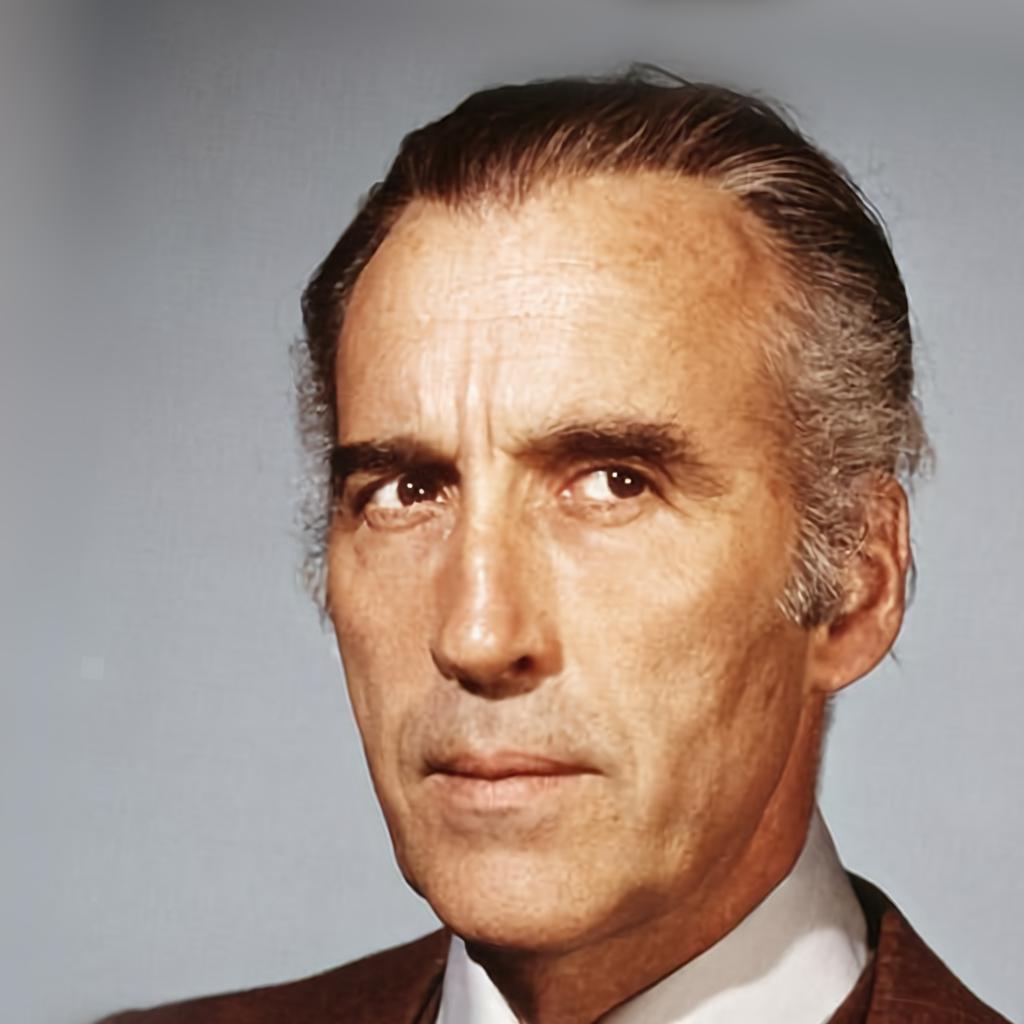}
         \caption{Original}
     \end{subfigure}
     \hfill
     \begin{subfigure}[b]{0.23\textwidth}
         \centering
         \includegraphics[width=\textwidth]{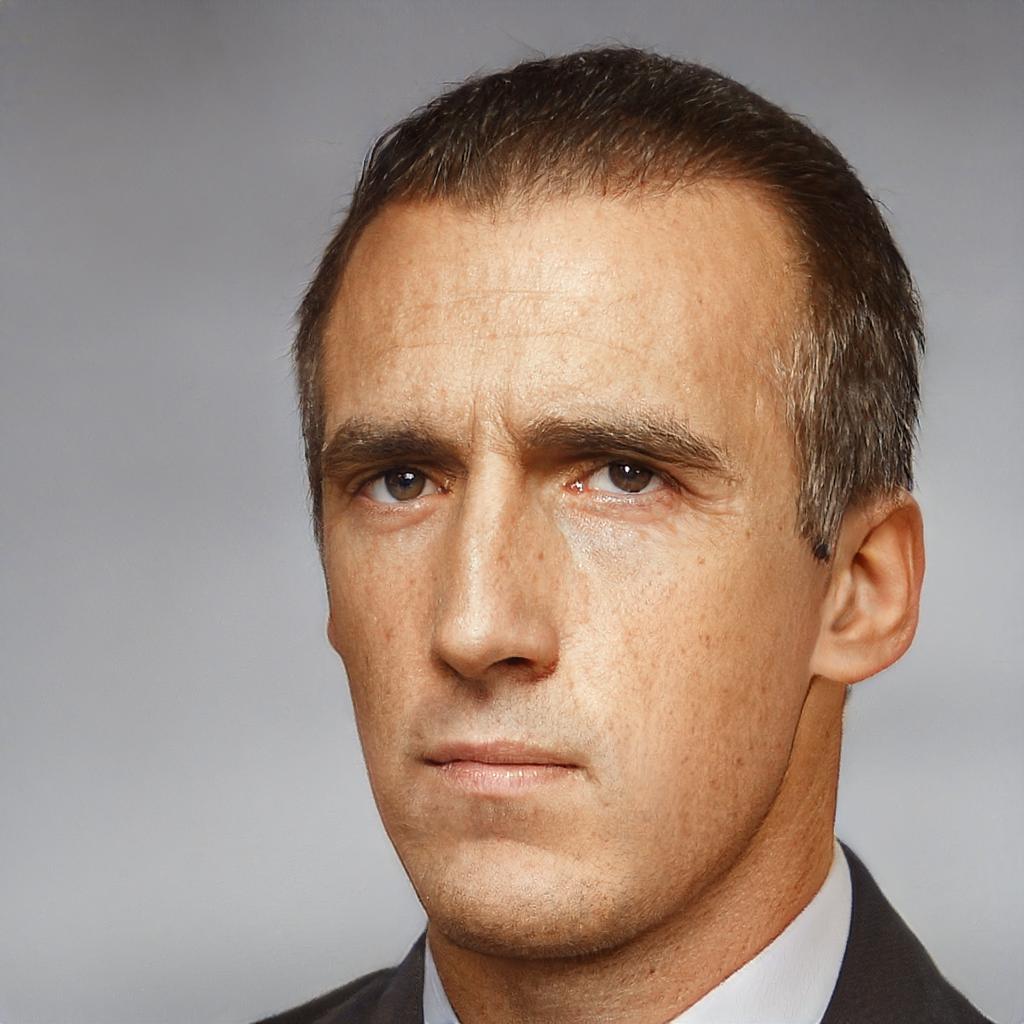}
         \caption{TCA$^2$}
     \end{subfigure}
     \vspace{-1em}
        \caption{The visualization of original face image and TCA$^2$ generated adversarial images. The given attribute is "\textit{young}" and its corresponding text prompt is "\textit{A young face}".}
        
\end{figure*}
\begin{figure*}
     \centering
     \begin{subfigure}[b]{0.23\textwidth}
         \centering
         \includegraphics[width=\textwidth]{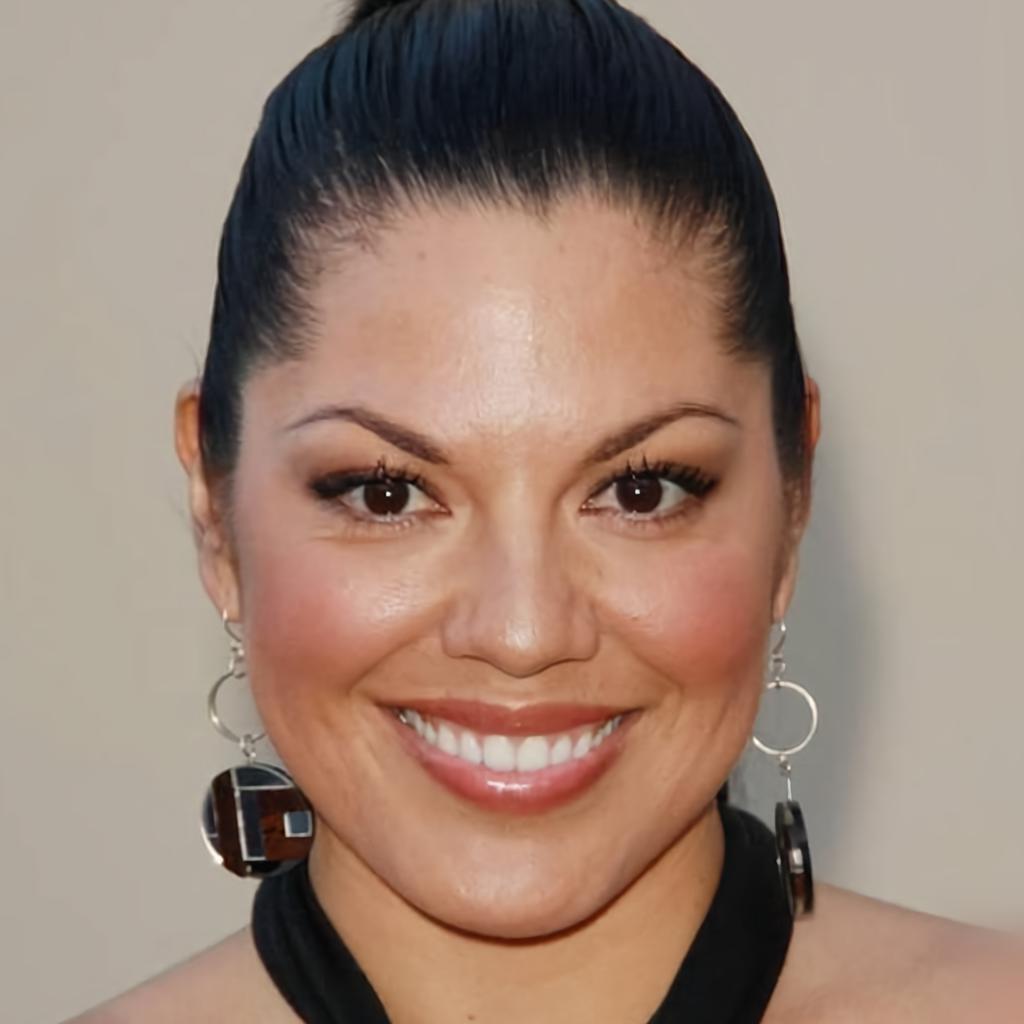}
         \caption{Original}
     \end{subfigure}
     \hfill
     \begin{subfigure}[b]{0.23\textwidth}
         \centering
         \includegraphics[width=\textwidth]{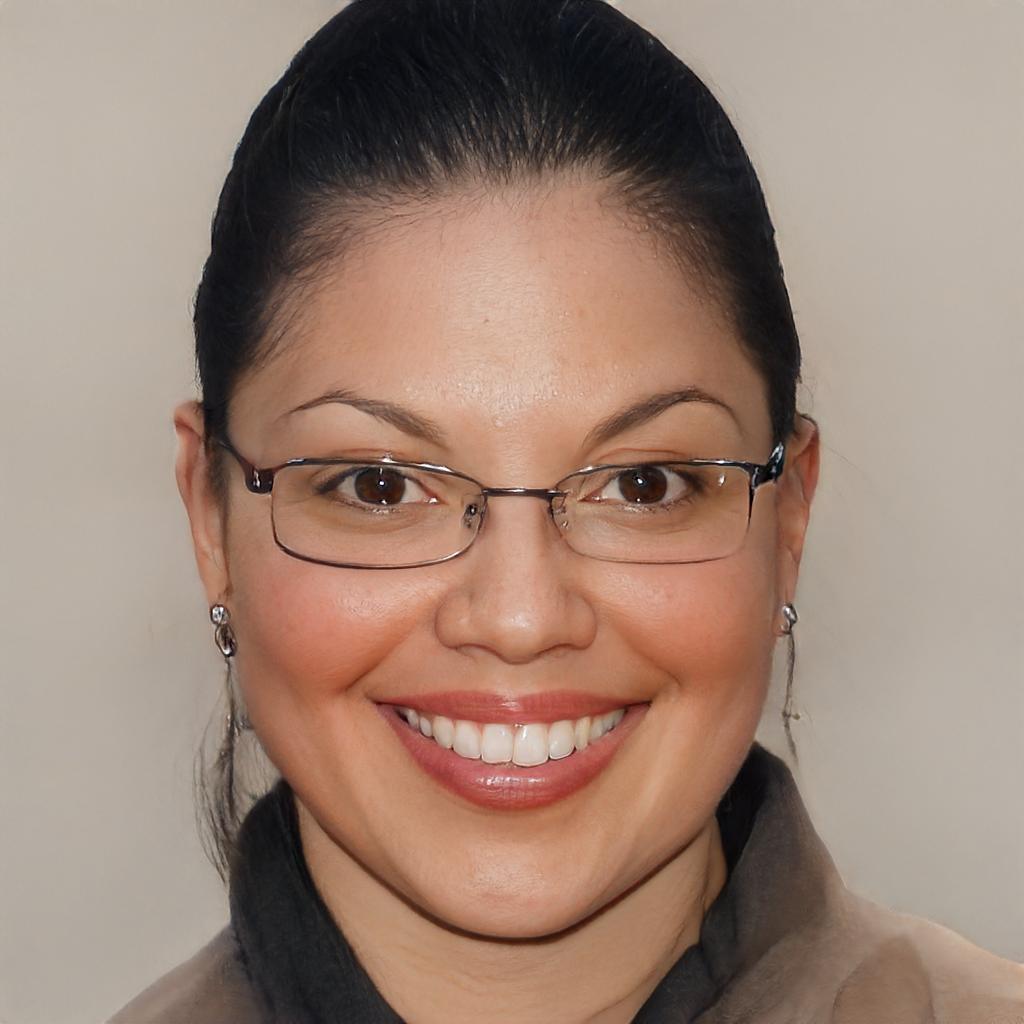}
         \caption{TCA$^2$}
     \end{subfigure}
     \hfill
     \begin{subfigure}[b]{0.23\textwidth}
         \centering
         \includegraphics[width=\textwidth]{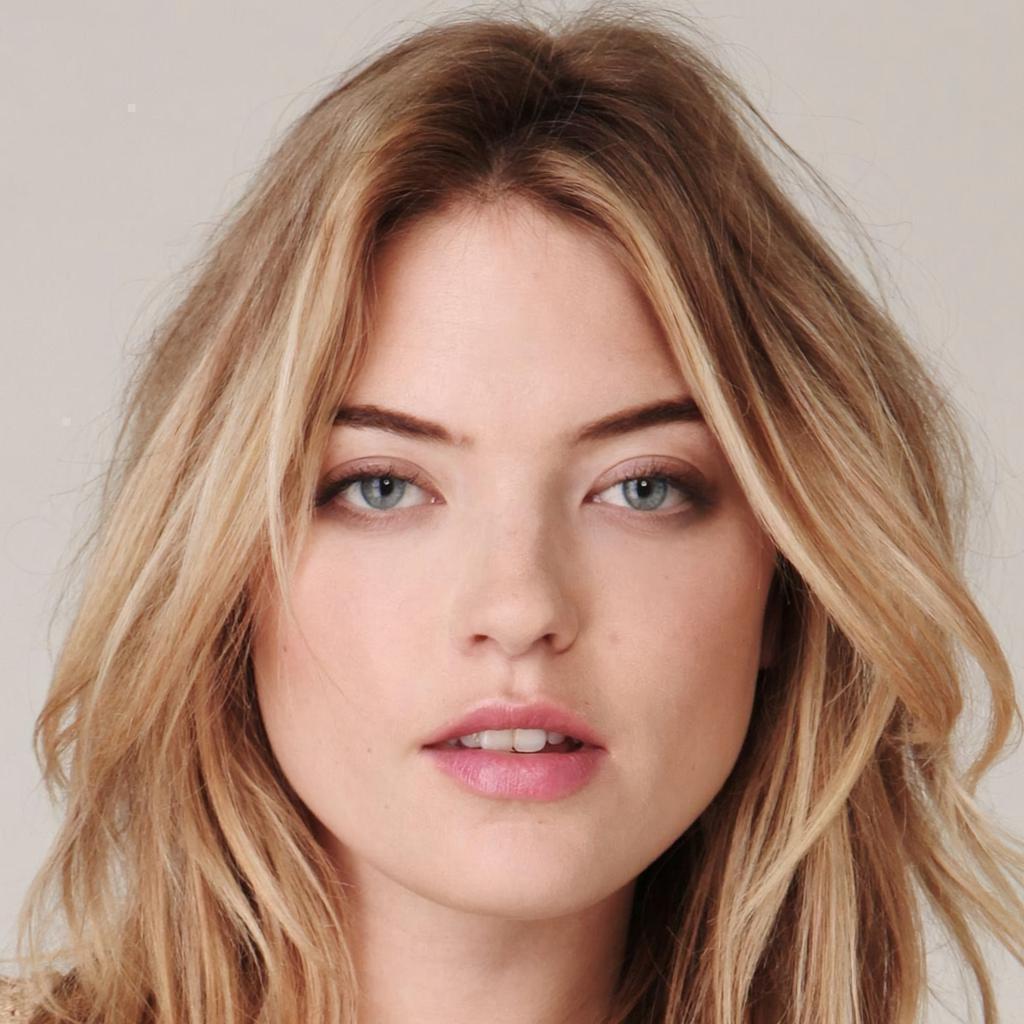}
         \caption{Original}
     \end{subfigure}
     \hfill
     \begin{subfigure}[b]{0.23\textwidth}
         \centering
         \includegraphics[width=\textwidth]{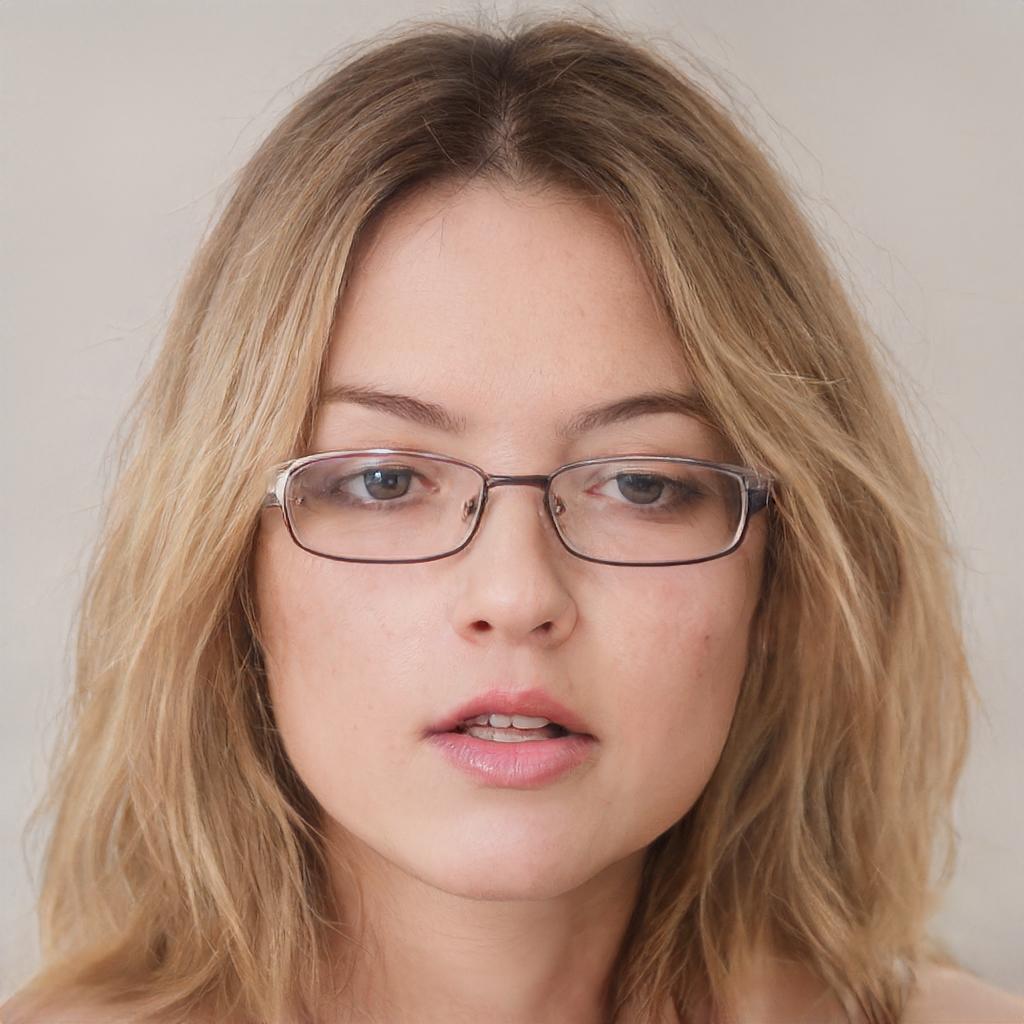}
         \caption{TCA$^2$}
     \end{subfigure}
     \vspace{-1em}
        \caption{The visualization of original face image and TCA$^2$ generated adversarial images. The given attribute is "\textit{eyeglasses}" and its corresponding text prompt is "\textit{A face with eyeglasses}".}
       
\end{figure*}
\begin{figure*}
     \centering
     \begin{subfigure}[b]{0.23\textwidth}
         \centering
         \includegraphics[width=\textwidth]{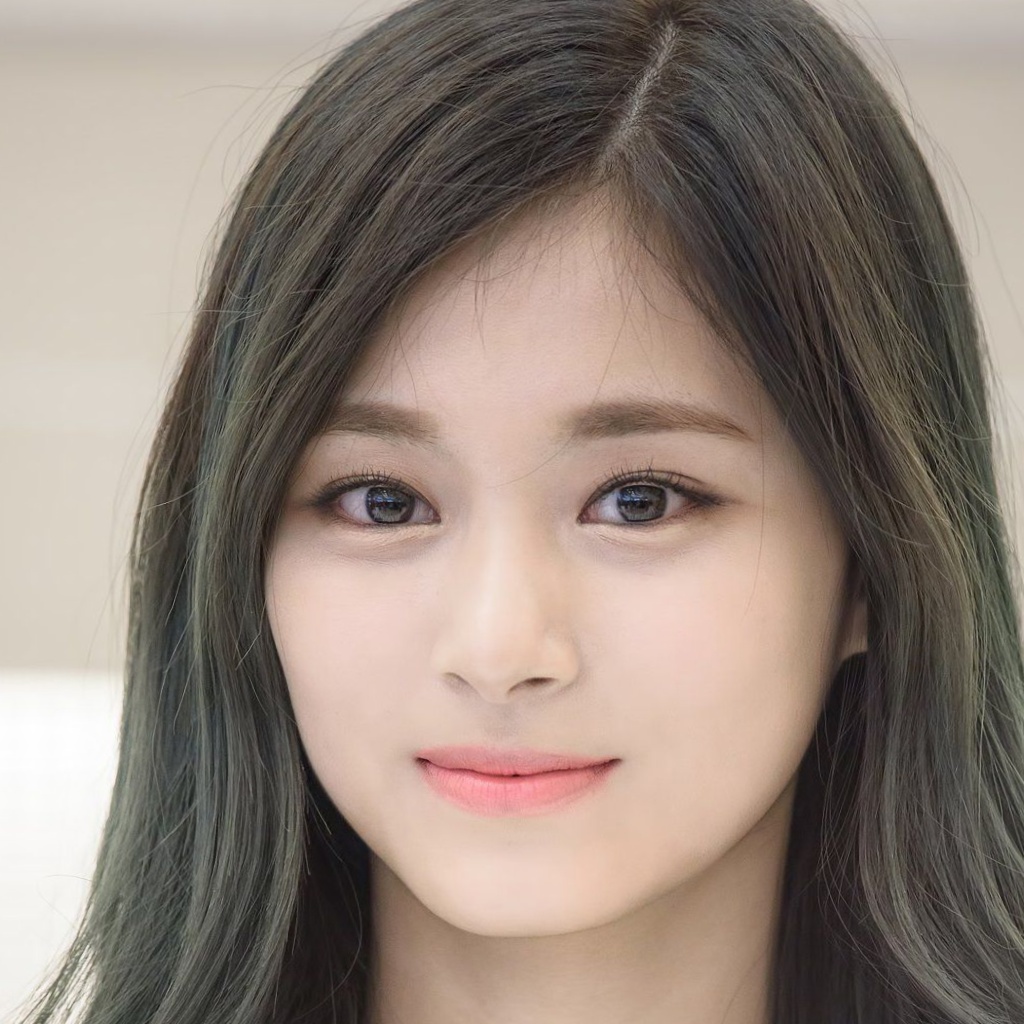}
         \caption{Original}
     \end{subfigure}
     \hfill
     \begin{subfigure}[b]{0.23\textwidth}
         \centering
         \includegraphics[width=\textwidth]{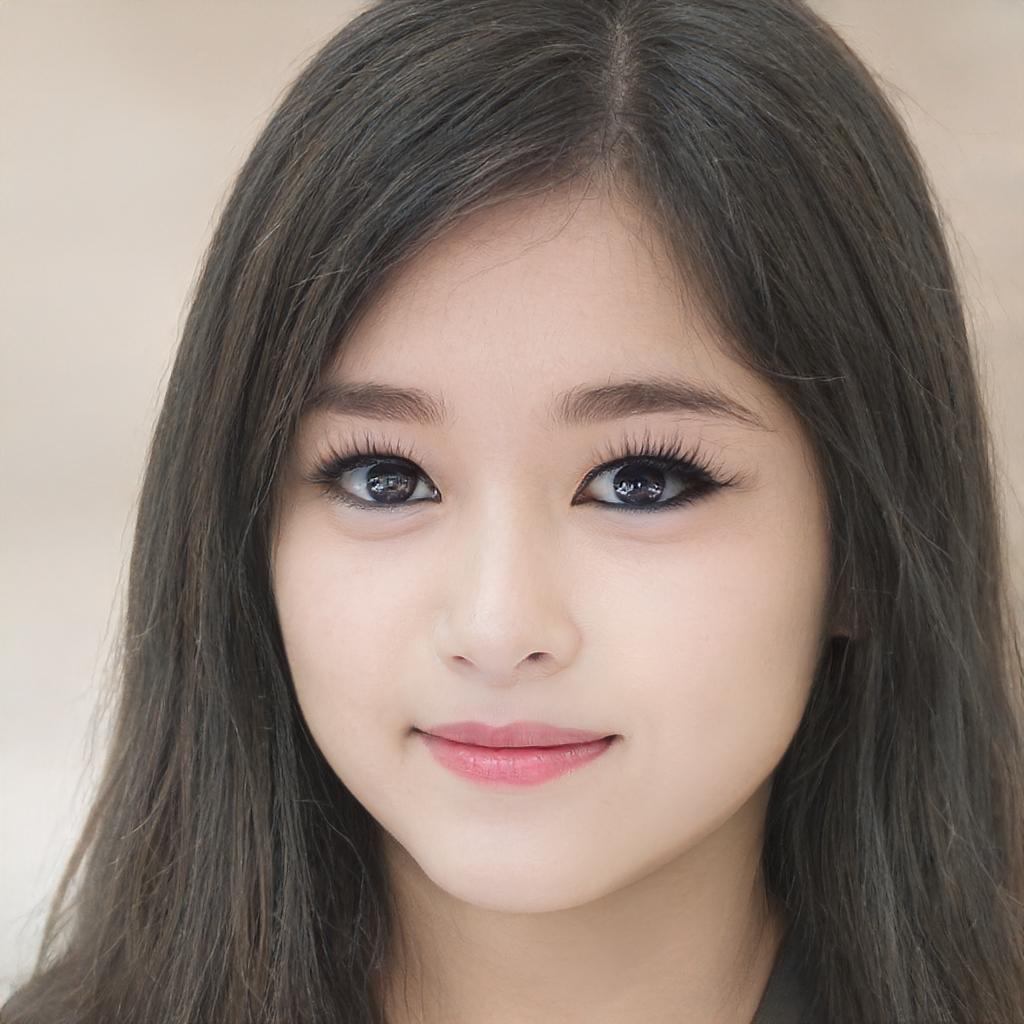}
         \caption{TCA$^2$}
     \end{subfigure}
     \hfill
     \begin{subfigure}[b]{0.23\textwidth}
         \centering
         \includegraphics[width=\textwidth]{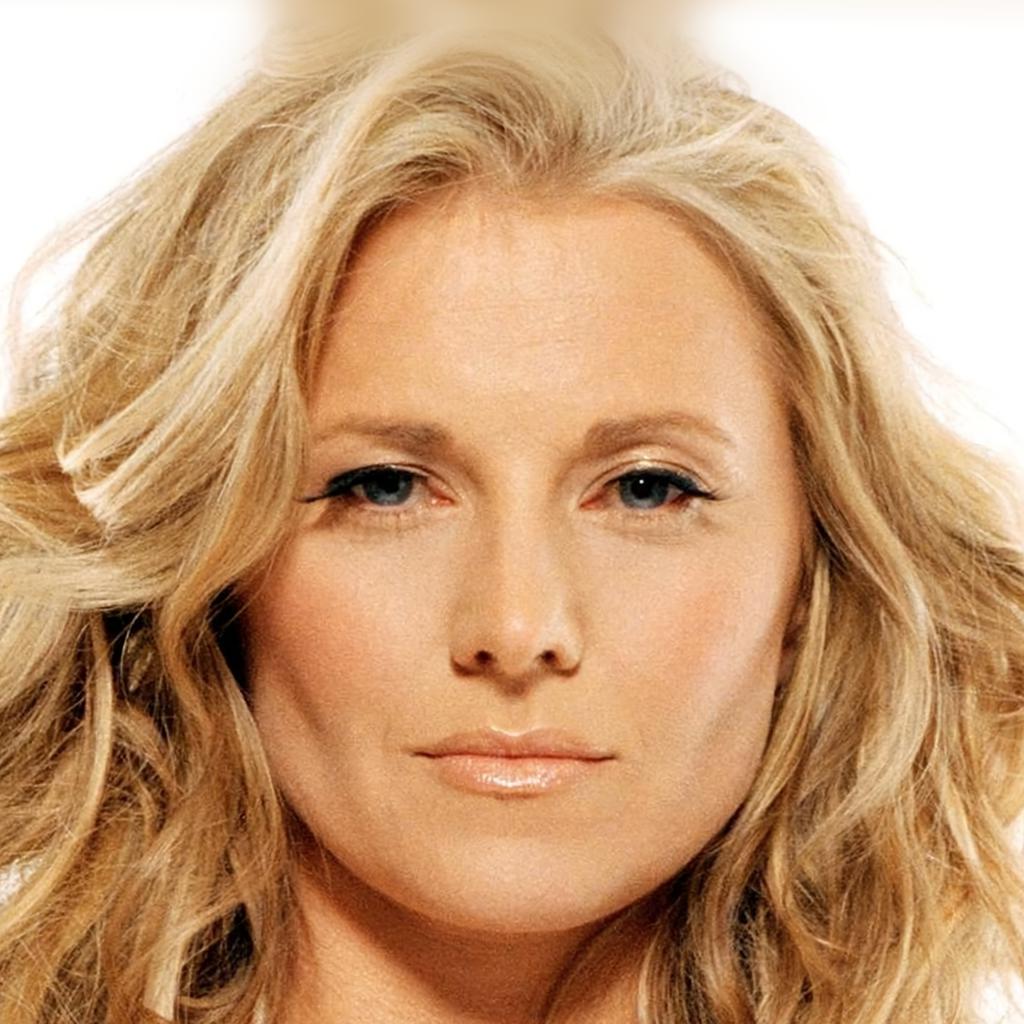}
         \caption{Original}
     \end{subfigure}
     \hfill
     \begin{subfigure}[b]{0.23\textwidth}
         \centering
         \includegraphics[width=\textwidth]{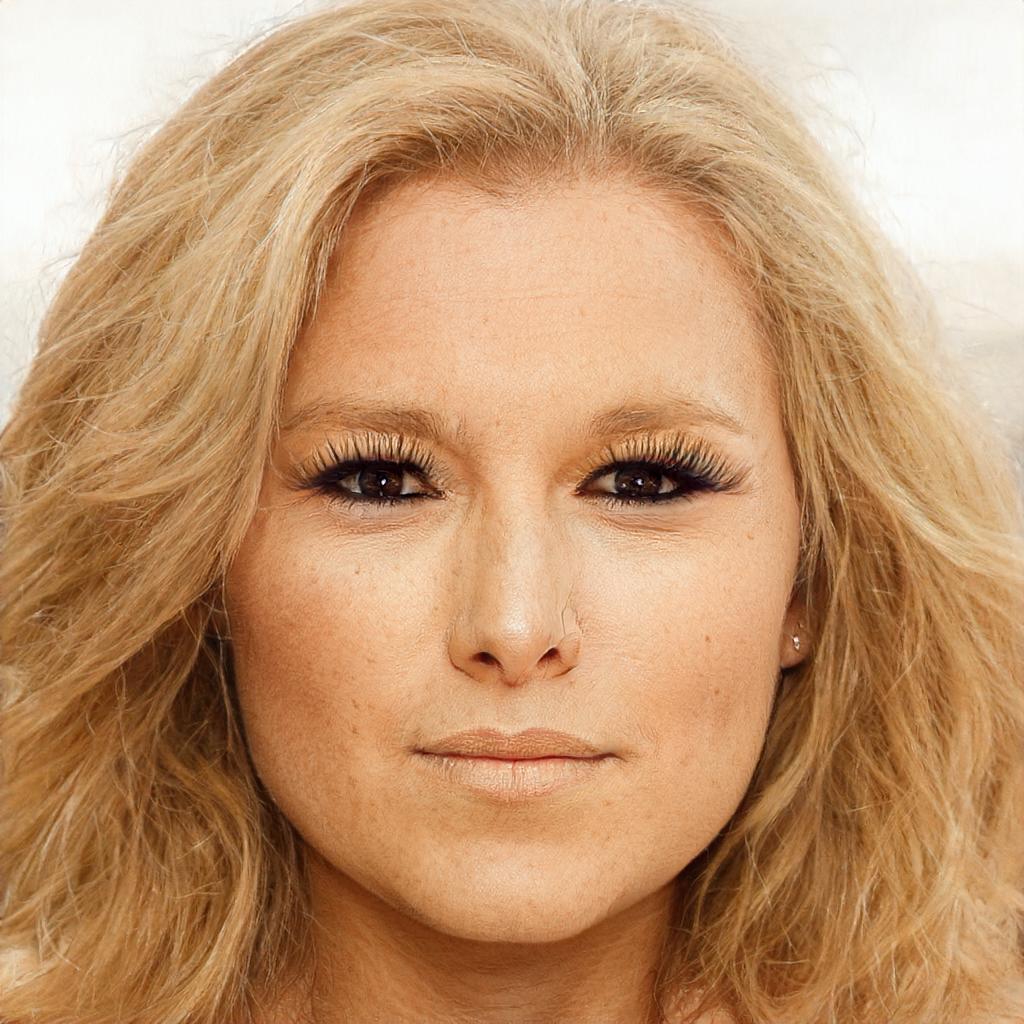}
         \caption{TCA$^2$}
     \end{subfigure}
     \vspace{-1em}
        \caption{The visualization of original face image and TCA$^2$ generated adversarial images. The given attribute is "\textit{heavy makeup}" and its corresponding text prompt is "\textit{A face with heavy makeup}".}
        
\end{figure*}
\begin{figure*}
     \centering
     \begin{subfigure}[b]{0.23\textwidth}
         \centering
         \includegraphics[width=\textwidth]{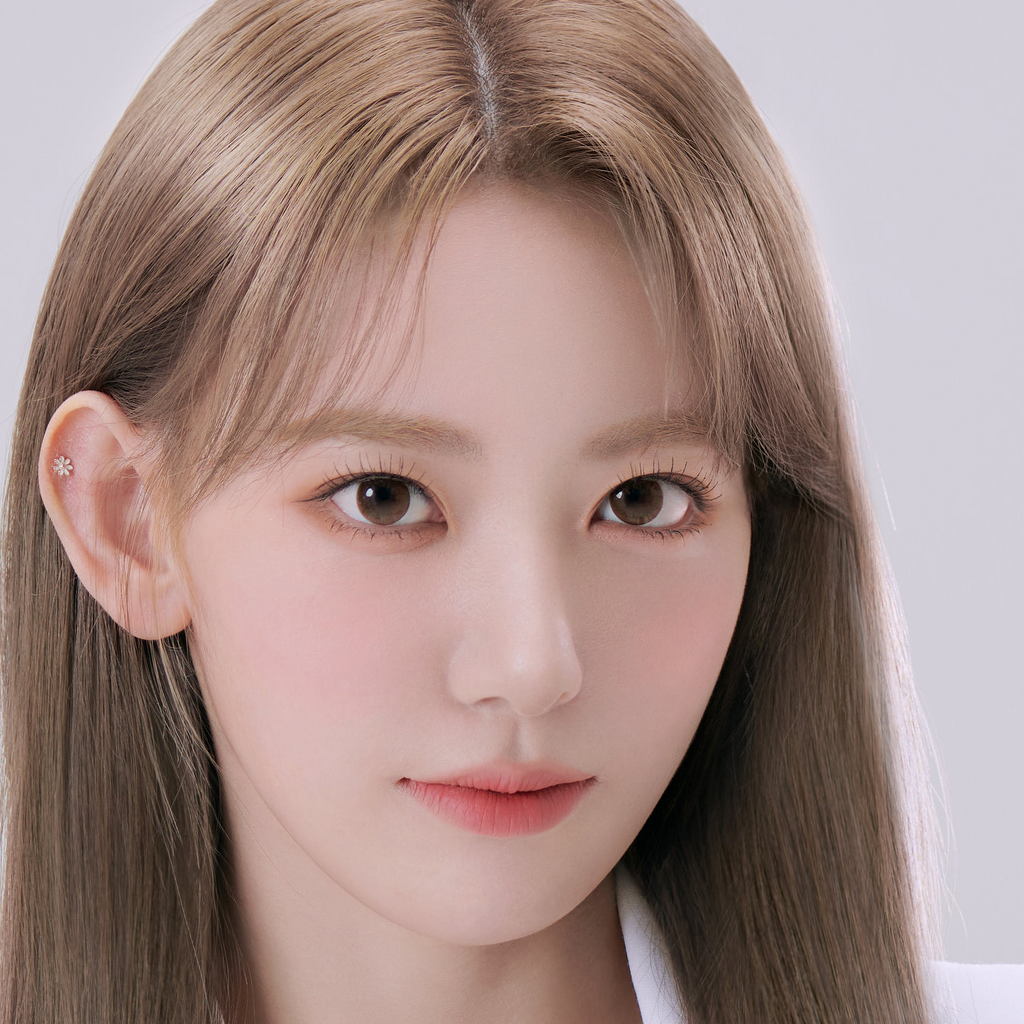}
         \caption{Original}
     \end{subfigure}
     \hfill
     \begin{subfigure}[b]{0.23\textwidth}
         \centering
         \includegraphics[width=\textwidth]{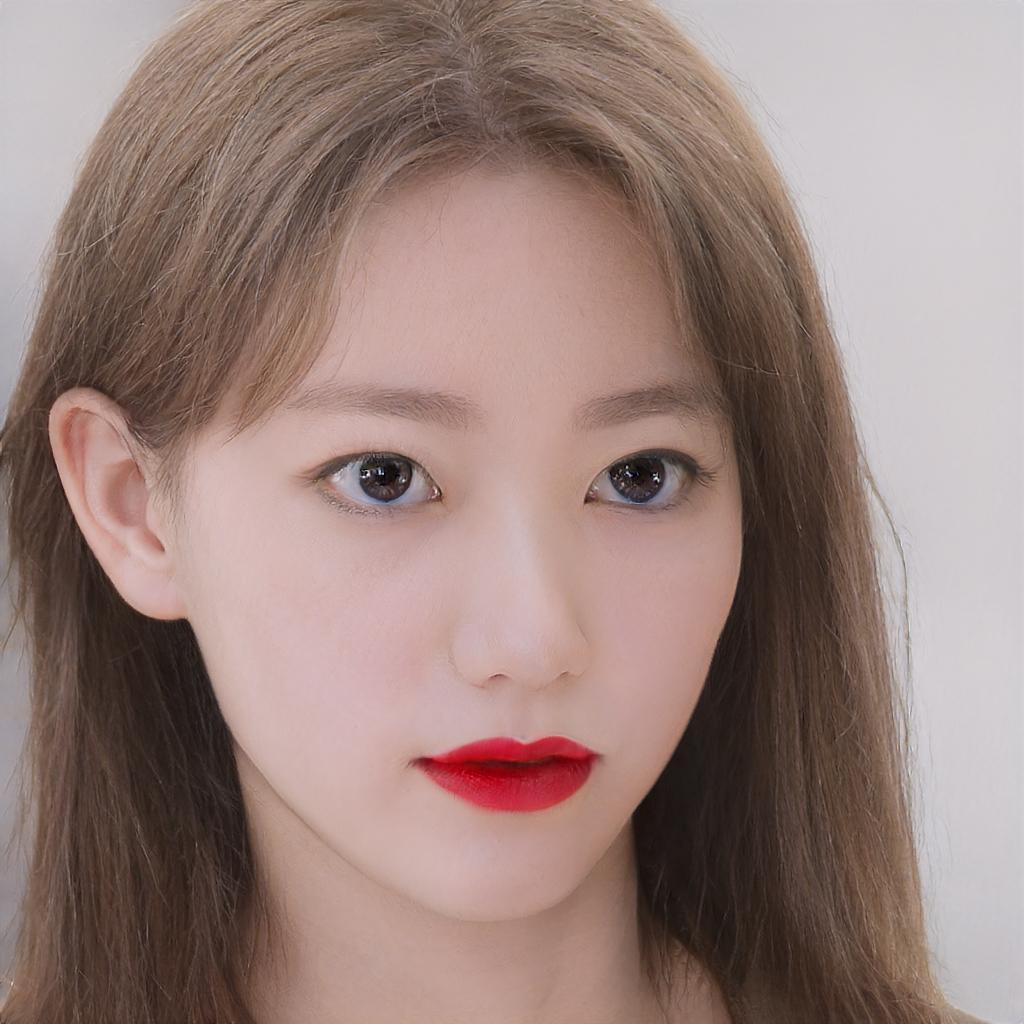}
         \caption{TCA$^2$}
     \end{subfigure}
     \hfill
     \begin{subfigure}[b]{0.23\textwidth}
         \centering
         \includegraphics[width=\textwidth]{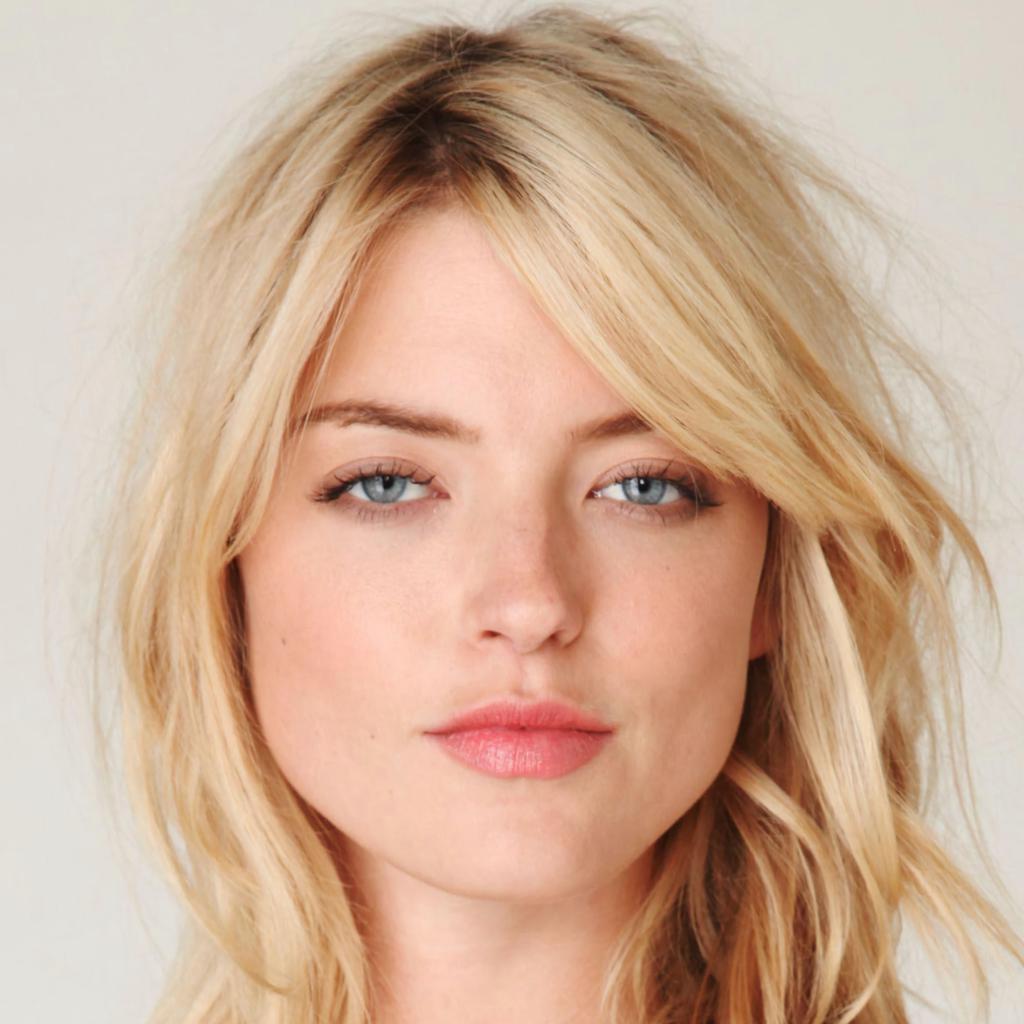}
         \caption{Original}
     \end{subfigure}
     \hfill
     \begin{subfigure}[b]{0.23\textwidth}
         \centering
         \includegraphics[width=\textwidth]{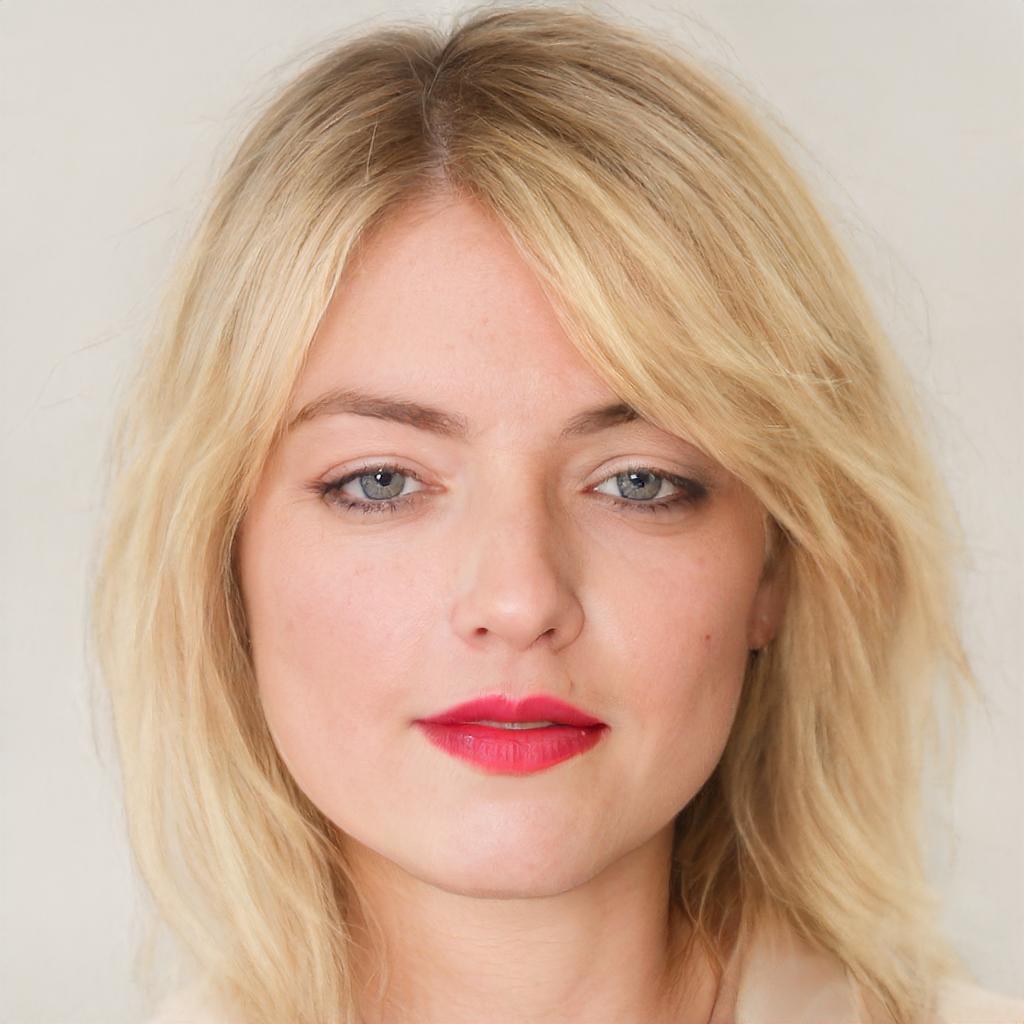}
         \caption{TCA$^2$}
     \end{subfigure}
     \vspace{-1em}
        \caption{The visualization of original face image and TCA$^2$ generated adversarial images. The given attribute is "\textit{wearing lipstick}" and its corresponding text prompt is "\textit{A face wearing lipstick}".}
       
\end{figure*}
\begin{figure*}
     \centering
     \begin{subfigure}[b]{0.23\textwidth}
         \centering
         \includegraphics[width=\textwidth]{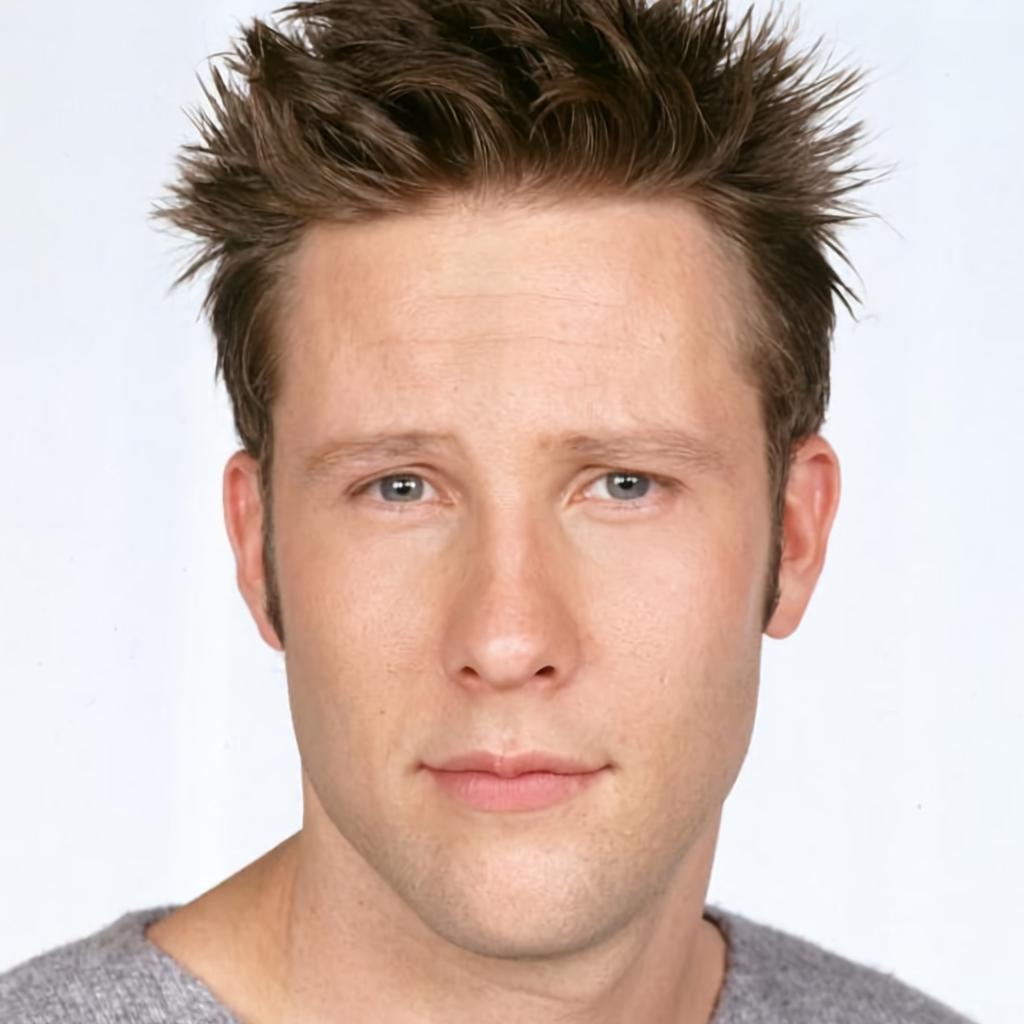}
         \caption{Original}
     \end{subfigure}
     \hfill
     \begin{subfigure}[b]{0.23\textwidth}
         \centering
         \includegraphics[width=\textwidth]{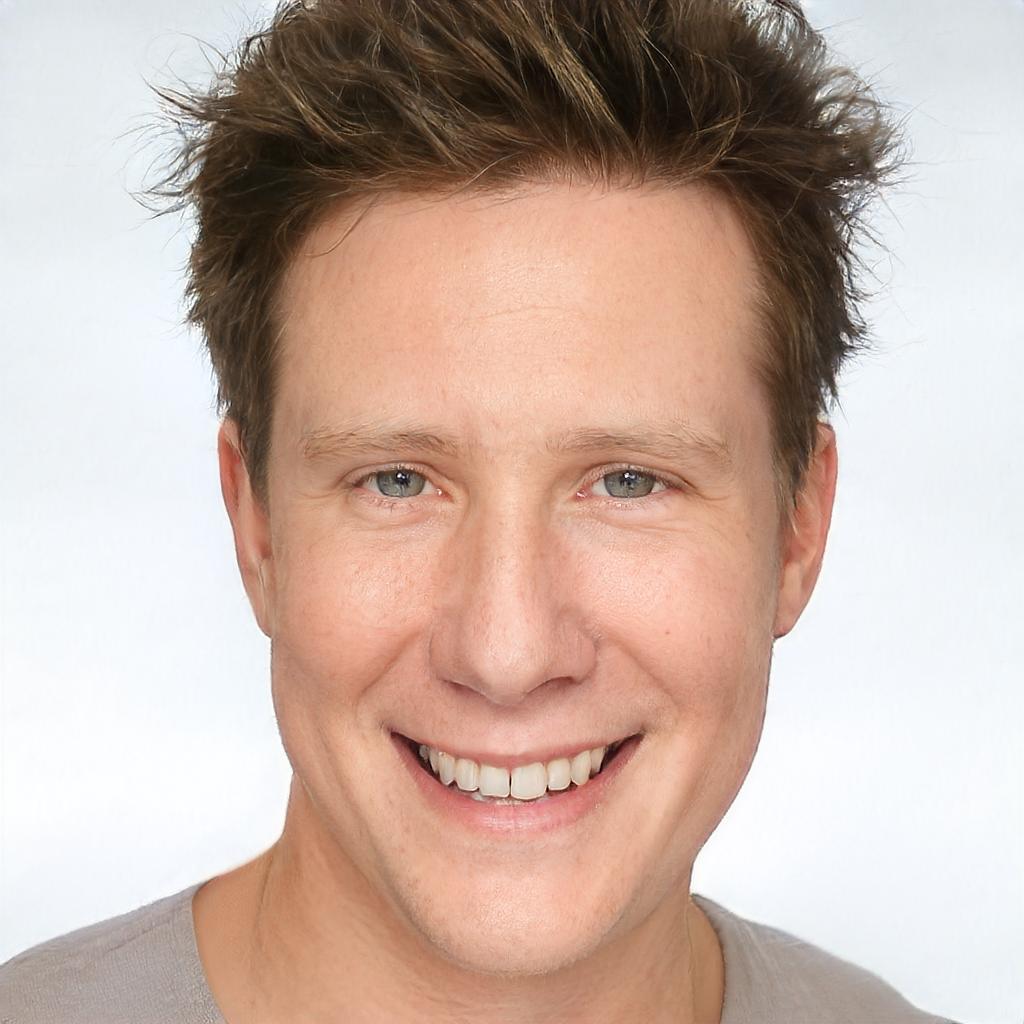}
         \caption{TCA$^2$}
     \end{subfigure}
     \hfill
     \begin{subfigure}[b]{0.23\textwidth}
         \centering
         \includegraphics[width=\textwidth]{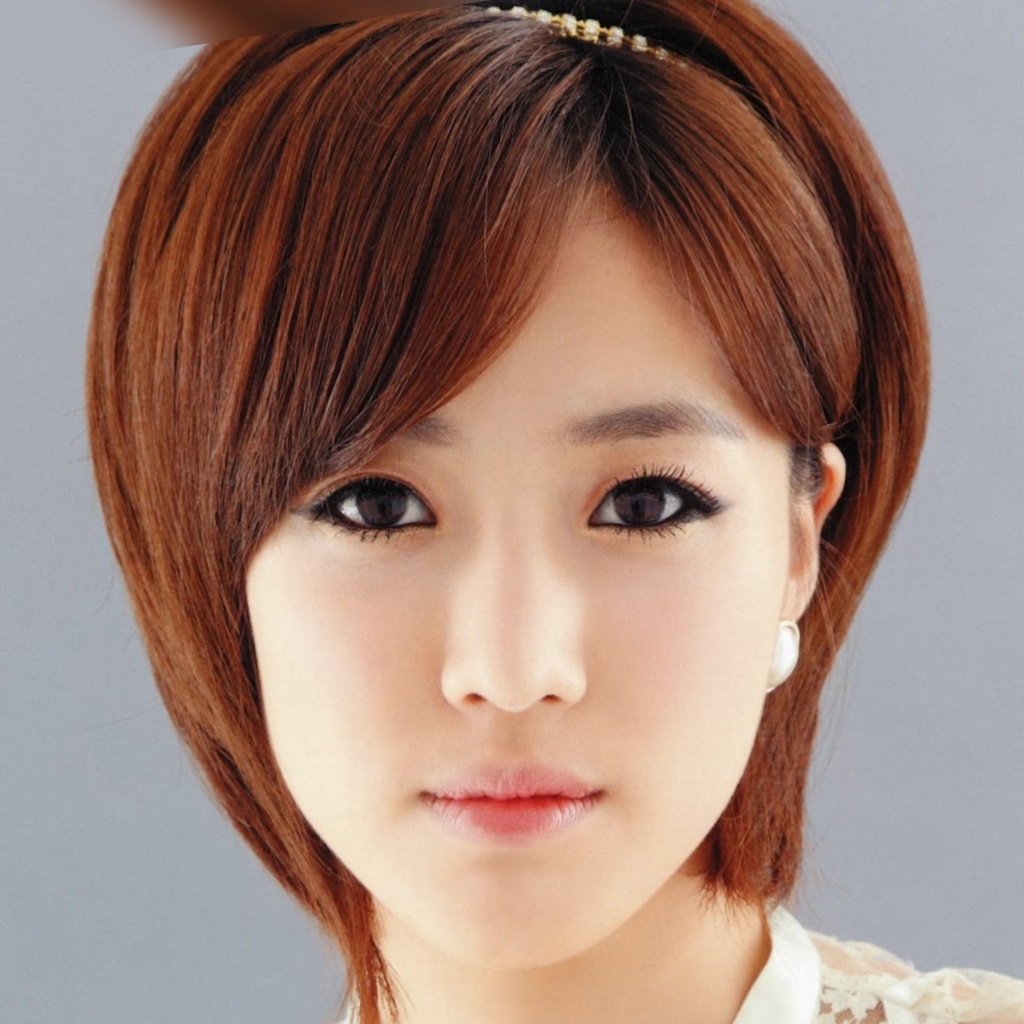}
         \caption{Original}
     \end{subfigure}
     \hfill
     \begin{subfigure}[b]{0.23\textwidth}
         \centering
         \includegraphics[width=\textwidth]{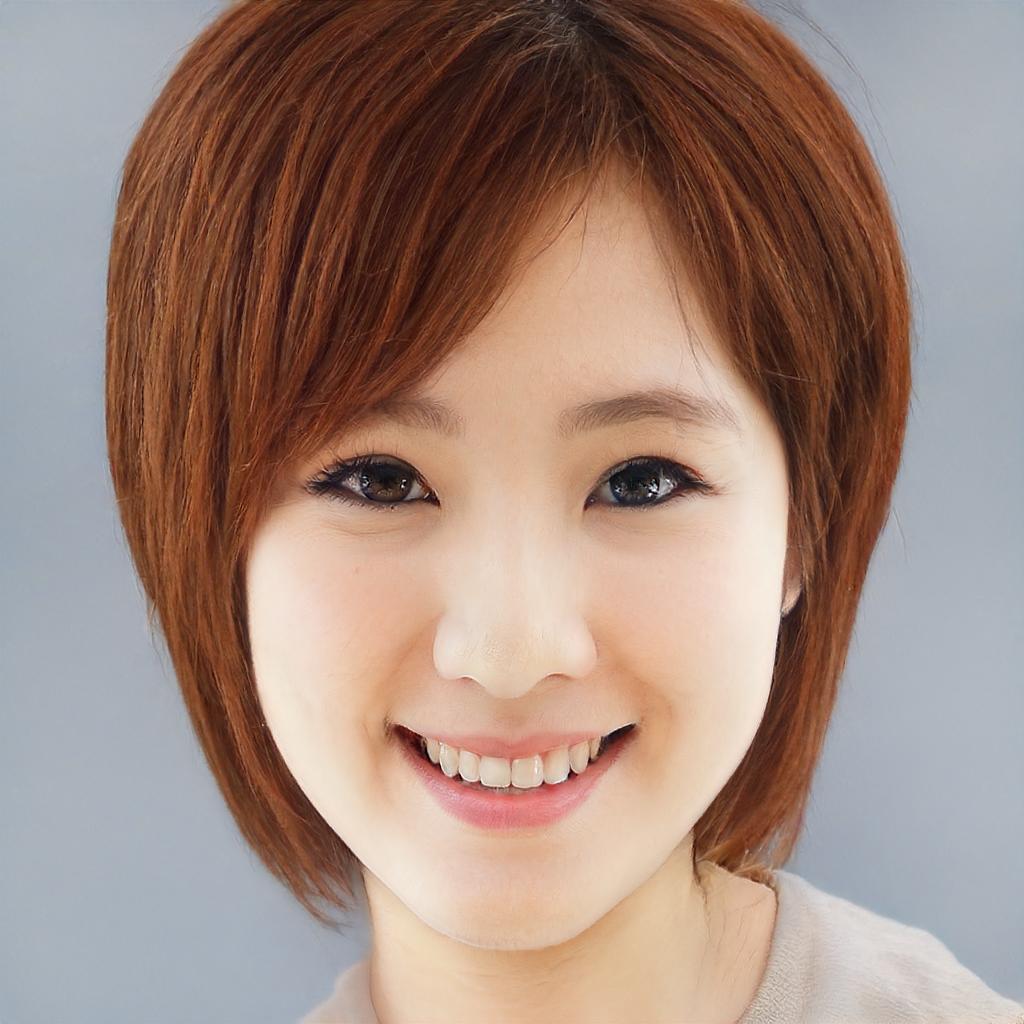}
         \caption{TCA$^2$}
     \end{subfigure}
     \vspace{-1em}
        \caption{The visualization of original face image and TCA$^2$ generated adversarial images. The given attribute is "\textit{smiling}" and its corresponding text prompt is "\textit{A 
smiling face}".}
        
\end{figure*}
\begin{figure*}
     \centering
     \begin{subfigure}[b]{0.23\textwidth}
         \centering
         \includegraphics[width=\textwidth]{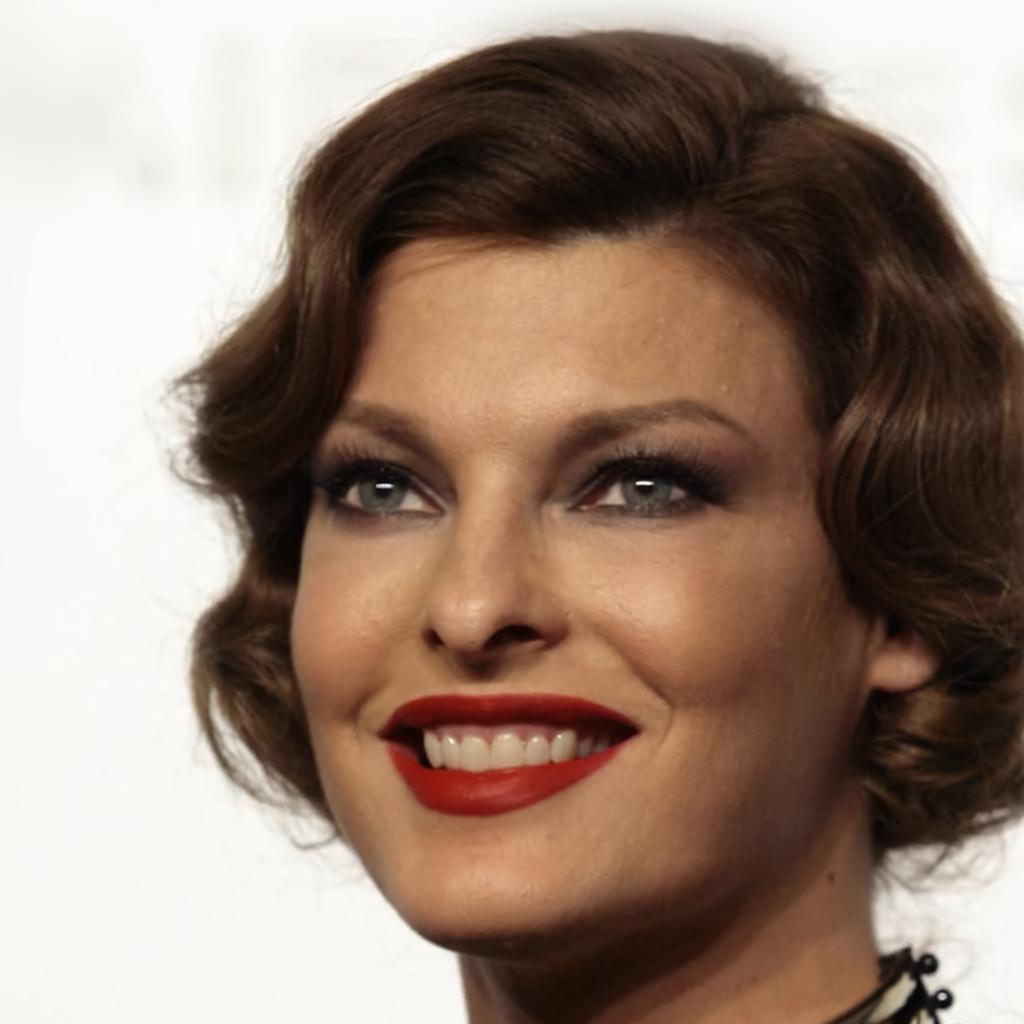}
         \caption{Original}
     \end{subfigure}
     \hfill
     \begin{subfigure}[b]{0.23\textwidth}
         \centering
         \includegraphics[width=\textwidth]{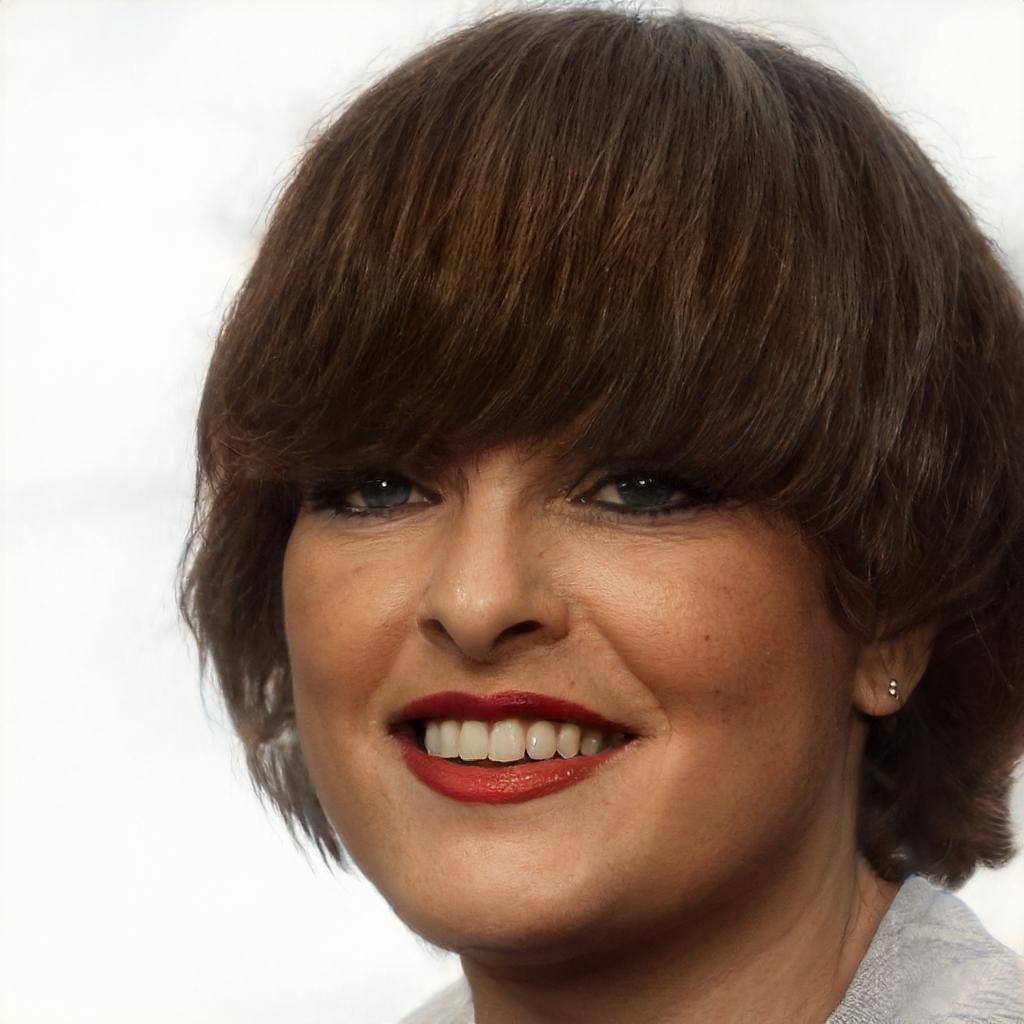}
         \caption{TCA$^2$}
     \end{subfigure}
     \hfill
     \begin{subfigure}[b]{0.23\textwidth}
         \centering
         \includegraphics[width=\textwidth]{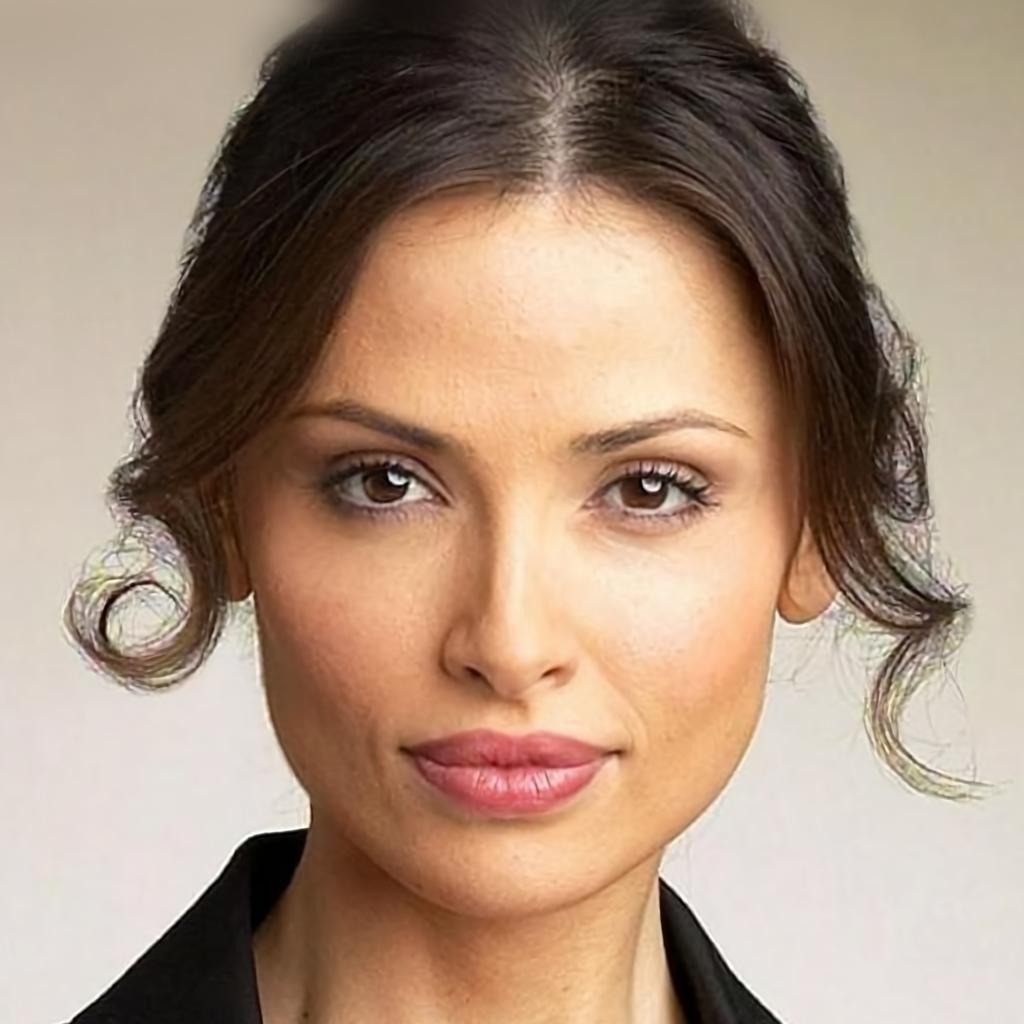}
         \caption{Original}
     \end{subfigure}
     \hfill
     \begin{subfigure}[b]{0.23\textwidth}
         \centering
         \includegraphics[width=\textwidth]{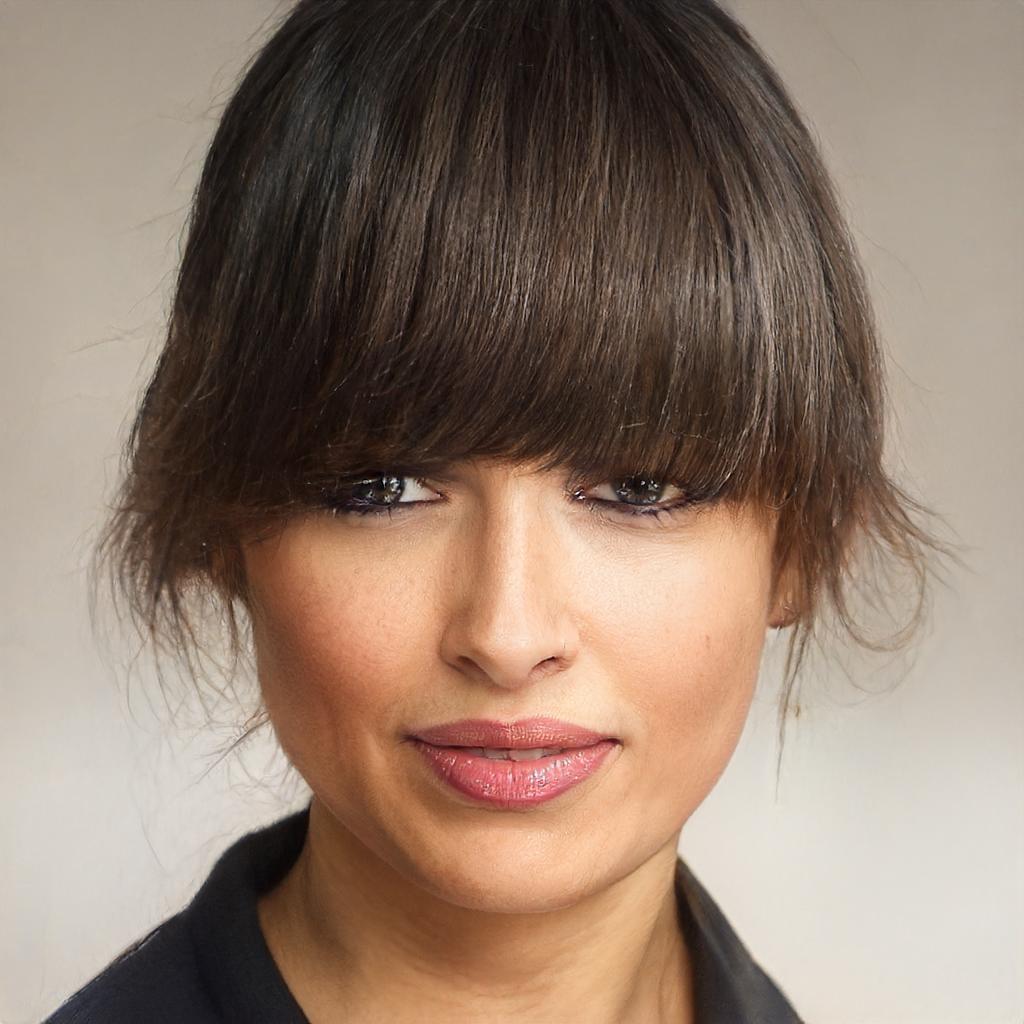}
         \caption{TCA$^2$}
     \end{subfigure}
     \vspace{-1em}
        \caption{The visualization of original face image and TCA$^2$ generated adversarial images. The given attribute is "\textit{bangs}" and its corresponding text prompt is "\textit{A face with bangs}".}
        
\end{figure*}
\begin{figure*}
     \centering
     \begin{subfigure}[b]{0.23\textwidth}
         \centering
         \includegraphics[width=\textwidth]{img1/16897.jpg}
         \caption{Original}
     \end{subfigure}
     \hfill
     \begin{subfigure}[b]{0.23\textwidth}
         \centering
         \includegraphics[width=\textwidth]{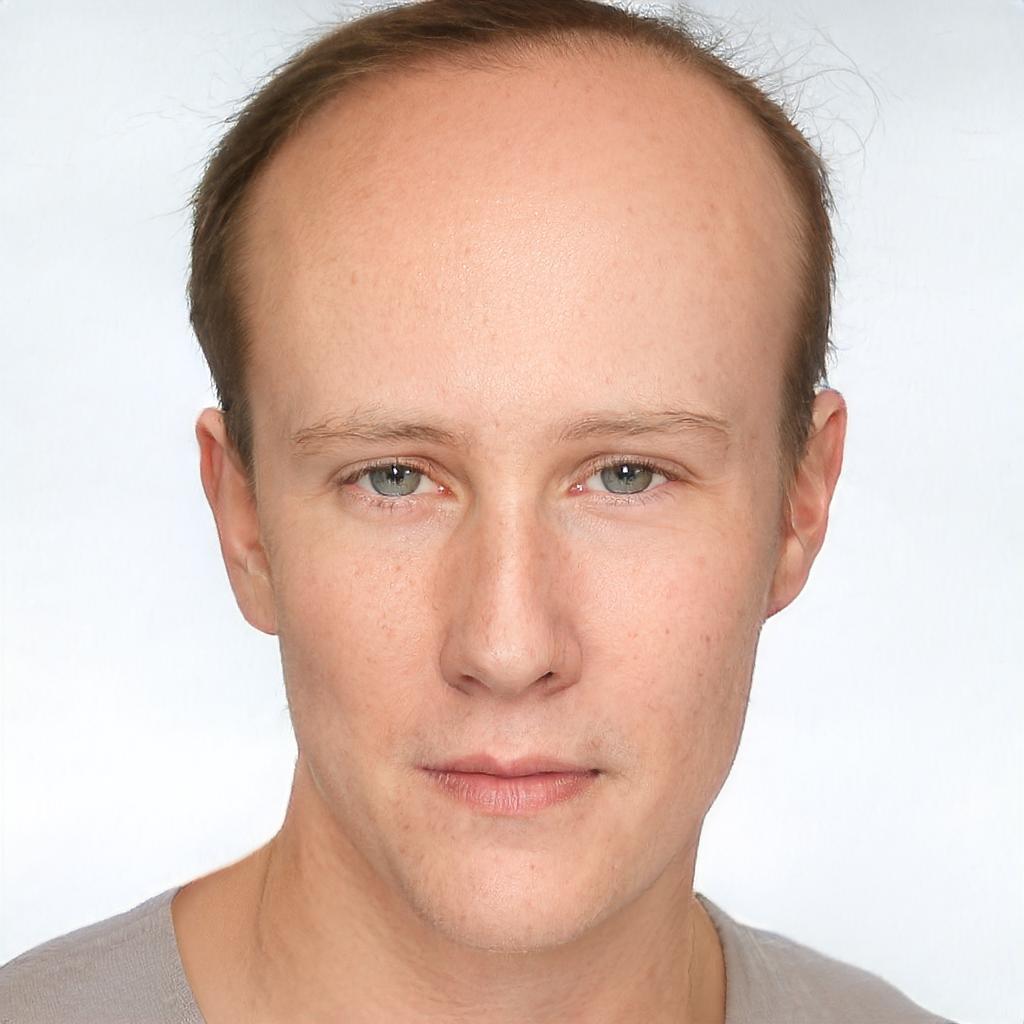}
         \caption{TCA$^2$}
     \end{subfigure}
     \hfill
     \begin{subfigure}[b]{0.23\textwidth}
         \centering
         \includegraphics[width=\textwidth]{img1/24499.jpg}
         \caption{Original}
     \end{subfigure}
     \hfill
     \begin{subfigure}[b]{0.23\textwidth}
         \centering
         \includegraphics[width=\textwidth]{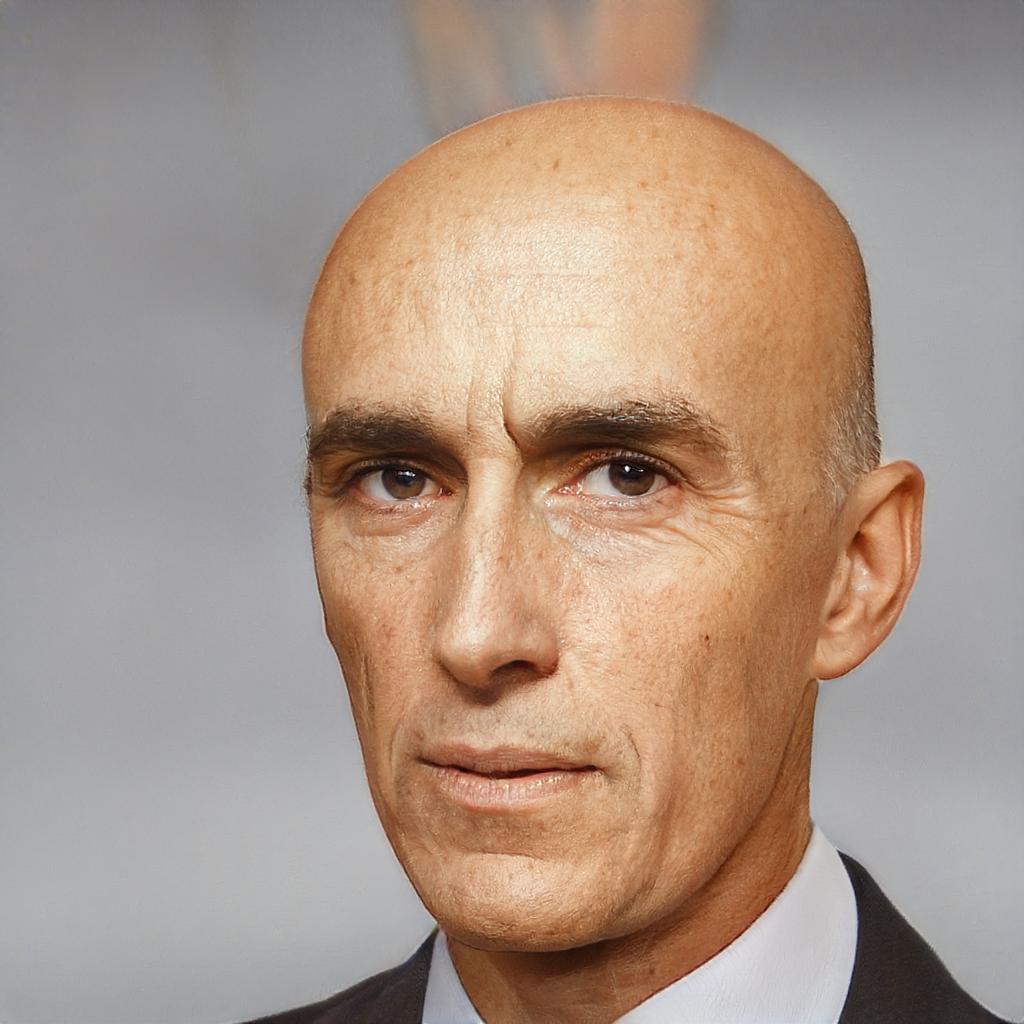}
         \caption{TCA$^2$}
     \end{subfigure}
     \vspace{-1em}
        \caption{The visualization of original face image and TCA$^2$ generated adversarial images. The given attribute is "\textit{receding hairline}" and its corresponding text prompt is "\textit{A face with receding hairline}".}
        
\end{figure*}

\end{document}